%% file: iclr2027_conference.tex
\documentclass{article} 
\usepackage{iclr2027_conference,times}

\input{math_commands.tex}

\usepackage{hyperref}
\usepackage{url}

\usepackage{enumitem}
\usepackage[utf8]{inputenc} 
\usepackage[T1]{fontenc}    
\usepackage{hyperref}       
\usepackage{url}            
\usepackage{booktabs}       
\usepackage{amsfonts}       
\usepackage{nicefrac}       
\usepackage{microtype}      
\usepackage{xcolor}         

\newcommand{\xhdr}[1]{{\noindent\bfseries #1}.}

\usepackage{xspace}
\newcommand{\method}{\texttt{ExpHarness}\xspace}

\usepackage{amsmath}
\usepackage{amssymb}
\usepackage{mathtools}
\usepackage{amsthm}

\usepackage[table]{xcolor}
\usepackage{multirow}
\usepackage{makecell}
\usepackage{pifont}       
\usepackage{bbding}      
\usepackage{fontawesome}
\definecolor{darkblue}{RGB}{0,0,139}
\usepackage{subcaption}
\usepackage{arydshln}
\usepackage{tabularx}
\usepackage{mdframed}
\usepackage[most]{tcolorbox}
\usepackage{arydshln}
\usepackage{adjustbox}
\usepackage{wrapfig}
\usepackage{titletoc}
\usepackage{booktabs,array,colortbl,graphicx}

\usepackage{algorithm}
\usepackage{algorithmic}

\definecolor{table-blue}{RGB}{173, 216, 230}
\definecolor{row-highlight}{RGB}{220, 230, 242}
\definecolor{qwen-header}{RGB}{235, 240, 248}
\definecolor{llama-header}{RGB}{248, 244, 235}
\definecolor{section-gray}{RGB}{245, 245, 245}
\definecolor{group-header}{RGB}{230, 235, 245}

\definecolor{row-alt}{RGB}{248, 250, 253}      
\definecolor{hl-llama}{RGB}{234, 242, 255}     
\definecolor{hl-qwen}{RGB}{236, 250, 240}      

\definecolor{zhz_gray}{rgb}{0.8,0.8,0.8}
\definecolor{darkgreen}{rgb}{0.0, 0.5, 0.0} 
\definecolor{darkred}{rgb}{0.5, 0.0, 0.0}   

\title{ExpHarness: Model-Agnostic Experience Learning through a Trainable Harness}

\author{%
\textbf{Tao Feng}\textsuperscript{1},\;\;
\textbf{Chongrui Ye}\textsuperscript{1}, \;\;
\textbf{Fangxu Yu}\textsuperscript{2}, \;\;
\textbf{Tianyang Luo}\textsuperscript{\textbf{1}},\;\;
\textbf{Jingjun Xu}\textsuperscript{\textbf{3}},\;\;
\textbf{Xueqiang Xu}\textsuperscript{\textbf{1}}, \\
\textbf{Haozhen Zhang}\textsuperscript{\textbf{4}},\;\; 
\textbf{Weizhi Zhang}\textsuperscript{\textbf{5}},\;\; 
\textbf{Zijie Lei}\textsuperscript{\textbf{6}},\;\;
\textbf{Zhigang Hua}\textsuperscript{\textbf{6}},\;\;
\textbf{Yan Xie}\textsuperscript{\textbf{6}},\;\;
\textbf{Shuang Yang}\textsuperscript{\textbf{6}},\;\;\\
\textbf{Jiaxuan You}\textsuperscript{\textbf{1}}\;\;
\\
\textsuperscript{1}University of Illinois Urbana-Champaign,\;\;
\textsuperscript{2}University of Maryland,\;\;
\textsuperscript{3}Harvard University,\;\;\\
\textsuperscript{4}Nanyang Technological University,\;\;
\textsuperscript{5}University of Illinois Chicago,\;\;
\textsuperscript{6}Meta Monetization AI
}

\iclrfinalcopy 
\begin{document}

\maketitle

\begin{abstract}
Large language model (LLM) agents increasingly operate within a harness, the
scaffolding that determines what enters the executor's context, yet the
experience they accumulate across tasks rarely flows back into this harness.
Existing approaches include executor fine-tuning and external memory retrieval,
but combining task-adaptive retrieval with experience reuse across frozen
executors remains challenging. To this end, we introduce \method, a learnable experience harness
that improves frozen and replaceable LLM executors without modifying their
parameters. Specifically, \method distills trajectories into reusable skills and
failure lessons within a self-evolving experience graph, and trains a
lightweight retrieval copilot that decides, per task, how broadly to explore the
graph and how strongly to favor historically useful experiences over merely
similar ones. The copilot is optimized with reinforcement learning from a
utility-grounded reward combining the with/without-experience score difference
and an absolute-performance term; the same reward updates the graph during training. Extensive
experiments on ExpSuite, spanning 10 static benchmarks and 2 agentic
environments, show that \method improves over the strongest baseline by
$12.1\%$ and $4.5\%$ on static tasks and by $21.4\%$ and $12.7\%$ on agentic
tasks with the smaller and larger executors, respectively, while reducing
interaction steps by up to $21.6\%$. Transfer experiments further examine reuse of the learned harness across
executors of different scales and reasoning capabilities, with joint graph and
copilot transfer performing closest to target-specific training among the
evaluated transfer variants.
\end{abstract}

\begin{center}
    \hypersetup{pdfborder={0 0 0}}
    \href{https://ulab-uiuc.github.io/ExpHarness/}{\textcolor{darkblue}{\faGlobe\ \texttt{Website}}}
    \quad
    \href{https://github.com/ulab-uiuc/ExpHarness}{\textcolor{darkblue}{\faGithub\ \texttt{Code}}}
\end{center}

\section{Introduction}
\label{sec:intro}

LLM agents have demonstrated strong capabilities in complex tasks requiring
reasoning, tool use, and multi-step environment interaction~\citep{yao2022react,shinn2023reflexion}.
In practice, such an agent is a frozen executor LLM wrapped in a \emph{harness},
the surrounding scaffolding that determines what information enters the
executor's context at each step.
A key bottleneck, however, is that this harness rarely learns: each task is
solved largely from scratch, and successful strategies, failure lessons, and
transferable insights accumulated from prior interactions are discarded rather
than systematically reused.
A natural remedy is to fine-tune the executor on its own experience, but this
solution becomes increasingly inflexible as LLM capabilities evolve rapidly.
Whenever a stronger executor is released, executor-specific training may need to be
repeated, and many capable LLMs are too large or inaccessible for
parameter-level modification in the first place.
These observations motivate a different locus for experience learning:
\textit{rather than training the executor, can an LLM agent learn from
accumulated experience by training its harness, keeping the executor itself
frozen and replaceable?}

Existing experience learning methods, as summarized in Table~\ref{tab:intro},
provide partial solutions to this problem, including textual experience
distillation~\citep{zhao2024expel}, memory organization~\citep{chhikara2025mem0},
utility-aware experience selection~\citep{zhang2026memrl}, and adaptive
retrieval or search policies~\citep{jiang2025s3}.
These methods emphasize different parts of the experience pipeline. We study
how relational storage, adaptive retrieval, and outcome-based feedback can be
combined in an external harness and evaluated when the executor changes.
Building such an experience harness is non-trivial for three reasons.
\textit{First, surface relevance is not the same as experience utility.}
An experience that appears similar to the current task may provide little
downstream benefit~\citep{shi2023large,cuconasu2024power}, while a useful experience may encode a transferable
strategy, shared sub-goal, or failure pattern that is not among the nearest
neighbors of the task embedding, since past experiences are often connected
through common strategies, environmental constraints, or recurring mistakes
rather than surface similarity~\citep{gutierrez2024hipporag}.
\textit{Second, the harness must adapt to each task.}
Some tasks benefit from broad exploration over related experiences, while
others require focused selection of a few high-utility ones, so a fixed
retrieval rule cannot serve all tasks~\citep{mallen2023not, jeong2024adaptive}.
\textit{Third, the harness must remain useful when the executor changes.}
Executors vary in reasoning, planning, and instruction-following ability, and a
harness learned with one executor should carry over to a stronger, weaker, or
reasoning-oriented executor without being rebuilt from scratch~\citep{wei2023larger, sclar2024quantifying}.

\begin{table*}[t]
    \caption{
        \textbf{Comparison with representative experience learning methods.}
        The columns summarize four design features for the representative
        methods listed. \method combines an experience graph,
        structure-aware expansion, utility-aware ranking, and adaptive retrieval
        around a frozen executor.
    }
    \vspace{-2.5mm}
    \label{tab:intro}
    \centering
    \setlength{\tabcolsep}{8pt}
    \resizebox{1\textwidth}{!}{
    \small
    \begin{tabular}{lcccc}
        \toprule
        \textbf{Method}
            & \textbf{Relational Experience Store}
            & \textbf{Structure-Aware Expansion}
            & \textbf{Utility-Aware Ranking}
            & \textbf{Adaptive Retrieval} \\
        \midrule
        ExpeL~\citep{zhao2024expel}
            & \textcolor{red}{\ding{55}}
            & \textcolor{red}{\ding{55}}
            & \textcolor{red}{\ding{55}}
            & \textcolor{red}{\ding{55}} \\
        Mem0~\citep{chhikara2025mem0}
            & \textcolor{green!50!black}{\ding{51}}
            & \textcolor{red}{\ding{55}}
            & \textcolor{red}{\ding{55}}
            & \textcolor{red}{\ding{55}} \\
        MemRL~\citep{zhang2026memrl}
            & \textcolor{red}{\ding{55}}
            & \textcolor{red}{\ding{55}}
            & \textcolor{green!50!black}{\ding{51}}
            & \textcolor{red}{\ding{55}} \\
        S3~\citep{jiang2025s3}
            & \textcolor{red}{\ding{55}}
            & \textcolor{red}{\ding{55}}
            & \textcolor{red}{\ding{55}}
            & \textcolor{green!50!black}{\ding{51}} \\
        \midrule
        \rowcolor{cyan!10}
        \method
            & \textcolor{green!50!black}{\ding{51}}
            & \textcolor{green!50!black}{\ding{51}}
            & \textcolor{green!50!black}{\ding{51}}
            & \textcolor{green!50!black}{\ding{51}} \\
        \bottomrule
    \end{tabular}}
\vspace{-20pt}
\end{table*}

To address these challenges, we propose \textbf{\method}, a learnable experience
harness that improves frozen LLM executors without modifying their parameters.
Rather than updating the executor, \method treats it as a replaceable task
solver and places all learning in the harness, so that the learned harness can
be carried over rather than rebuilt when stronger LLMs become available.
Specifically, the harness comprises two core components---an experience graph
and a retrieval copilot---jointly updated through utility-grounded feedback.
First, the \emph{Experience Graph}, a graph-structured experience store,
stores compact experience units summarized from historical trajectories, including skills distilled from successful
trajectories and lessons distilled from failures, and links related units so
that retrieval can reach experiences sharing strategies, sub-goals, or failure
patterns beyond the nearest neighbors of the task.
Second, a lightweight \emph{Retrieval Copilot} predicts task-adaptive controls
that determine how broadly retrieval explores the graph and how strongly the
final ranking favors historical utility over semantic relevance.
To update both components, \emph{utility-grounded feedback} combines the executor's score with
retrieved experiences and its score difference relative to a no-experience run.
This feedback optimizes the copilot with reinforcement learning and updates
shared reward estimates for the retrieved entries, supplementing semantic
relevance with historical task outcomes.

We evaluate \method on ExpSuite, covering both single-turn static tasks
(question answering, mathematical reasoning, and code generation) and
multi-step agentic environments (ALFWorld and AppWorld).
Across static tasks, \method improves over the strongest baseline by
$12.1\%$ with the smaller executor and $4.5\%$ with the larger executor.
The advantage becomes more pronounced in agentic environments, where \method
improves the weighted average score over the strongest baseline by $21.4\%$
and $12.7\%$ with the smaller and larger executors, respectively, while
reducing average interaction steps by $12.7\%$ and $21.6\%$ compared with the
competing baseline with the lowest average step count.
Furthermore, transfer experiments compare reuse of the store, copilot, and
full harness across executors, with joint transfer performing closest to
target-specific training among the evaluated variants. Ablation studies show
performance degradation under four variants that modify experience filtering,
memory organization, graph diffusion, and utility-aware ranking.
\section{Preliminaries}
\label{sec:preliminaries}

\vspace{-2.5mm}

\subsection{Model-Agnostic LLM Agents}
\label{sec:setting}

\vspace{-1.8mm}

We consider an LLM agent that solves a task $x \in \mathcal{X}$ by producing an output $y$,
which can be an action, answer, or code sequence, and receives a task score
$s = S(x, y) \in \mathbb{R}$ from the environment or evaluator.
The agent is built around an executor LLM $\pi_{\mathrm{exec}}$, which maps the task input
to an output, i.e., $y = \pi_{\mathrm{exec}}(x)$.
We adopt a \emph{model-agnostic} view of the executor: the experience learning mechanism
should not depend on the internal architecture, parameters, gradients, logits, or training
procedure of $\pi_{\mathrm{exec}}$.
Instead, $\pi_{\mathrm{exec}}$ is treated as a frozen and replaceable task solver, which can be
closed-source, costly to update, specialized for a domain, or newly released.
The improvement mechanism interacts with the executor only through its input-output behavior
and task-level feedback.
This setting decouples experience learning from executor training, allowing the same external
experience system to improve different frozen executors through input context rather than
parameter updates.

\vspace{-2.5mm}

\subsection{Experience-Augmented Learning}
\label{sec:experience_learning}

\vspace{-1.8mm}

To improve the executor through its input, we equip the agent with an external experience
system $\mathcal{M}$ that stores reusable knowledge distilled from historical trajectories.
A trajectory is denoted as $\tau = (x, \xi, y, s)$, where $x$ is the task input, $\xi$ denotes
the intermediate execution process, $y$ is the final response or action sequence, and
$s \in \mathbb{R}$ is the task score.
Specifically, $\xi$ corresponds to the agent interaction trace in agentic environments, and
to the intermediate thinking process in question-answering or reasoning tasks.
Each trajectory is summarized into a compact natural-language experience unit
$e = \mathrm{Summarize}(\tau)$ and stored in $\mathcal{M}$. At execution time, the agent retrieves a subset of experiences $E \subseteq \mathcal{M}$ and
injects them into the executor's input context, yielding
$y = \pi_{\mathrm{exec}}(x, E)$.
The task score $s = S(x, y)$ then depends indirectly on the choice of $E$.

Since $\pi_{\mathrm{exec}}$ is fixed, the learnable component is the \emph{retrieval policy}
$\pi_{\mathrm{ret}}$, which selects $E$ from $\mathcal{M}$ given the current task.
The learning objective is to maximize expected task performance over a task distribution
$\mathcal{D}$:
\vspace{-1mm}
\begin{equation}
    \max_{\pi_{\mathrm{ret}}}
    \;\mathbb{E}_{x \sim \mathcal{D}}
    \!\left[\,
        S\!\left(x,\, \pi_{\mathrm{exec}}\!\left(x,\, E\right)\right)
    \,\right],
    \quad E \sim \pi_{\mathrm{ret}}(\cdot \mid x, \mathcal{M}).
    \label{eq:objective}
\end{equation}
This formulation decouples execution from experience learning: $\pi_{\mathrm{exec}}$ solves the
task, while $\pi_{\mathrm{ret}}$ decides which experiences should be provided as context.

\section{\method: A Learnable Experience Harness for Frozen Executors}
\label{sec:method}

\vspace{-2.5mm}

\subsection{Overview}
\label{sec:overview}

\vspace{-1.8mm}

\begin{figure*}[t]
    \centering
    \vspace{-3mm}
    \includegraphics[width=1\linewidth]{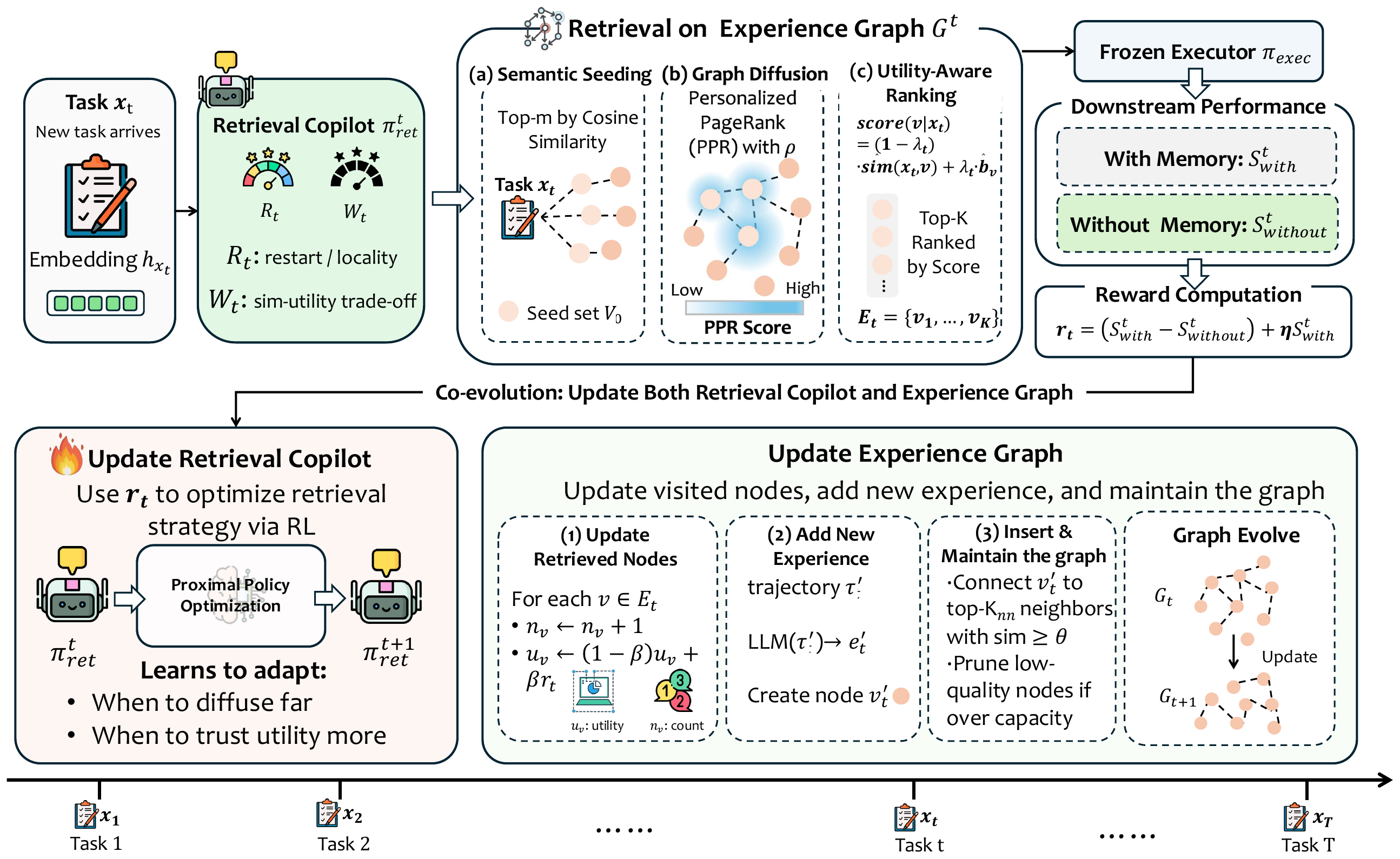}
    \vspace{-6mm}
\caption{\textbf{Overview of \method.}
\method enables a frozen and replaceable executor LLM to improve through a learnable experience harness, consisting of a self-evolving experience graph and a trainable retrieval copilot.
For each incoming task $x_t$, the task is embedded as $h_{x_t}$ and passed to the retrieval copilot $\pi_{\mathrm{ret}}^t$, which predicts two adaptive controls: $R_t$ for restart/locality control and $W_t$ for the similarity--utility trade-off.
Retrieval is performed on the current experience graph $G^t$ through three steps:
\textbf{(a) Semantic Seeding}, which selects a seed set $V_0$ by cosine similarity;
\textbf{(b) Graph Diffusion}, which expands from the seeds using personalized PageRank controlled by $\rho$;
and \textbf{(c) Utility-Aware Ranking}, which combines semantic relevance and utility confidence to select the top-$K$ experiences $E_t$.
During training, the frozen executor $\pi_{\mathrm{exec}}$ is evaluated with and without retrieved experiences, producing $s_{\mathrm{with}}^t$ and $s_{\mathrm{without}}^t$.
Their difference, together with the absolute task score, forms the reward $r_t=(s_{\mathrm{with}}^t-s_{\mathrm{without}}^t)+\eta s_{\mathrm{with}}^t$.
This reward drives a co-evolution process: it updates the retrieval copilot via PPO and updates the experience graph by refining visited-node utilities, adding new experience nodes, connecting them to similar neighbors, and pruning low-quality nodes when necessary.
The feedback and update paths operate during training. At evaluation, the
copilot and graph are fixed and only the experience-conditioned execution path
is used; the executor LLM remains frozen throughout.}
    \label{fig:framework}
\vspace{-10pt}
\end{figure*}

\method is a learnable experience harness for LLM agents. Given a task $x$, \method improves a frozen executor LLM $\pi_{\mathrm{exec}}$ by deciding which external experiences enter its context, without modifying the executor. The key idea is to decouple execution from experience learning: all learnable state resides in the harness $\mathcal{H}=(G,\pi_{\mathrm{ret}})$, consisting of an experience graph $G$ and a retrieval copilot $\pi_{\mathrm{ret}}$, which allows the same interface to be used with different frozen executors; the effectiveness of transfer is evaluated empirically. As shown in Figure~\ref{fig:framework}, these two components are coupled by a feedback-driven training loop. First, historical trajectories are compressed into experience units and maintained in an experience graph (§\ref{sec:graph_construction}). Second, the retrieval copilot predicts task-adaptive controls to search the graph via semantic seeding, structure-aware expansion, and utility-aware ranking (§\ref{sec:retrieval}). Third, downstream feedback updates both the copilot and the utility statistics of the graph, biasing future retrieval toward experiences that are empirically useful rather than merely semantically related (§\ref{sec:feedback}).

\vspace{-2.5mm}

\subsection{Experience Graph Construction and Maintenance}
\label{sec:graph_construction}

\vspace{-1.8mm}

\xhdr{Trajectory-to-experience conversion}
Let a historical trajectory be denoted as $\tau = (x, \xi, y, s)$, where $x$ is
the task input, $\xi$ denotes the intermediate execution process, $y$ is the
final response or action sequence, and $s \in \mathbb{R}$ is the task score
returned by the environment or evaluator.
Specifically, $\xi$ corresponds to the agent interaction trace in agentic environments, and to the intermediate thinking process in question-answering or reasoning tasks.
\method converts each trajectory into a compact natural-language experience unit:
\vspace{-1mm}
\begin{equation}
    e = \mathrm{Summarize}(\tau).
    \label{eq:summarize}
\end{equation}
The summarizer does not aim to preserve the full trajectory.
Instead, it extracts reusable knowledge from the trajectory.
High-scoring trajectories are distilled into \emph{skills}, such as successful reasoning patterns, planning strategies, or task-specific heuristics.
Low-scoring trajectories are distilled into \emph{lessons}, such as failure modes, invalid actions, or constraints to avoid.

Each experience unit becomes an entry $v = (e_v, h_v, u_v, n_v) \in V$ in the experience graph, where $e_v$ is the textual experience, $h_v$ is its embedding, $u_v$ is its shared-reward estimate, and $n_v$ is its retrieval count. The utility $u_v$ and count $n_v$ are updated online from downstream feedback, allowing the graph to gradually distinguish useful experiences from merely relevant ones.

\xhdr{Relational organization}
Rather than keeping entries as isolated records, \method links related experiences so that retrieval can later move from a matched entry to its neighbors.
Formally, the experience graph is a sparse undirected graph $G=(V,E)$.
When inserting a new entry $v_i$, \method connects it to semantically similar existing entries:
\vspace{-1mm}
\begin{equation}
(v_i, v_j) \in E \iff v_j \in \mathcal{N}_{K_{\text{nn}}}(v_i) \ \text{and}\ \cos(h_{v_i}, h_{v_j}) \ge \theta,
    \label{eq:edge}
\end{equation}
where $\mathcal{N}_{K_{\text{nn}}}(v_i)$ denotes the $K_{\text{nn}}$ nearest neighbors of $v_i$ under cosine similarity, and $\theta$ is a similarity threshold.
This construction keeps the graph sparse while preserving local neighborhoods of related experiences.

\vspace{-2.5mm}
\subsection{Copilot-Controlled Utility-Guided Retrieval}
\label{sec:retrieval}
\vspace{-1.8mm}

\textbf{Retrieval copilot}. \method{} trains a lightweight retrieval copilot $\pi_{\text{ret}}$ to control how the graph is searched. The copilot never produces the final answer or action; this is always done by the frozen executor $\pi_{\text{exec}}$. Instead, given a task $x$, the copilot outputs two discrete control variables:
\begin{equation}
    (R,\, W) \;\sim\; \pi_{\text{ret}}(\cdot \mid x),
    \label{eq:copilot_output}
\end{equation}
where $R \in \{0,\dots,100\}$ controls the restart/locality of graph diffusion and $W \in \{0,\dots,100\}$ controls the trade-off between semantic relevance and historical utility.
We rescale them as $\rho = R/100$ and $\lambda = W/100$, where
$\rho,\lambda \in [0,1]$.
Intuitively, $\rho$ controls how closely retrieval stays to the initial semantic seeds: a larger $\rho$ induces a higher restart probability, confining expansion to the immediate neighborhood of the seeds, while a smaller $\rho$ allows probability mass to propagate further across the graph; $\lambda$ determines how strongly the final ranking favors historically useful experiences over purely similar ones.

\xhdr{Semantic seeding}
Given task $x$ with embedding $h_x$, \method first retrieves the top-$m$ nodes by cosine similarity:
\vspace{-1mm}
\begin{equation}
    V_0 = \operatorname{Top\text{-}m}_{v \in V} \cos(h_x, h_v).
    \label{eq:semantic_seed}
\end{equation}
The seed set $V_0$ provides an initial set of task-relevant experiences.
However, these seeds are not directly used as the final retrieval result, because high semantic similarity does not necessarily imply high downstream utility.

\textbf{Structure-aware expansion}. Starting from $V_0$, \method{} expands retrieval along the relational links of the graph via graph diffusion, implemented as personalized PageRank.
The restart distribution is defined as
\vspace{-1mm}
\begin{equation}
    q(v) = \begin{cases} 1/|V_0|, & v \in V_0,\\ 0, & v \notin V_0,\end{cases}
    \label{eq:restart_distribution}
\end{equation}
and the diffusion iterates as
\vspace{-1mm}
\begin{equation}
    p_{t+1} \;=\; \alpha(\rho)\, q \;+\; \bigl(1 - \alpha(\rho)\bigr)\, A_{\mathrm{norm}}^\top\, p_t,
    \label{eq:ppr}
\end{equation}
where $A_{\mathrm{norm}}$ is the row-normalized adjacency matrix and $\alpha(\rho)\in(0,1)$ is a restart probability that increases monotonically with $\rho$.
A larger $\rho$ yields a higher restart probability, which confines probability mass closer to the semantic seeds and reduces the scope of diffusion; a smaller $\rho$ allows the distribution to spread further across the graph.
After convergence, the top-ranked nodes form the candidate set
$C=\operatorname{TopL}_{v\in V} p(v)$.

\xhdr{Utility-aware ranking}
For each candidate $v\in C$, \method estimates its utility with an upper-confidence score:
\vspace{-1mm}
\begin{equation}
    b_v = u_v + c\sqrt{\frac{\log(N_{\text{ret}}+1)}{\max(n_v,1)}},
    \label{eq:ucb}
\end{equation}
where $N_{\text{ret}}$ is the total number of retrieval events and $c>0$ is an exploration coefficient.
This score encourages the retriever to exploit experiences with high estimated utility while still exploring under-tested nodes.

The final retrieval score combines semantic relevance and utility confidence:
\vspace{-1mm}
\begin{equation}
    \mathrm{score}(v \mid x)
    \;=\;
    (1 - \lambda)\,\widehat{\mathrm{sim}}(x, v)
    \;+\;
    \lambda\,\widehat{b}_v,
    \label{eq:score}
\end{equation}
where $\widehat{\mathrm{sim}}(x,v)$ and $\widehat{b}_v$ are normalized within $C$.
The retrieved experience set is
\vspace{-1mm}
\begin{equation}
    E_{R,W}(x) = \operatorname{TopK}_{v \in C}\; \mathrm{score}(v \mid x),
    \label{eq:retrieved_set}
\end{equation}
and its textual contents are concatenated with the task input and passed to the frozen executor:
\vspace{-1mm}
\begin{equation}
    y_{\mathrm{with}} = \pi_{\mathrm{exec}}\bigl(x,\, E_{R,W}(x)\bigr).
    \label{eq:exec_with_exp}
\end{equation}

\subsection{Harness Learning from Utility Feedback}
\label{sec:feedback}

\xhdr{Utility-grounded reward}
To learn which retrieved experiences are actually useful, \method evaluates the frozen executor under two conditions.
With retrieved experiences:
\vspace{-1mm}
\begin{equation}
    s_{\mathrm{with}} = S\bigl(x,\, \pi_{\mathrm{exec}}(x,\, E_{R,W}(x))\bigr),
    \label{eq:s_with}
\end{equation}
and without any retrieved experience:
\vspace{-1mm}
\begin{equation}
    s_{\mathrm{without}} = S\bigl(x,\, \pi_{\mathrm{exec}}(x,\, \emptyset)\bigr),
    \label{eq:s_without}
\end{equation}
where $S(\cdot)$ denotes the task-specific evaluator.
The retrieval reward is
\vspace{-1mm}
\begin{equation}
    r(x, E_{R,W})
    =
    s_{\mathrm{with}}(x, E_{R,W})
    -
    s_{\mathrm{without}}(x)
    +
    \eta\, s_{\mathrm{with}}(x, E_{R,W}) .
    \label{eq:reward}
\end{equation}
The score difference $s_{\mathrm{with}}-s_{\mathrm{without}}$ records the
observed change relative to the no-experience run, while $\eta s_{\mathrm{with}}$
adds an absolute-performance term. The coefficient $\eta\geq0$ sets its weight.
The combined reward is therefore not a pure estimate of retrieval gain: when
$s_{\mathrm{with}}=s_{\mathrm{without}}>0$ and $\eta>0$, it remains positive.

\xhdr{Copilot policy optimization}
The retrieval copilot is optimized with PPO \citep{schulman2017proximal} to maximize the expected utility-grounded reward:
\vspace{-1mm}
\begin{equation}
    \max_{\pi_{\mathrm{ret}}}
    \;\mathbb{E}_{x \sim \mathcal{D}}\,
    \mathbb{E}_{(R,W) \sim \pi_{\mathrm{ret}}(\cdot \mid x)}
    \!\left[\, r(x,\, E_{R,W}) \,\right].
    \label{eq:ppo_obj}
\end{equation}
This objective trains retrieval controls using downstream task scores rather than semantic similarity alone.
The executor LLM $\pi_{\mathrm{exec}}$ is never updated.

\xhdr{Experience graph evolution during training}
The same reward also updates the experience graph, so that both halves of the harness co-evolve from a single signal.
For each retrieved entry $v \in E_{R,W}(x)$, \method updates its retrieval count and utility estimate:
\vspace{-1mm}
\begin{equation}
    n_v \leftarrow n_v + 1,
    \label{eq:count_update}
\end{equation}
\begin{equation}
    u_v \leftarrow (1 - \beta)\, u_v + \beta\, r(x,\, E_{R,W}),
    \label{eq:utility_update}
\end{equation}
where $\beta\in(0,1]$ is the update rate. Each retrieved entry receives the
same set-level reward. Thus, $u_v$ is a historical shared-reward estimate for
retrieval sets containing $v$, rather than an isolated causal contribution of
that entry; it also reflects the absolute-performance term in Eq.~\ref{eq:reward}.
In addition, the completed trajectory $\tau'=(x,\xi',y_{\mathrm{with}},s_{\mathrm{with}})$ is summarized into a new candidate experience entry, filtered for near-duplicates, and inserted into the graph using Eq.~\ref{eq:edge}.
When the graph exceeds its capacity budget, entries with low utility and low retrieval frequency are evicted.
In this way, the graph continuously absorbs new skills and lessons while suppressing experiences that are rarely useful
(see full training algorithm in Appendix~\ref{app:training}).


\section{Experiments} \label{sec:exp}


\begin{table*}[t]
\centering
\large
\setlength{\tabcolsep}{2pt}
\renewcommand{\arraystretch}{1.08}
\caption{\textbf{Performance comparison on ExpSuite-Static with 10 diverse tasks.}
Results are grouped by task category: Question Answering, Mathematical Reasoning,
and Code Generation. Weighted Avg. weights each benchmark score by its number of
test instances (Table~\ref{tab:static_data_stats}); all scores are percentages.
\textbf{Bold} and \underline{underline} denote the best and second-best results.}
\vspace{-2.5mm}
\label{tab:standardized_benchmarks}

\resizebox{\textwidth}{!}{
\begin{tabular}{
>{\centering\arraybackslash}m{1.25cm}
l
ccccc
ccc
cc
!{\vrule width 0.8pt}
c
}
\toprule

\multirow{3}{*}{\textbf{Model}} &
\multirow{3}{*}{\textbf{Method}} &
\multicolumn{5}{c}{\textbf{Question Answering}} &
\multicolumn{3}{c}{\textbf{Reasoning}} &
\multicolumn{2}{c}{\textbf{Coding}} & \\

\cmidrule(lr){3-7}
\cmidrule(lr){8-10}
\cmidrule(lr){11-12}

& &
ARC-C &
CommonsenseQA &
\shortstack{GPQA-\\Diamond} &
MMLU &
OBQA &
GSM8K &
GSM-Symbolic &
MATH &
HumanEval+ &
MBPP+ &
\textbf{Weighted Avg.} \\

\midrule


\multirow{13}{*}{
\rotatebox[origin=c]{90}{
\shortstack{
\textbf{Llama-3.2-3B-Instruct}\\
\textbf{(Small LLM)}
}}}

& No Memory
& 51.56
& 54.44
& 18.33
& 42.89
& 54.22
& 69.56
& 60.44
& 37.56
& 43.59
& 38.75
& 51.88 \\

& \multicolumn{12}{l}{
\cellcolor{gray!15}
\textit{Retrieval-Centric Experience Learning Baselines}
} \\

& ReasoningBank
& 71.78
& 61.56
& 21.67
& 54.00
& 72.67
& 70.44
& 56.44
& 45.56
& 51.28
& 57.50
& 60.83 \\

& ExpeL
& 44.67
& 57.78
& 16.67
& 46.44
& 43.11
& 77.11
& \underline{80.44}
& 23.33
& 20.51
& 22.50
& 51.49 \\

& LightMem
& 71.56
& \textbf{64.00}
& \underline{26.67}
& \underline{58.00}
& \underline{72.89}
& 44.89
& 42.00
& 49.11
& 48.72
& 42.50
& 56.47 \\

& Mem0
& 68.00
& 59.56
& \textbf{28.33}
& 56.44
& 65.78
& 52.00
& 39.78
& 28.22
& 41.03
& \underline{60.00}
& 52.42 \\

& AWM
& \underline{72.44}
& 60.44
& 15.00
& 53.33
& 69.33
& 55.33
& 38.89
& 45.78
& \underline{53.85}
& \underline{60.00}
& 55.81 \\

& MemRL
& 67.33
& 60.00
& 23.33
& 40.22
& 67.11
& \underline{83.78}
& 68.00
& \underline{55.33}
& 46.15
& 57.50
& \underline{62.06} \\

& \multicolumn{12}{l}{
\cellcolor{gray!15}
\textit{LLM-Centric Experience Learning Baselines}
} \\

& IRCoT
& 62.67
& 59.78
& 15.00
& 38.00
& 63.33
& 77.78
& 72.67
& 29.56
& 51.28
& 50.00
& 56.65 \\

& Search-o1
& 66.67
& 60.22
& \textbf{28.33}
& 50.00
& 62.44
& 78.44
& 73.33
& 38.44
& 30.77
& 28.75
& 59.63 \\

& S3
& 64.89
& 59.11
& 23.33
& 44.00
& 64.00
& 82.00
& 74.00
& 53.11
& 48.72
& 55.00
& 61.94 \\

& \cellcolor{hl-llama}\textbf{\method}
& \cellcolor{hl-llama}\textbf{74.00}
& \cellcolor{hl-llama}\underline{63.11}
& \cellcolor{hl-llama}\underline{26.67}
& \cellcolor{hl-llama}\textbf{60.00}
& \cellcolor{hl-llama}\textbf{74.00}
& \cellcolor{hl-llama}\textbf{84.89}
& \cellcolor{hl-llama}\textbf{82.00}
& \cellcolor{hl-llama}\textbf{57.11}
& \cellcolor{hl-llama}\textbf{56.41}
& \cellcolor{hl-llama}\textbf{62.50}
& \cellcolor{hl-llama}\textbf{69.57} \\

\midrule


\multirow{13}{*}{
\rotatebox[origin=c]{90}{
\shortstack{
\textbf{Llama-3.1-8B-Instruct}\\
\textbf{(Large LLM)}
}}}

& No Memory
& 70.22
& 63.11
& 23.33
& 53.11
& 69.78
& 70.89
& 62.67
& 39.78
& 43.59
& 58.75
& 60.41 \\

& \multicolumn{12}{l}{
\cellcolor{gray!15}
\textit{Retrieval-Centric Experience Learning Baselines}
} \\

& ReasoningBank
& 82.22
& 70.44
& \underline{28.33}
& 63.78
& 78.44
& 80.67
& 74.22
& 47.56
& 58.97
& 62.50
& 69.93 \\

& ExpeL
& 82.67
& \textbf{75.78}
& 21.67
& 66.67
& 82.67
& 91.11
& \underline{88.89}
& 45.78
& 35.90
& 66.25
& 74.53 \\

& LightMem
& 81.56
& 72.22
& 23.33
& 60.89
& 80.89
& 81.56
& 72.00
& 49.11
& 51.28
& 68.75
& 70.02 \\

& Mem0
& 83.33
& 70.00
& 20.00
& 67.11
& 80.67
& 90.67
& 86.00
& 47.78
& \underline{64.10}
& \underline{77.50}
& 74.02 \\

& AWM
& 81.56
& 74.00
& 18.33
& 60.22
& 78.67
& 81.78
& 76.44
& 47.11
& \underline{64.10}
& 76.25
& 70.47 \\

& MemRL
& \underline{84.44}
& 73.78
& 23.33
& \underline{67.56}
& \underline{84.00}
& \underline{93.33}
& 80.00
& \underline{56.44}
& 35.90
& 65.00
& \underline{75.34} \\

& \multicolumn{12}{l}{
\cellcolor{gray!15}
\textit{LLM-Centric Experience Learning Baselines}
} \\

& IRCoT
& 83.11
& \underline{75.33}
& \textbf{30.00}
& 67.11
& 82.22
& 88.67
& 84.00
& 52.00
& 48.72
& 71.25
& 74.80 \\

& Search-o1
& 81.33
& 71.78
& \underline{28.33}
& 64.44
& 78.00
& 89.33
& 84.67
& 48.00
& 46.15
& 63.75
& 72.54 \\

& S3
& 82.00
& 73.56
& 26.67
& 65.56
& 81.56
& 90.89
& 86.00
& 52.89
& 58.97
& 72.50
& 74.89 \\

& \cellcolor{hl-qwen}\textbf{\method}
& \cellcolor{hl-qwen}\textbf{86.00}
& \cellcolor{hl-qwen}75.11
& \cellcolor{hl-qwen}\underline{28.33}
& \cellcolor{hl-qwen}\textbf{68.89}
& \cellcolor{hl-qwen}\textbf{86.00}
& \cellcolor{hl-qwen}\textbf{95.11}
& \cellcolor{hl-qwen}\textbf{90.00}
& \cellcolor{hl-qwen}\textbf{58.00}
& \cellcolor{hl-qwen}\textbf{66.67}
& \cellcolor{hl-qwen}\textbf{78.75}
& \cellcolor{hl-qwen}\textbf{78.76} \\

\bottomrule
\end{tabular}
}
\end{table*}

\begin{table*}[t]
\centering
\caption{\textbf{Performance comparison on ExpSuite-Agentic across ALFWorld and AppWorld tasks.}
We report success rate (SR) for ALFWorld, pass rate (PR) for AppWorld, and the weighted average score across all test tasks (weighted by the number of test tasks in each split; Table~\ref{tab:agentic_data_stats}); all SR, PR, and Score values are percentages.
\textbf{Bold} and \underline{underline} denote the best and second-best results.}
\label{tab:expsuite_agentic}
\setlength{\tabcolsep}{3.2pt}
\resizebox{\textwidth}{!}{
\begin{tabular}{l cccccccccc !{\vrule width 0.8pt} cccccccccc}
\toprule
& \multicolumn{10}{c}{\textbf{Qwen3-32B (Small LLM)}} &
  \multicolumn{10}{c}{\textbf{Gemini-3.1-Flash-Lite (Large LLM)}} \\
\cmidrule(lr){2-11} \cmidrule(lr){12-21}
& \multicolumn{4}{c}{\textbf{ALFWorld}}
& \multicolumn{4}{c}{\textbf{AppWorld}}
& \multicolumn{2}{c}{\textbf{Avg.}}
& \multicolumn{4}{c}{\textbf{ALFWorld}}
& \multicolumn{4}{c}{\textbf{AppWorld}}
& \multicolumn{2}{c}{\textbf{Avg.}} \\
\cmidrule(lr){2-5} \cmidrule(lr){6-9} \cmidrule(lr){10-11}
\cmidrule(lr){12-15} \cmidrule(lr){16-19} \cmidrule(lr){20-21}
& \multicolumn{2}{c}{\textbf{ALF-Seen}}
& \multicolumn{2}{c}{\textbf{ALF-Unseen}}
& \multicolumn{2}{c}{\textbf{Test-N}}
& \multicolumn{2}{c}{\textbf{Test-C}}
& \multirow{2}{*}{\textbf{Score}}
& \multirow{2}{*}{\textbf{\#Steps$\downarrow$}}
& \multicolumn{2}{c}{\textbf{ALF-Seen}}
& \multicolumn{2}{c}{\textbf{ALF-Unseen}}
& \multicolumn{2}{c}{\textbf{Test-N}}
& \multicolumn{2}{c}{\textbf{Test-C}}
& \multirow{2}{*}{\textbf{Score}}
& \multirow{2}{*}{\textbf{\#Steps$\downarrow$}} \\
\cmidrule(lr){2-3} \cmidrule(lr){4-5} \cmidrule(lr){6-7} \cmidrule(lr){8-9}
\cmidrule(lr){12-13} \cmidrule(lr){14-15} \cmidrule(lr){16-17} \cmidrule(lr){18-19}
\textbf{Method}
& \textbf{SR} & \textbf{\#Steps$\downarrow$}
& \textbf{SR} & \textbf{\#Steps$\downarrow$}
& \textbf{PR} & \textbf{\#Steps$\downarrow$}
& \textbf{PR} & \textbf{\#Steps$\downarrow$}
& &
& \textbf{SR} & \textbf{\#Steps$\downarrow$}
& \textbf{SR} & \textbf{\#Steps$\downarrow$}
& \textbf{PR} & \textbf{\#Steps$\downarrow$}
& \textbf{PR} & \textbf{\#Steps$\downarrow$}
& & \\
\midrule

No-Memory
& 19.3 & 41.2 & 35.8 & 42.5 & 30.1 & 30.8 & 21.6 & 31.5 & 25.1 & 34.7
& 62.7 & 27.9 & 64.9 & 26.3 & 34.2 & 33.0 & 23.2 & 36.6 & 38.3 & 32.9 \\

\rowcolor{gray!15}
\multicolumn{21}{l}{\textit{Prompt-based Agentic Baselines}} \\
ReAct
& 31.4 & 36.7 & \underline{47.0} & 38.6 & 30.0 & 32.2 & 21.8 & 31.9 & 28.9 & 33.8
& 65.7 & 24.1 & 63.4 & 25.1 & 32.7 & 24.8 & 32.1 & 16.1 & 42.6 & 20.5 \\
Reflexion
& 23.6 & 40.6 & 40.3 & 38.7 & 31.6 & 31.3 & 21.9 & 32.7 & 26.9 & 34.6
& 63.1 & 23.3 & 60.4 & 23.7 & 48.2 & 21.9 & 31.1 & \underline{15.6} & 44.2 & \underline{19.4} \\

\rowcolor{gray!15}
\multicolumn{21}{l}{\textit{Retrieval-Centric Experience Learning Baselines}} \\
ReasoningBank
& 37.9 & 35.3 & 26.1 & 42.6 & 29.4 & 25.6 & 22.2 & 24.3 & 26.8 & 29.2
& 67.2 & 29.1 & 81.3 & 24.1 & 45.3 & 23.1 & \underline{41.2} & 25.5 & 52.5 & 25.4 \\
ExpeL
& 54.3 & 27.5 & 29.1 & 41.4 & 26.7 & 35.1 & 28.2 & 25.6 & 32.3 & 30.2
& 71.6 & \underline{21.4} & 63.4 & 28.7 & 43.6 & 21.1 & 29.9 & 28.3 & 44.6 & 25.8 \\
LightMem
& 49.2 & 29.6 & 25.0 & 39.2 & 28.5 & 34.2 & 23.7 & 37.5 & 29.0 & 35.8
& 70.1 & 27.8 & 70.9 & 27.8 & 34.2 & 31.1 & 29.7 & 34.2 & 43.6 & 31.6 \\
Mem0
& 30.0 & 37.3 & 18.7 & 42.8 & 20.8 & 33.4 & 15.7 & 35.2 & 19.5 & 36.4
& 58.2 & 30.6 & 63.4 & 22.4 & 32.7 & 31.5 & 22.9 & 34.8 & 36.9 & 31.5 \\
AWM
& 37.9 & 35.3 & 32.8 & 40.1 & 28.0 & 28.8 & 22.8 & 33.0 & 27.8 & 33.7
& 73.1 & 24.7 & 80.6 & \underline{18.8} & 50.7 & 20.9 & 31.5 & 25.8 & 49.7 & 23.6 \\
MemRL
& 25.7 & 40.3 & 20.2 & 49.6 & 45.2 & 18.6 & 28.9 & 24.7 & 30.2 & 29.9
& 61.9 & 28.9 & 71.6 & 26.1 & 40.5 & \underline{13.5} & 31.7 & 16.7 & 44.6 & 19.5 \\

\rowcolor{gray!15}
\multicolumn{21}{l}{\textit{LLM-Centric Experience Learning Baselines}} \\
IRCoT
& 33.6 & 37.9 & 27.6 & 40.6 & 52.8 & 11.4 & \underline{36.1} & 17.8 & 37.6 & 23.4
& 75.4 & 27.0 & \underline{87.3} & 19.6 & 46.9 & 18.7 & 34.9 & 25.4 & 52.0 & 23.4 \\
Search-o1
& 32.1 & 38.0 & 11.2 & 47.3 & 45.4 & 16.5 & 26.5 & \underline{14.2} & 28.7 & 23.7
& 76.1 & 22.4 & 79.4 & 23.5 & 52.3 & 20.7 & 39.8 & 24.6 & 54.3 & 23.3 \\
S3
& \underline{58.6} & \underline{23.5}
& 42.5 & \underline{22.5}
& \underline{55.4} & \underline{9.7}
& 35.0 & 15.0
& \underline{44.0} & \underline{16.5}
& \underline{77.1} & 21.8
& 80.6 & 21.5
& \underline{53.0} & 18.5
& 40.8 & 22.0
& \underline{55.3} & 21.1 \\

\midrule
\rowcolor{blue!8}
\method
& \textbf{70.0} & \textbf{20.0}
& \textbf{75.4} & \textbf{18.0}
& \textbf{57.1} & \textbf{8.8}
& \textbf{39.3} & \textbf{13.6}
& \textbf{53.4} & \textbf{14.4}
& \textbf{85.0} & \textbf{17.6}
& \textbf{88.1} & \textbf{17.0}
& \textbf{57.1} & \textbf{12.8}
& \textbf{48.4} & \textbf{14.8}
& \textbf{62.3} & \textbf{15.2} \\

\bottomrule
\end{tabular}
}
\vspace{-2mm}
\end{table*}

We evaluate \method on ExpSuite, comprising ExpSuite-Static for single-turn reasoning and generation tasks and ExpSuite-Agentic for multi-step interactive decision-making. Across both settings, the executor LLM is kept frozen, and all methods are evaluated without manually curated few-shot examples. We compare against prompting baselines, retrieval-centric experience learning methods, and LLM-centric experience learning methods. Dataset statistics, implementation details, and extended analysis are provided in Appendix~\ref{app:dataset_statistics}, \ref{app:hyper}, and~\ref{app:exp}.

\xhdr{Task description}
ExpSuite consists of two settings. \textbf{(i) ExpSuite-Static} includes ten benchmarks across three categories: question answering (ARC-C~\citep{arc}, CommonsenseQA~\citep{CommonsenseQA}, GPQA-Diamond~\citep{rein2024gpqa}, MMLU~\citep{MMLU}, OBQA~\citep{OpenbookQA}), mathematical reasoning (GSM8K~\citep{gsm8k}, GSM-Symbolic~\citep{gsm_sym}, MATH~\citep{math}), and code generation (HumanEval+~\citep{liu2023your}, MBPP+~\citep{liu2023your}). We report accuracy, exact match, and Pass@1~\citep{chen2021evaluating} respectively. \textbf{(ii) ExpSuite-Agentic} includes ALFWorld~\citep{shridhar2020alfworld} (Seen/Unseen splits; SR and \#Steps) and AppWorld~\citep{trivedi2024appworld} (Test-Normal/Test-Challenge splits; PR and \#Steps). For executor models, we use Llama-3.2-3B-Instruct (Small LLM) and Llama-3.1-8B-Instruct
(Large LLM)~\citep{grattafiori2024llama} for ExpSuite-Static, and Qwen3-32B (Small
LLM)~\citep{yang2025qwen3} and Gemini-3.1-Flash-Lite (Large
LLM)~\citep{google2026gemini31flashlite} for ExpSuite-Agentic.

\xhdr{Baselines}
We compare \method with four groups of baselines. \textbf{(i) No-Memory} directly uses the frozen executor without external experiences. \textbf{(ii) Retrieval-centric} baselines maintain external experience records, including ReasoningBank~\citep{ouyang2025reasoningbank}, ExpeL~\citep{zhao2024expel}, LightMem~\citep{fang2026lightmem}, Mem0~\citep{chhikara2025mem0}, AWM~\citep{wang2024agent}, and MemRL~\citep{zhang2026memrl}. \textbf{(iii) LLM-centric} baselines use an LLM-based policy to retrieve or exploit external information, including IRCoT~\citep{trivedi2023interleaving}, Search-o1~\citep{li2025search}, and S3~\citep{jiang2025s3}. \textbf{(iv) Prompt-based agentic} baselines (ExpSuite-Agentic only) include ReAct~\citep{yao2022react} and Reflexion~\citep{shinn2023reflexion}. For retrieval-centric baselines, the memory-construction protocol provides
the same underlying historical trajectories wherever applicable
(Appendix~\ref{app:hyper}). The resulting numbers and lengths of memory items
can differ across representations.

\vspace{-2.5mm}

\subsection{\method Outperforms General Prompt-based Baselines and Experience Learning Methods}
\label{sec:4.1}

\vspace{-1.8mm}

We evaluate \method on ExpSuite, covering both single-turn static tasks and multi-step agentic environments.
Results are reported in Table~\ref{tab:standardized_benchmarks} and Table~\ref{tab:expsuite_agentic}.
We have the following observations.

\xhdr{\method Achieves the Highest Weighted Average Scores in the Evaluated Settings}
On ExpSuite-Static, \method improves the test-instance-weighted average score over
MemRL, the strongest baseline under this aggregation, by 12.1\% with the small
executor and 4.5\% with the large executor. These are relative improvements,
computed as $(S_{\mathrm{ours}}-S_{\mathrm{baseline}})/S_{\mathrm{baseline}}$;
they do not imply the best result on every individual benchmark.
On ExpSuite-Agentic, the corresponding relative improvements over S3 are 21.4\%
and 12.7\%. Average environment interaction steps are 12.7\% and 21.6\% lower
than those of the baselines with the lowest average step counts, respectively.
Step counts describe environment interaction and do not include all retrieval,
LLM inference, or training costs; they can also vary with task success.

\xhdr{Relative Gains Vary Across Executors and Task Settings}
In these evaluations, the relative gains are larger for the executor labeled
small within each setting, and larger in ExpSuite-Agentic than in ExpSuite-Static.
These comparisons are descriptive: executor identities, benchmarks, and baseline
scores differ across settings. They are consistent with useful experience reuse
in interactive tasks, but do not isolate model scale or task horizon as the
cause of the differences.

\captionsetup[subfigure]{justification=raggedright}
\begin{figure*}[t]
    \vspace{-3mm}
    \centering
    \begin{subfigure}{0.32\textwidth}
        \centering
        \includegraphics[height=4.5cm]{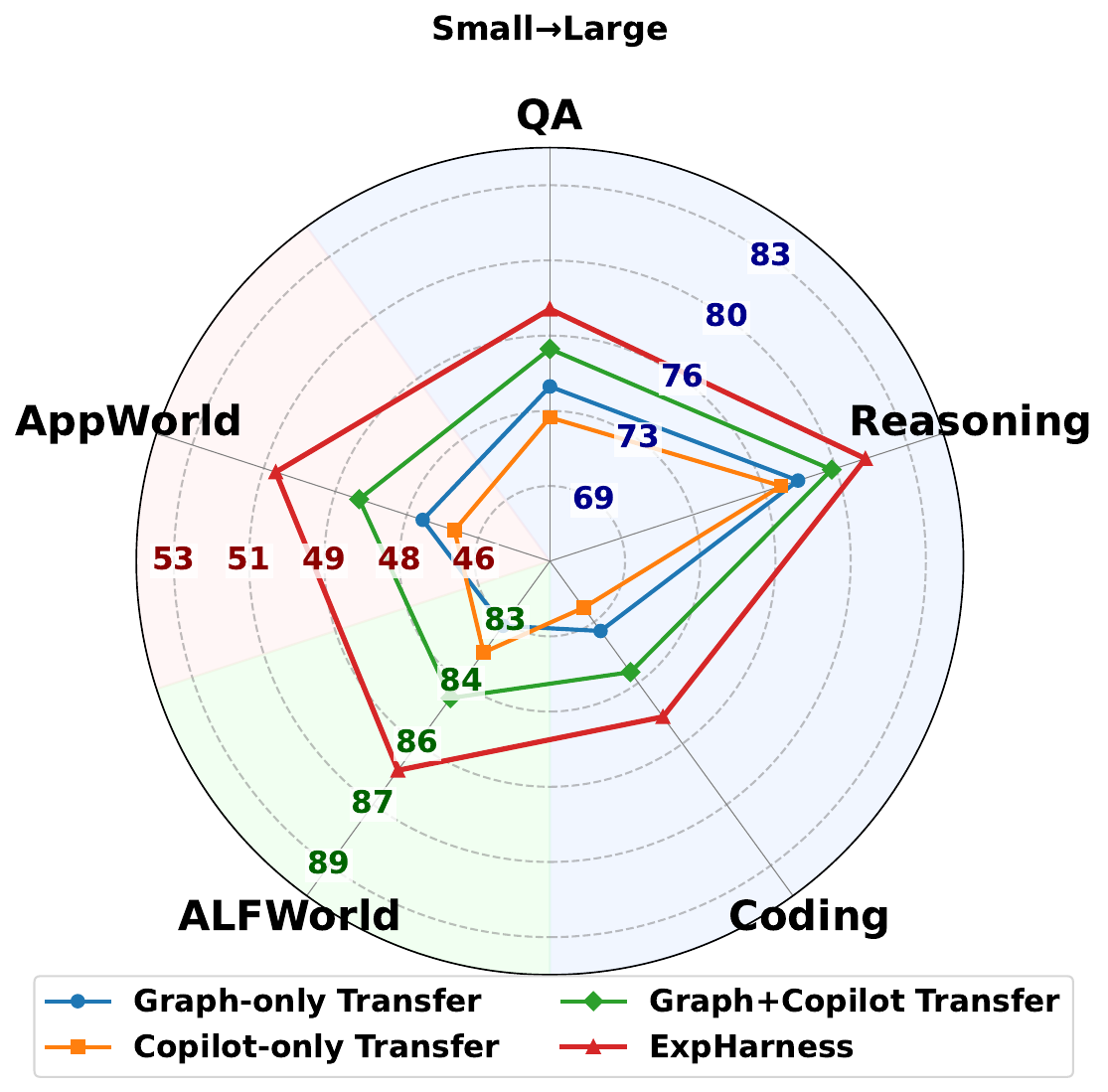}
        \vspace{-5mm}
        \caption*{\hspace{0.1cm}(a)}
        \label{fig:transfer_small_to_large}
    \end{subfigure}
    \begin{subfigure}{0.32\textwidth}
        \centering
        \includegraphics[height=4.5cm]{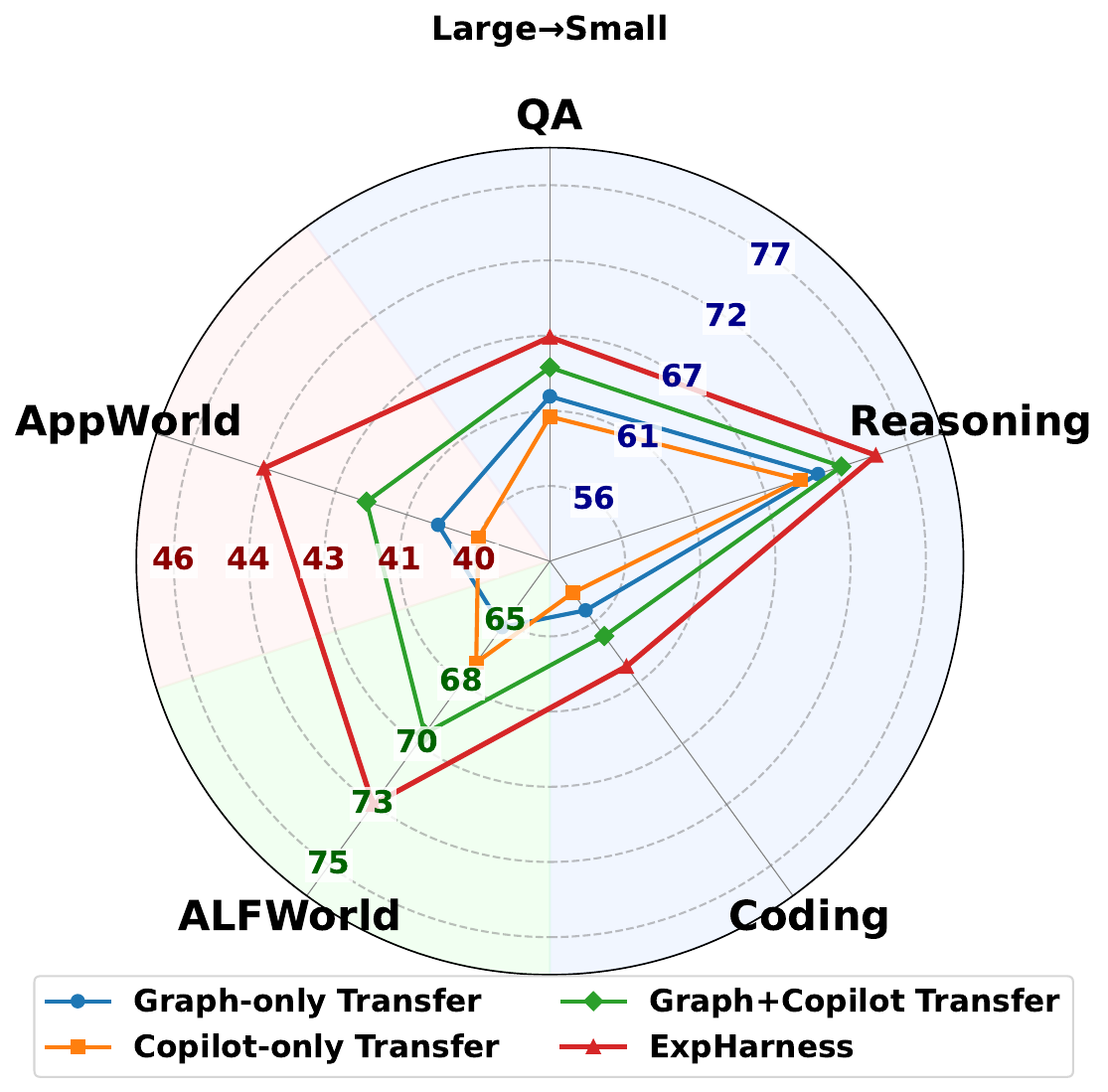}
        \vspace{-5mm}
        \caption*{\hspace{0.2cm}(b)}
        \label{fig:transfer_large_to_small}
    \end{subfigure}
    \begin{subfigure}{0.32\textwidth}
        \centering
        \includegraphics[height=4.5cm]{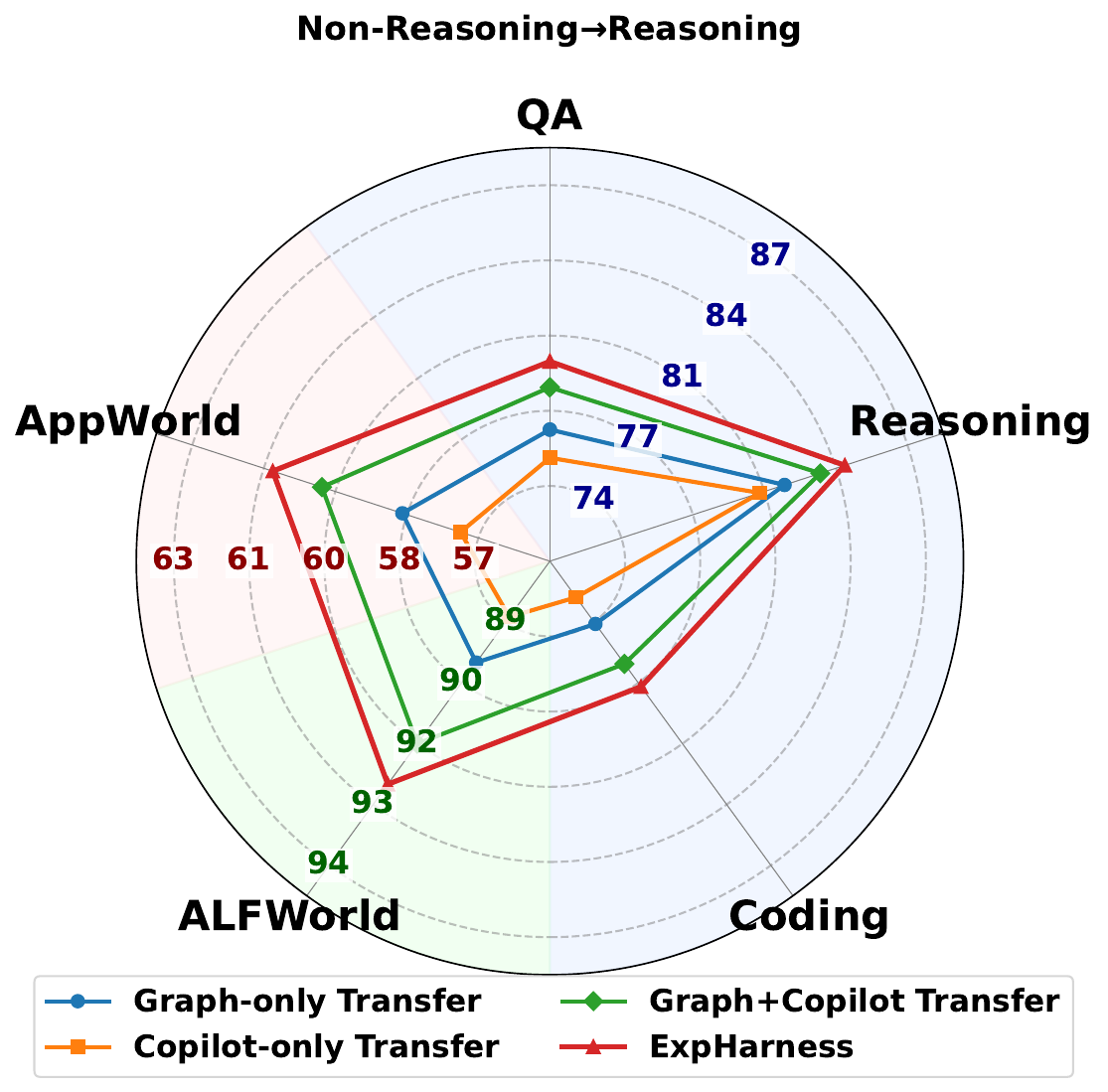}
        \vspace{-5mm}
        \caption*{\hspace{0.3cm}(c)}
        \label{fig:transfer_non_reasoning_to_reasoning}
    \end{subfigure}
    \vspace{-3mm}
 \caption{\textbf{Harness transfer across different executor shifts.}
(a) \textit{Small-to-large transfer}: transferring the learned harness (experience graph and retrieval copilot) from a smaller executor to a larger executor.
(b) \textit{Large-to-small transfer}: transferring harness components from a larger executor to a smaller executor.
(c) \textit{Non-reasoning-to-reasoning transfer}: transferring harness components across executors with different reasoning capabilities.
}
    \label{fig:executor_transfer_radar}
    \vspace{-15pt}
\end{figure*}

\vspace{-2.5mm}

\subsection{The Learned Harness Transfers Across Different LLM Executors}
\label{sec:executor_generalization}

\vspace{-1.8mm}

Because all learning in \method resides in the harness, a crucial question is whether a harness learned with one executor remains useful when the executor is replaced. To investigate this, we compare harness transfer under three settings.
\textit{Small-to-large transfer} uses the small executor as source and the large executor as target, testing whether experiences from cheaper models benefit stronger frozen LLMs. 
\textit{Large-to-small transfer} reverses this direction, testing whether high-quality experiences from stronger models remain useful for weaker executors. 
\textit{Non-reasoning-to-reasoning transfer} uses non-reasoning executors as source and reasoning-capable executors as target, testing whether experiences transfer across reasoning capability gaps. Specifically, for ExpSuite-Static, we transfer from Llama-3.1-8B-Instruct to DeepSeek-R1-Distill-Llama-8B \citep{guo2025deepseek}; for ExpSuite-Agentic, we transfer from Gemini-3.1-Flash-Lite to Claude-Sonnet-4 \citep{anthropic2025claude4}.
For each setting, we compare Graph-only Transfer (experience graph only), Copilot-only Transfer (copilot only), Graph+Copilot Transfer (full harness), and target-specific \method trained on the target executor. The comparisons below concern performance relative to these variants and the target-trained reference.

\xhdr{Joint Transfer Is Closest to Target-Specific Training in the Small-to-Large Setting}
In Figure~\ref{fig:executor_transfer_radar}(a), Graph+Copilot Transfer is closest
to target-specific \method across the displayed domains. It also outperforms
the two partial-transfer variants, supporting joint reuse of the experience
store and retrieval policy in this setting.

\xhdr{Large-to-Small Transfer Retains a Gap to Target-Specific Training}
In Figure~\ref{fig:executor_transfer_radar}(b), the transferred variants remain
below target-specific \method, while Graph+Copilot Transfer performs best among
the transfer variants across most displayed domains. The panels use different
axis ranges, so their polygon areas cannot quantify the relative difficulty of
the transfer directions. Without a target No-Memory reference in this figure,
these comparisons alone do not establish improvement over the unaugmented
target executor.

\xhdr{Joint Transfer Also Performs Well for Reasoning-Oriented Targets}
In Figure~\ref{fig:executor_transfer_radar}(c), Graph+Copilot Transfer approaches
target-specific \method, including on ALFWorld and AppWorld. This observation
is consistent with the reuse of experience across the evaluated model pairs.
Because model identity and reasoning capability change together, the experiment
does not isolate reasoning capability as the cause of transfer performance.

\begin{figure}[t]
    \centering
    \includegraphics[width=1\linewidth]{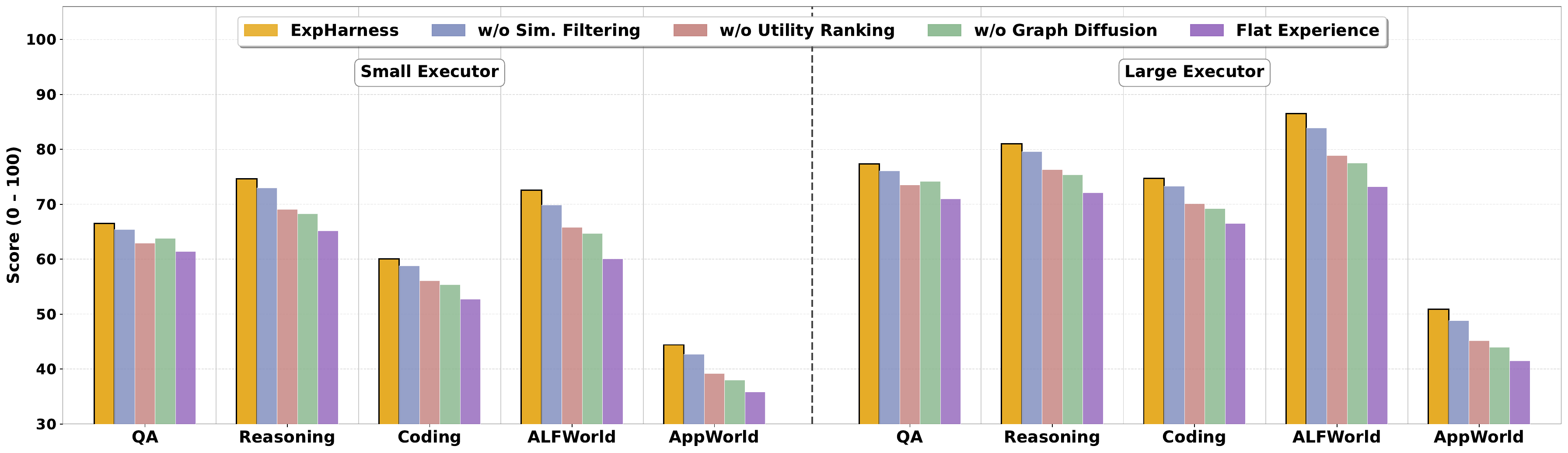}
    \vspace{-7mm}
    \caption{\textbf{Performance under four ablation settings.}
We compare \method with variants modifying similarity filtering, memory
organization, graph diffusion, and utility-aware ranking across QA, Reasoning,
Coding, ALFWorld, and AppWorld.
}
    \vspace{-5mm}
    \label{fig:ablation_studies}
\end{figure}

\vspace{-2.5mm}

\subsection{Ablation Studies of Four Experience-Retrieval Design Choices}
\label{sec:ablation}

\vspace{-1.8mm}

To examine four experience-retrieval design choices, we conduct ablation studies across five evaluation domains: QA, Reasoning, Coding, ALFWorld, and AppWorld, with results reported in Figure~\ref{fig:ablation_studies}.
We compare \method with four variants: i) \textbf{w/o Similarity Filtering} removes similarity-based filtering during experience insertion, allowing redundant experiences to be repeatedly retained; ii) \textbf{Flat Experience} replaces the experience graph with a flat experience pool and retrieves top-$K$ experiences purely by semantic similarity; iii) \textbf{w/o Graph Diffusion} keeps the relational links but disables structure-aware expansion, retrieving only from the initial semantic seeds; and iv) \textbf{w/o Utility Ranking} ranks candidate experiences only by semantic similarity, without using historical utility statistics.

Across the reported domains, each of these four variants performs below the
full method. \textbf{Flat Experience} shows the largest drop, but it changes both
candidate generation and the ranking rule, so its gap cannot be attributed to
relational structure alone. \textbf{w/o Graph Diffusion} and \textbf{w/o Utility
Ranking} show lower scores under their respective retrieval changes, while
\textbf{w/o Similarity Filtering} shows a smaller decline. These results support
the combined retrieval design under the tested configurations. They do not
separately isolate the effects of copilot training, the reward terms, or the
node-utility update rule.

\section{Additional Related Work}


Reinforcement learning has become an important mechanism for improving LLMs and LLM agents, from PPO-based alignment~\citep{ouyang2022training,liu2024deepseek} to DPO~\citep{rafailov2023direct}, process reward models~\citep{lightman2024let}, self-play~\citep{chen2024self}, self-correction~\citep{kumar2024training}, and recent reasoning-oriented variants such as GRPO~\citep{shao2024deepseekmath}, Dr.GRPO~\citep{liu2025understanding}, GSPO~\citep{zheng2025group}, and Clip-Cov~\citep{cui2025entropy}.
Beyond directly optimizing LLM policies, experience-learning systems store and reuse prior interactions: ReAct~\citep{yao2022react} and Reflexion~\citep{shinn2023reflexion} use reasoning traces and verbal feedback, while ExpeL~\citep{zhao2024expel}, ReasoningBank~\citep{ouyang2025reasoningbank}, LightMem~\citep{fang2026lightmem}, Mem0~\citep{chhikara2025mem0}, and AWM~\citep{wang2024agent} construct external memories from past trajectories; IRCoT~\citep{trivedi2023interleaving}, Search-o1~\citep{li2025search}, S3~\citep{jiang2025s3}, and MemRL~\citep{zhang2026memrl} further introduce learned retrieval or episodic-memory policies.
These methods differ in memory organization, retrieval control, and the use
of downstream feedback. \method studies their combination in an external
experience harness comprising an experience graph and a trainable retrieval
copilot, jointly updated using outcome-based feedback around a frozen executor. Its transfer
experiments further examine reuse of this harness across executor models.

\section{Conclusion}
We introduce \textsc{ExpHarness}, a learnable experience harness that improves
frozen LLM executors through external experience reuse rather than parameter
updates. To select experiences that are useful rather than merely similar, we
organize them in a self-evolving experience graph and train a retrieval copilot
with utility-grounded feedback, fostering a harness that adapts to each task
and transfers across executors. Transfer experiments support reuse of the learned harness across the evaluated executor pairs, with joint graph and copilot transfer closest to target-specific training among the compared variants. 
\section*{AI use disclosure}
We used generative AI tools solely to polish the writing of this paper. The authors reviewed and edited all AI-assisted text and take full responsibility for the content.

\bibliography{iclr2027_conference}
\bibliographystyle{iclr2027_conference}

\appendix
\section{Limitations, Future Work, and Broader Impact}
\label{app:limit_future_impact}

\paragraph{Limitations.}
\label{app:limitations}
Although \method provides a model-agnostic way to improve frozen LLM agents through graph-structured experience reuse, our study has several limitations. 
First, our experiments focus on a representative set of static reasoning benchmarks, including question answering~\citep{arc, CommonsenseQA, rein2024gpqa, MMLU, OpenbookQA}, mathematical reasoning~\citep{gsm8k, gsm_sym, math}, code generation~\citep{liu2023your}, and agentic environments~\citep{shridhar2020alfworld, trivedi2024appworld}. While these tasks cover diverse forms of experience reuse, future work could further evaluate \method on longer-horizon real-world applications, such as web browsing~\citep{he2024webvoyager, pan2024webcanvas, wei2025browsecompsimplechallengingbenchmark}, scientific discovery~\citep{jansen2024discoveryworld, yuksekgonul2024textgrad, tang2025chemagent, xu2026zero}, and collaborative multi-agent workflows~\citep{jiang2025adaptation, park2026choicemates}. Second, the experience graph is constructed using embedding-based semantic similarity with fixed 
hyperparameters, including neighbor size, similarity threshold, graph capacity, and utility update rate. Although these choices work well empirically, more adaptive graph construction and pruning strategies may further improve 
robustness across domains with different memory distributions.

\paragraph{Future Work.}
\label{app:future_works}
Several directions remain open for future investigation.
First, the experience graph is currently constructed using embedding-based semantic similarity with fixed hyperparameters; replacing these with adaptive graph construction and pruning strategies could improve robustness across diverse domains.
Second, the retrieval copilot is trained with scalar utility-grounded rewards from the downstream task outcomes, richer feedback signals, such as process-level credit from intermediate task milestones~\citep{ma2023let, khalifa2025process}, could sharpen the learning signal further. Third, the agent currently leverages retrieved experiences through prompt-based conditioning. However, prompt-level instructions are known to translate inconsistently into the model's downstream behavior~\citep{han2025personalityillusionrevealingdissociation, han2026steer2adapt, turpin2023language}, leaving room for mechanisms that more tightly couple experience utilization with the policy itself, such as fine-tuning on experience-grounded trajectories~\citep{lu2025onpolicydistillation}, distilling experience-conditioned reasoning into the model's parameters~\citep{ye2026onpolicycontextdistillationlanguage, zhao2026self}.

\paragraph{Broader Impact.}
\label{app:broader_impact}
\method enables LLM agents to accumulate and transfer task knowledge through an 
external graph-structured memory, without modifying model parameters. This design avoids executor parameter updates, although training the copilot
and constructing memory incur additional costs. It also permits the use of
closed-source executors through their input-output interfaces. Keeping their
parameters frozen does not guarantee the safety of behavior conditioned on
retrieved experience. We encourage future deployments of \method to be
accompanied by robust evaluation protocols, human-in-the-loop checkpoints, and 
auditable retrieval logs.

\section{Implementation Details}
\label{app:hyper}

We implement \method with two main components: a lightweight retrieval copilot $\pi_{\mathrm{ret}}$ and a self-evolving experience graph $G=(V,E)$. The executor LLM $\pi_{\mathrm{exec}}$ is kept frozen throughout all experiments, and only the retrieval copilot and graph memory are updated during training.

\paragraph{Retrieval copilot.}
The retrieval copilot $\pi_{\mathrm{ret}}$ is instantiated with Qwen2.5-3B-Instruct\footnote{\url{https://huggingface.co/Qwen/Qwen2.5-3B-Instruct}}.
Given a task $x$, the copilot outputs two discrete control variables $(R,W) \sim \pi_{\mathrm{ret}}(\cdot \mid x)$, where $R$ controls the restart/locality of graph diffusion (larger $R$ keeps retrieval closer to the semantic seeds) and $W$ controls the trade-off between semantic relevance and historical utility.
We rescale them as $\rho=R/100$ and $\lambda=W/100$.
The copilot is optimized with Proximal Policy Optimization (PPO)~\citep{schulman2017proximal} using the utility-grounded reward in Eq.~\ref{eq:reward}, with $\eta=1$.
We use the verl distributed RL framework with Fully Sharded Data Parallel (FSDP) for parameter and gradient offloading.
The actor learning rate is $5\times10^{-6}$ and the critic learning rate is $1\times10^{-5}$, both using AdamW~\citep{loshchilov2017decoupled} with cosine learning-rate decay and no warmup.
The KL penalty coefficient is set to $0.01$, and gradient checkpointing is enabled to reduce memory usage.
The maximum prompt length is 4096 tokens and the maximum response length is 500 tokens.
During training, we sample copilot outputs with temperature $1.0$; during evaluation, we use greedy decoding.
We train with a batch size of 16 for up to 500 gradient steps across 2 epochs.
Checkpoints are saved every 20 steps and validation is performed every 30 steps.

\paragraph{Experience graph.}
The experience graph stores each experience as a node $v=(e_v,h_v,u_v,n_v)$, where $e_v$ is the natural-language experience, $h_v$ is its text embedding, $u_v$ is its shared-reward estimate, and $n_v$ is its retrieval count.
We use Contriever\footnote{\url{https://huggingface.co/facebook/contriever}} as the embedding model for both task inputs and experience nodes.
When a new node is inserted into the graph, it is connected to its top-$K_{\mathrm{nn}}=5$ nearest neighbors under cosine similarity if the similarity is above the threshold $\theta=0.3$, following Eq.~\ref{eq:edge}.
The graph capacity is capped at $|V|_{\max}=2{,}000$ nodes.
When the capacity is exceeded, we evict nodes with low utility and low retrieval frequency.
For semantic seeding, we retrieve the top-$m=10$ nodes by cosine similarity as the seed set $V_0$ in Eq.~\ref{eq:semantic_seed}.
After personalized PageRank expansion in Eq.~\ref{eq:ppr}, we construct the candidate set and apply utility-aware ranking using Eqs.~\ref{eq:ucb}--\ref{eq:score}.
The final retrieved set contains the top-$K=10$ experiences, as defined in Eq.~\ref{eq:retrieved_set}.
The UCB exploration coefficient is set to $c=1.0$.
For online graph updates, we update the retrieval count and utility estimate according to Eqs.~\ref{eq:count_update}--\ref{eq:utility_update}, with exponential moving average rate $\beta=0.1$.

\paragraph{Experience summarization.}
Each completed trajectory $\tau'=(x,\xi',y_{\mathrm{with}},s_{\mathrm{with}})$ is summarized into a compact experience unit $e'=\mathrm{Summarize}(\tau')$ following Eq.~\ref{eq:summarize}.
The summarizer extracts reusable information rather than preserving the full trajectory.
High-scoring trajectories are summarized into skills, such as successful reasoning patterns, planning strategies, or task-specific heuristics.
Low-scoring trajectories are summarized into lessons, such as failure modes, invalid actions, or constraints to avoid.
For \textsc{ExpSuite-Static}, we use Qwen2.5-3B-Instruct in zero-shot mode as the summarizer.
For \textsc{ExpSuite-Agentic}, we use Gemini-3.1-Flash-Lite to generate environment-specific experience descriptions.
Detailed prompts are provided in Appendix~\ref{app:prompt}.

\paragraph{Cold-start initialization.}
Before RL training, we initialize the experience graph with a cold-start procedure.
We sample a subset of training tasks, execute them with the frozen executor $\pi_{\mathrm{exec}}$ without retrieved experiences, summarize the resulting trajectories, and insert the generated experience nodes into an initially empty graph.
This initialization provides the copilot with a non-empty graph for early-stage retrieval and stabilizes training.

\paragraph{Executor LLMs.}
To evaluate whether \method is executor-agnostic, we use different frozen executor LLMs across the two ExpSuite settings.
For \textsc{ExpSuite-Static}, we use Llama-3.2-3B-Instruct as the small executor and Llama-3.1-8B-Instruct as the large executor, both accessed through the NVIDIA NIM API\footnote{\url{https://build.nvidia.com}}.
For \textsc{ExpSuite-Agentic}, we use Qwen3-32B\footnote{\url{https://huggingface.co/Qwen/Qwen3-32B}} as the small executor and Gemini-3.1-Flash-Lite as the large executor.
Qwen3-32B is accessed through the NVIDIA NIM API, while Gemini-3.1-Flash-Lite is accessed through the Google Gemini API.
In all cases, the executor is used only through its input-output interface and is never updated during training or evaluation.

\paragraph{Baseline memory construction.}
For fair comparison, all retrieval-centric experience learning baselines are provided with the same historical trajectories as \method whenever applicable.
This includes ReasoningBank, ExpeL, LightMem, Mem0, AWM, and MemRL.
Each baseline converts the trajectories into its own memory representation using the corresponding memory construction strategy and the prompts described in Appendix~\ref{app:prompt}.
This controls the underlying trajectory source where applicable, but does
not imply identical memory-item counts, context lengths, or compute budgets.

\paragraph{Training and evaluation.}
During training, each sampled task is evaluated twice: once with retrieved experiences and once without retrieved experiences.
The two scores are used to compute the utility-grounded reward in Eq.~\ref{eq:reward}, which simultaneously updates the retrieval copilot through PPO and updates graph utility statistics through exponential moving average.
After each task, the completed trajectory is summarized into a new candidate experience, filtered for near-duplicates, and inserted into the graph.
During evaluation, both the copilot and graph are fixed.
The copilot outputs deterministic control variables via greedy decoding, retrieves the top-$K$ experiences, and passes them to the frozen executor without any further graph or policy updates.

\paragraph{Infrastructure.}
All copilot training experiments are conducted on 4 NVIDIA A6000 GPUs with 48GB memory each using BF16 mixed precision.
We use vLLM~\citep{kwon2023efficient} for efficient rollout generation with tensor parallelism size 1 and GPU memory utilization set to $0.3$.
The experience graph server runs on CPU and communicates with the training loop through HTTP.
A single training run for \textsc{ExpSuite-Agentic} on ALFWorld with 500 steps takes approximately 40 hours, while a training run for \textsc{ExpSuite-Static} with 500 steps takes approximately 20 hours.

\section{Training Procedure of \method}
\label{app:training}

Algorithm~\ref{alg:training} summarizes the full training procedure of \method.
At each iteration, the copilot samples retrieval controls $(R,W)$ for a task $x$ and uses them to retrieve experiences from the current graph.
The frozen executor is then evaluated with and without the retrieved experiences to compute the utility-grounded reward.
This reward is used in two ways: it updates the retrieval copilot through PPO and updates the utility statistics of retrieved graph nodes through an exponential moving average.
Finally, the completed trajectory is summarized into a new experience, filtered for near-duplicates, and inserted into the graph.
At inference time, both the copilot and the graph are fixed; the copilot outputs deterministic control variables via greedy decoding, and the retrieved experiences are directly passed to the frozen executor without further updates.

\begin{algorithm}[t]
\caption{Training Procedure of \method}
\label{alg:training}
\begin{algorithmic}[1]
\REQUIRE Task distribution $\mathcal{D}$, executor $\pi_{\mathrm{exec}}$,
         initial copilot $\pi_{\mathrm{ret}}$, initial graph $G = (V, E)$,
         reward weight $\eta$, utility update rate $\beta$
\FOR{each training iteration}
    \STATE Sample task $x \sim \mathcal{D}$
    \STATE \textit{// Retrieval}
    \STATE Sample $(R, W) \sim \pi_{\mathrm{ret}}(\cdot \mid x)$ \hfill (Eq.~\ref{eq:copilot_output})
    \STATE Compute seed set $V_0$ via semantic seeding \hfill (Eq.~\ref{eq:semantic_seed})
    \STATE Expand candidate set $C$ via personalized PageRank \hfill (Eq.~\ref{eq:ppr})
    \STATE Retrieve $E_{R,W}(x)$ via utility-aware ranking \hfill (Eqs.~\ref{eq:ucb}--\ref{eq:retrieved_set})
    \STATE \textit{// Execution}
    \STATE $s_{\mathrm{with}} \leftarrow S\bigl(x,\, \pi_{\mathrm{exec}}(x,\, E_{R,W}(x))\bigr)$ \hfill (Eq.~\ref{eq:s_with})
    \STATE $s_{\mathrm{without}} \leftarrow S\bigl(x,\, \pi_{\mathrm{exec}}(x,\, \emptyset)\bigr)$ \hfill (Eq.~\ref{eq:s_without})
    \STATE \textit{// Reward}
    \STATE $r \leftarrow (s_{\mathrm{with}} - s_{\mathrm{without}}) + \eta \cdot s_{\mathrm{with}}$ \hfill (Eq.~\ref{eq:reward})
    \STATE \textit{// Copilot update}
    \STATE Update $\pi_{\mathrm{ret}}$ with PPO using reward $r$ \hfill (Eq.~\ref{eq:ppo_obj})
    \STATE \textit{// Graph update}
    \FOR{each retrieved node $v \in E_{R,W}(x)$}
        \STATE $n_v \leftarrow n_v + 1$ \hfill (Eq.~\ref{eq:count_update})
        \STATE $u_v \leftarrow (1 - \beta)\, u_v + \beta\, r$ \hfill (Eq.~\ref{eq:utility_update})
    \ENDFOR
    \STATE \textit{// Graph growth}
    \STATE Summarize $\tau' = (x, \xi', y_{\mathrm{with}}, s_{\mathrm{with}})$ into experience $e'$ \hfill (Eq.~\ref{eq:summarize})
    \STATE Filter near-duplicates; evict low-utility nodes if capacity exceeded
    \STATE Insert $e'$ into $G$ \hfill (Eq.~\ref{eq:edge})
\ENDFOR
\ENSURE Trained copilot $\pi_{\mathrm{ret}}$, evolved experience graph $G$
\end{algorithmic}
\end{algorithm}

\section{Dataset Descriptions}
\label{appendix:dataset_descriptions}
We describe all evaluation datasets in ExpSuite below, categorized by their corresponding settings in Table \ref{tab:expsuite_overview}. ExpSuite consists of two complementary groups: ExpSuite-Static and ExpSuite-Agentic. ExpSuite-Static evaluates experience reuse in single-turn input-output tasks, including question answering, mathematical reasoning, and code generation. ExpSuite-Agentic evaluates experience-guided decision-making in multi-step interactive environments. For all datasets, we report task-specific performance metrics, and for interactive tasks we additionally report the number of environment steps to characterize environment interaction length.
\begin{table}[ht]
\small
\setlength{\tabcolsep}{5pt}
\centering
\renewcommand{\arraystretch}{0.95}
\caption{\textbf{Detailed summary of datasets used in ExpSuite.} We categorize datasets by setting, task type, and evaluation metric.}
\label{tab:expsuite_overview}
\begin{tabular}{lcc}
    \toprule
    \textbf{Dataset} & \textbf{Task} & \textbf{Metric} \\
    \midrule
    \multicolumn{3}{l}{\textbf{ExpSuite-Static}} \\
    \midrule
    ARC-C             & Question Answering                  & Accuracy \\
    CommonsenseQA     & Question Answering                  & Accuracy \\
    GPQA-Diamond              & Question Answering                  & Accuracy \\
    MMLU              & Question Answering                  & Accuracy \\
    OBQA              & Question Answering                  & Accuracy \\
    GSM8K             & Mathematical Reasoning              & Exact Match \\
    GSM-Symbolic      & Mathematical Reasoning              & Exact Match \\
    MATH              & Mathematical Reasoning              & Exact Match \\
    HumanEval+        & Code Generation                     & Pass@1 \\
    MBPP+             & Code Generation                     & Pass@1 \\
    \midrule
    \multicolumn{3}{l}{\textbf{ExpSuite-Agentic}} \\
    \midrule
    ALFWorld-Seen     & Seen Interactive Task Completion    & SR / \#Steps \\
    ALFWorld-Unseen   & Unseen Interactive Task Completion  & SR / \#Steps \\
    AppWorld-Test-N   & Normal App-based Workflow Execution & PR / \#Steps \\
    AppWorld-Test-C   & Compositional App-based Workflow Execution & PR / \#Steps \\
    \bottomrule
\end{tabular}
\end{table}

\subsection{ExpSuite-Static}

\subsubsection{Question Answering}

\textbf{ARC} (AI2 Reasoning Challenge)~\citep{arc} contains 7.8K science questions drawn from grade-school standardized exams. Following common practice, we adopt the Challenge split (ARC-C), which retains only questions that cannot be solved by simple retrieval or lexical co-occurrence baselines and therefore demand genuine reasoning.

\textbf{CommonsenseQA}~\citep{CommonsenseQA} consists of 12.2K multiple-choice questions built on top of ConceptNet, where answering correctly requires drawing on implicit, everyday world knowledge about concepts and their relations rather than facts stated in the question itself.

\textbf{GPQA-Diamond}~\citep{rein2024gpqa} is the 198-question high-quality subset of GPQA (Graduate-Level Google-Proof QA), covering biology, physics, and chemistry. It retains questions answered correctly by both expert validators and incorrectly by a majority of non-expert validators, probing subject-matter expertise rather than surface lookup.

\textbf{MMLU} (Massive Multitask Language Understanding)~\citep{MMLU} covers 57 subjects across STEM, the humanities, and the social sciences. Its multiple-choice questions span difficulties from elementary school to professional level, jointly testing factual recall and reasoning.

\textbf{OpenBookQA}~\citep{OpenbookQA} contains 5.9K elementary-school science questions modeled after open-book exams. Solving each item requires combining a given scientific fact with additional commonsense knowledge, which probes multi-hop reasoning over scientific concepts.

\subsubsection{Mathematical Reasoning}

\textbf{GSM8K}~\citep{gsm8k} is a collection of 8.5K grade-school math word problems, each solvable in 2--8 sequential arithmetic steps over the four basic operations. We evaluate by exact match on the final numeric answer.

\textbf{GSM-Symbolic} \citep{gsm_sym} is built from symbolic templates derived from GSM8K test questions. Each template specifies which names and numerical values can vary, along with constraints that keep the problem valid. Sampling different instantiations yields many surface variants of the same underlying problem, all still solved as ordinary numerical word problems. This tests whether a model's arithmetic reasoning holds up under changes to names and numbers, rather than depending on memorized surface forms. We evaluate by exact match on the final numeric answer.

\textbf{MATH}~\citep{math} consists of 12.5K competition-level problems spanning algebra, geometry, number theory, probability, counting, and precalculus. Drawn from contests such as AMC and AIME, the problems require both multi-step reasoning and substantial mathematical background.

\subsubsection{Code Generation}

\textbf{HumanEval+} is an extension of HumanEval~\citep{liu2023your} with a substantially expanded test suite that reduces false positives. It comprises 164 Python programming problems specified by a function signature and docstring, where models must produce code that passes all hidden tests.

\textbf{MBPP+}~\citep{liu2023your} strengthens the Mostly Basic Python Problems benchmark with more rigorous tests, covering 974 crowd-sourced entry-level Python tasks that target standard algorithmic patterns and basic programming proficiency.

\subsection{ExpSuite-Agentic}

\subsubsection{ALFWorld}

\textbf{ALFWorld-Seen}~\citep{shridhar2020alfworld} is a text-based interactive household benchmark in which an agent reads natural-language observations and issues action commands step by step to complete household goals. The \textbf{Seen} evaluation split (\texttt{valid\_seen}) comprises 140 episodes spanning six task types (e.g., \textit{pick-and-place}, \textit{pick-heat-then-place}, \textit{look-at-in-light}); room layouts overlap with training scenarios, so the agent encounters familiar spatial configurations but with novel task instances, isolating generalization at the instruction level.

\textbf{ALFWorld-Unseen}~\citep{shridhar2020alfworld} shares the same six task types but evaluates on \texttt{valid\_unseen}: 134 episodes set in entirely new room layouts with novel object placements, providing a stricter out-of-distribution test of the agent's ability to generalize beyond environments seen during training. For both splits we report task success rate (SR) and average environment steps per episode (\#Steps).

\subsubsection{AppWorld}

\textbf{AppWorld-Test-N}~\citep{trivedi2024appworld} is a controlled benchmark in which an agent executes day-to-day workflows across nine simulated applications (e.g., Gmail, Spotify, Amazon, Todoist, Venmo) by writing Python code that calls their APIs. The \textbf{Normal} test split (\texttt{test\_normal}) consists of 56 scenarios (168 tasks) representing routine single-app or straightforward cross-app operations.

\textbf{AppWorld-Test-C}~\citep{trivedi2024appworld} evaluates agents on the \textbf{Compositional} split (\texttt{test\_challenge}): 139 scenarios (417 tasks) that require coordinating complex multi-app sequences, resolving cross-service dependencies, and handling compositional reasoning across 457 APIs—forming a substantially harder probe of planning and tool-use capabilities. We report pass rate (PR) and mean environment steps (\#Steps) for both splits.

\section{Dataset Statistics}
\label{app:dataset_statistics}
In this section, we present detailed statistics for each dataset. Specifically, the statistics for ExpSuite-Static and ExpSuite-Agentic are provided in Table~\ref{tab:static_data_stats} and Table~\ref{tab:agentic_data_stats}, respectively. For each ExpSuite-Static dataset, we randomly sample 1,500 instances and divide them into training, validation, and test sets following a 5:2:3 ratio; for datasets whose original size is smaller than 1,500 (GPQA-Diamond, HumanEval+, and MBPP+), we apply the same ratio over all available instances. For ExpSuite-Agentic, we adopt the official splits released by each benchmark: ALFWorld provides separate \textit{Seen} and \textit{Unseen} test environments, while AppWorld shares a single training set across two test variants, \textit{Test-N} (normal) and \textit{Test-C} (challenge).

\begin{table}[h]
\centering
\small
\setlength{\tabcolsep}{3.5pt}
\caption{\textbf{ExpSuite-Static Data Statistics.} HEval+ denotes HumanEval+.}
\label{tab:static_data_stats}
\begin{tabularx}{\linewidth}{l *{10}{>{\centering\arraybackslash}X}}
\toprule
\multirow{2}{*}{Split} & \multicolumn{3}{c}{Math} & \multicolumn{5}{c}{QA} & \multicolumn{2}{c}{Code} \\
\cmidrule(lr){2-4}\cmidrule(lr){5-9}\cmidrule(lr){10-11}
& GSM8K & GSM-sym & MATH & MMLU & CSQA & OBQA & ARC-C & \shortstack{GPQA-\\Diamond} & HEval+ & MBPP+ \\
\midrule
Train    & 750 & 750 & 750 & 750 & 750 & 750 & 750 & 99 & 65 & 132 \\
Valid    & 300 & 300 & 300 & 300 & 300 & 300 & 300 & 39 & 26 & 52  \\
Test     & 450 & 450 & 450 & 450 & 450 & 450 & 450 & 60 & 39 & 80  \\
\bottomrule
\end{tabularx}
\end{table}

\begin{table}[h]
\centering
\small
\setlength{\tabcolsep}{3.5pt}
\caption{\textbf{ExpSuite-Agentic Data Statistics.}}
\label{tab:agentic_data_stats}
\begin{tabularx}{\linewidth}{l *{4}{>{\centering\arraybackslash}X}}
\toprule
\multirow{2}{*}{Split} & \multicolumn{2}{c}{ALFWorld} & \multicolumn{2}{c}{AppWorld} \\
\cmidrule(lr){2-3}\cmidrule(lr){4-5}
& Seen & Unseen & Test-N & Test-C \\
\midrule
Train    & 3,553 & 3,553 & 90  & 90  \\
Test     & 140   & 134   & 168 & 417 \\
\bottomrule
\end{tabularx}
\end{table}

\section{Prompt Usage}
\label{app:prompt}
This section separates retrieval-control prompts, experience-construction
prompts, and executor prompts for \method and the documented baselines. In
\method, the retrieval copilot outputs $(R,W)$, the summarizer converts completed
trajectories into experience, and the frozen executor produces answers, code,
or environment actions. In ExpSuite-Static, this executor is Llama-3.2-3B-Instruct
or Llama-3.1-8B-Instruct; the Qwen retrieval copilot does not answer the task.
The role-separated \method templates below specify these interfaces rather
than byte-for-byte API messages; model-specific chat serialization is distinct
from the textual output tags.

\subsection{Copilot Prompting}
\label{app:copilot_prompt}

The retrieval copilot $\pi_{\mathrm{ret}}$ receives a task description and outputs two control parameters $(R, W)$ that govern how experiences are retrieved from the experience graph (see \S\ref{sec:retrieval}).
We design task-category-specific prompt templates that share a common structure but differ in role descriptions and task framing.
All templates instruct the copilot to output its parameters in the format \texttt{<search>R:W</search>}, optionally preceded by brief reasoning in \texttt{<think>...</think>} tags.
Below we describe the three prompt categories and their design rationale.

\paragraph{ExpSuite-Agentic: ALFWorld.}
For ALFWorld, the copilot is framed as a \emph{retrieval strategy controller for an embodied household agent}.
The prompt presents the task description (e.g., \texttt{pick\_and\_place-Mug-None-Cabinet-312}) and explains how $R$ and $W$ control graph diffusion scope and utility weighting, respectively.
Because ALFWorld tasks fall into a small number of recurring categories (e.g., pick-and-place, heat-then-place, examine-in-light), the copilot can learn to associate task types with effective retrieval strategies—for instance, using high $W$ for familiar task types where historical utility is reliable, and low $R$ for novel configurations that benefit from broader exploration.
The controller template is shown in Table~\ref{tab:prompt-copilot-alfworld}.

\paragraph{ExpSuite-Agentic: AppWorld.}
For AppWorld, the copilot is framed as a \emph{retrieval strategy controller for a coding agent in an app-based environment}.
The prompt structure is identical to ALFWorld, with the task description replaced by a natural-language instruction (e.g., ``What is the title of the most-liked song in my Spotify playlists'').
AppWorld tasks are more diverse than ALFWorld in terms of required APIs and multi-app dependencies, so the copilot must learn to vary its retrieval strategy across a wider range of task structures.
The controller template is shown in Table~\ref{tab:prompt-copilot-appworld}.

\paragraph{ExpSuite-Static: QA, Math, and Code.}
The copilot acts as a retrieval strategy controller for all three categories.
It receives the task specification and returns retrieval controls in
\texttt{<search>R:W</search>}; it does not return an answer letter, numerical
solution, or Python implementation. The retrieved experience text and original
task are passed to a separate frozen Llama executor. Controller templates are
shown in Tables~\ref{tab:prompt-copilot-qa}, \ref{tab:prompt-copilot-math}, and
\ref{tab:prompt-copilot-code}; executor templates are given in
Tables~\ref{tab:prompt-executor-qa}--\ref{tab:prompt-executor-code}.
These controller templates describe the $(R,W)$ output interface in
Eq.~\ref{eq:copilot_output}, rather than an answer-generating search dialogue.

\begin{table}[h]
\centering
\caption{\textbf{Copilot prompt template for ALFWorld} (ExpSuite-Agentic). The copilot outputs retrieval parameters $(R, W)$ for graph-based experience retrieval. \texttt{\{task\_description\}} is replaced with the ALFWorld task identifier.}
\label{tab:prompt-copilot-alfworld}
\small
\begin{tabular}{p{13cm}}
\toprule[1.1pt]
\texttt{<|im\_start|>}\\
\textit{user}\\[4pt]
You are a retrieval strategy controller for an embodied agent in the ALFRED household environment.\\[4pt]
Task: \textbf{\{task\_description\}}\\[4pt]
You control how experiences are retrieved from a knowledge graph by setting two parameters:\\[4pt]
R (0-100): Restart/locality control for graph diffusion.\\
\quad Low R (0-30): Explore widely through connected experiences in the graph (broad discovery).\\
\quad High R (70-100): Stay close to the most similar experiences (precise, local search).\\[4pt]
W (0-100): Selection preference between semantic similarity and historical utility.\\
\quad Low W (0-30): Prefer candidates most similar to this task.\\
\quad High W (70-100): Give greater weight to historical shared-reward estimates.\\[4pt]
When W=0, candidates are ranked only by cosine similarity within the graph-derived candidate set. Increase W to incorporate historical feedback.\\[4pt]
Output format: \texttt{<search>R:W</search>}\\
You may reason briefly in \texttt{<think>...</think>} tags first.\\[4pt]
Examples:\\
- New/unfamiliar task type $\rightarrow$ \texttt{<search>20:10</search>}\\
- Common task with good past data $\rightarrow$ \texttt{<search>80:70</search>}\\
- Moderate confidence $\rightarrow$ \texttt{<search>50:40</search>}\\[4pt]
Output the complete retrieval-control tag. Do not output an environment action or a task answer.\\
\texttt{<|im\_end|>}\\
\bottomrule[1.1pt]
\end{tabular}
\end{table}

\begin{table}[h]
\centering
\caption{\textbf{Copilot prompt template for AppWorld} (ExpSuite-Agentic). Structure is identical to ALFWorld; only the role description and task content differ.}
\label{tab:prompt-copilot-appworld}
\small
\begin{tabular}{p{13cm}}
\toprule[1.1pt]
\texttt{<|im\_start|>}\\
\textit{user}\\[4pt]
You are a retrieval strategy controller for a coding agent in the AppWorld environment.\\[4pt]
Task: \textbf{\{task\_description\}}\\[4pt]
You control how experiences are retrieved from a knowledge graph by setting two parameters:\\[4pt]
R (0-100): Restart/locality control for graph diffusion.\\
\quad Low R (0-30): Explore widely through connected experiences in the graph (broad discovery).\\
\quad High R (70-100): Stay close to the most similar experiences (precise, local search).\\[4pt]
W (0-100): Selection preference between semantic similarity and historical utility.\\
\quad Low W (0-30): Prefer candidates most similar to this task.\\
\quad High W (70-100): Give greater weight to historical shared-reward estimates.\\[4pt]
When W=0, candidates are ranked only by cosine similarity within the graph-derived candidate set. Increase W to incorporate historical feedback.\\[4pt]
Output format: \texttt{<search>R:W</search>}\\
You may reason briefly in \texttt{<think>...</think>} tags first.\\[4pt]
Examples:\\
- New/unfamiliar task type $\rightarrow$ \texttt{<search>20:10</search>}\\
- Common task with good past data $\rightarrow$ \texttt{<search>80:70</search>}\\
- Moderate confidence $\rightarrow$ \texttt{<search>50:40</search>}\\[4pt]
Output the complete retrieval-control tag. Do not output an environment action or a task answer.\\
\texttt{<|im\_end|>}\\
\bottomrule[1.1pt]
\end{tabular}
\end{table}

\begin{table}[h]
\centering
\caption{\textbf{Retrieval-copilot template for Question Answering} (ExpSuite-Static). Qwen outputs retrieval controls; the separate frozen Llama executor generates the task response.}
\label{tab:prompt-copilot-qa}
\small
\begin{tabular}{p{13cm}}
\toprule[1.1pt]
\textit{System role:} You are a retrieval strategy controller for a frozen task-solving model.\\[4pt]
\textit{User input:}\\
Question: \textbf{\{question\}}\\
\textbf{\{choices\}}\\
R (0--100): Restart/locality control. Low R allows broader graph diffusion; high R keeps retrieval closer to the semantic seeds.\\[4pt]
W (0--100): Similarity--utility trade-off. Low W favors semantic similarity; high W gives more weight to historical utility.\\[4pt]
When W=0, ranking uses only similarity within the graph-derived candidate set.\\[4pt]
You may explain your retrieval choice briefly in \texttt{<think>...</think>} tags.\\
Output the retrieval parameters as \texttt{<search>R:W</search>}.\\
Do not solve the task or output its final answer or implementation.\\
\bottomrule[1.1pt]
\end{tabular}
\end{table}

\begin{table}[h]
\centering
\caption{\textbf{Retrieval-copilot template for Mathematical Reasoning} (ExpSuite-Static). Qwen outputs retrieval controls; the separate frozen Llama executor generates the task response.}
\label{tab:prompt-copilot-math}
\small
\begin{tabular}{p{13cm}}
\toprule[1.1pt]
\textit{System role:} You are a retrieval strategy controller for a frozen task-solving model.\\[4pt]
\textit{User input:}\\
Problem: \textbf{\{question\}}\\
R (0--100): Restart/locality control. Low R allows broader graph diffusion; high R keeps retrieval closer to the semantic seeds.\\[4pt]
W (0--100): Similarity--utility trade-off. Low W favors semantic similarity; high W gives more weight to historical utility.\\[4pt]
When W=0, ranking uses only similarity within the graph-derived candidate set.\\[4pt]
You may explain your retrieval choice briefly in \texttt{<think>...</think>} tags.\\
Output the retrieval parameters as \texttt{<search>R:W</search>}.\\
Do not solve the task or output its final answer or implementation.\\
\bottomrule[1.1pt]
\end{tabular}
\end{table}

\begin{table}[h]
\centering
\caption{\textbf{Retrieval-copilot template for Code Generation} (ExpSuite-Static). Qwen outputs retrieval controls; the separate frozen Llama executor generates the task response.}
\label{tab:prompt-copilot-code}
\small
\begin{tabular}{p{13cm}}
\toprule[1.1pt]
\textit{System role:} You are a retrieval strategy controller for a frozen task-solving model.\\[4pt]
\textit{User input:}\\
Task: \textbf{\{task\}}\\
Function signature: \textbf{\{function\_signature\}}\\
R (0--100): Restart/locality control. Low R allows broader graph diffusion; high R keeps retrieval closer to the semantic seeds.\\[4pt]
W (0--100): Similarity--utility trade-off. Low W favors semantic similarity; high W gives more weight to historical utility.\\[4pt]
When W=0, ranking uses only similarity within the graph-derived candidate set.\\[4pt]
You may explain your retrieval choice briefly in \texttt{<think>...</think>} tags.\\
Output the retrieval parameters as \texttt{<search>R:W</search>}.\\
Do not solve the task or output its final answer or implementation.\\
\bottomrule[1.1pt]
\end{tabular}
\end{table}

\subsection{Experience Construction}

Each baseline that maintains an experience bank processes training rollout trajectories via LLM calls to produce structured memory items, but the construction strategy differs substantially across methods. For ExpSuite-Static, the method-specific construction templates are described below.
Any \texttt{answer}, \texttt{expected\_answer}, or response field in this
subsection belongs to a completed training trajectory or its feedback; it is
an input to experience construction, not an instruction for the retrieval
copilot to answer a new task. For ExpSuite-Agentic, each method additionally uses environment-specific prompts tailored to ALFWorld household manipulation and AppWorld API-based workflows, described alongside the static prompts for each method below.

\textbf{ReasoningBank} uses a unified extractor that processes each trajectory into a structured memory item. For ExpSuite-Static, successful trajectories are processed with the prompt in Table~\ref{tab:prompt-success-memory}; failed trajectories use Table~\ref{tab:prompt-failure-memory}.
For ExpSuite-Agentic, ALFWorld trajectories use Table~\ref{tab:reasoningbank-alfworld}, and AppWorld trajectories use Table~\ref{tab:reasoningbank-appworld}.

\textbf{ExpeL}~\citep{zhao2024expel} maintains a global rule set updated via two critique passes. A \emph{compare-critique} pass pairs each success with failures from the same dataset and asks the LLM to emit ADD/EDIT/REMOVE/AGREE operations over existing rules (Table~\ref{tab:expel-compare}). An \emph{all-success} pass fires every $S=8$ successes per dataset and similarly refines rules from positive evidence alone (Table~\ref{tab:expel-success}). In addition, every rollout is stored verbatim as an exemplar memory item (with \texttt{title=``exemplar''}), and finalized rules are stored as items with \texttt{title=``rule''}. For ExpSuite-Agentic, ExpeL uses environment-specific critique prompts for ALFWorld (Table~\ref{tab:expel-alfworld}) and AppWorld (Table~\ref{tab:expel-appworld}).

\textbf{LightMem}~\citep{fang2026lightmem} implements a three-stage STM$\to$LTM pipeline. A short-term memory buffer (size $K=5$) is periodically flushed to long-term memory via a MemoryExtractor LLM call using Table~\ref{tab:prompt-success-memory} and Table~\ref{tab:prompt-failure-memory}. A separate offline consolidation pass then reconciles similar LTM entries using the UPDATE/DELETE/IGNORE prompt in Table~\ref{tab:lightmem-consolidation}. For ExpSuite-Agentic, LightMem uses environment-specific extraction prompts for ALFWorld (Table~\ref{tab:lightmem-alfworld}) and AppWorld (Table~\ref{tab:lightmem-appworld}).

\textbf{Mem0}~\citep{chhikara2025mem0} runs a two-step write pipeline per rollout. Step~1 extracts salient facts from the (problem, response) pair via Table~\ref{tab:mem0-extract}. Step~2 retrieves the top-$K$ most similar existing memories, then asks the LLM to emit ADD/UPDATE/DELETE/NONE operations via Table~\ref{tab:mem0-decide}. For successful rollouts, a third reflect step generates a procedural summary via Table~\ref{tab:mem0-reflect}. For ExpSuite-Agentic, Mem0 uses environment-specific prompts for ALFWorld (Table~\ref{tab:mem0-alfworld}) and AppWorld (Table~\ref{tab:mem0-appworld}).

\textbf{AWM}~\citep{wang2024agent} induces a per-dataset workflow from a pool of up to $N=5$ successful trajectories. The induction prompt (Table~\ref{tab:awm-induction}) asks the LLM to extract repetitive reasoning patterns as step-by-step workflows with \texttt{\{placeholder\}} names; the resulting workflow string is stored as a plain key--value entry indexed by dataset name and prepended verbatim at inference time without retrieval. For ExpSuite-Agentic, AWM uses environment-specific induction prompts for ALFWorld (Table~\ref{tab:awm-alfworld}) and AppWorld (Table~\ref{tab:awm-appworld}).

\textbf{\method} uses a summarizer to extract experience text from completed
executor trajectories; the graph component embeds the resulting entries and
constructs their links as described in \S\ref{sec:graph_construction}. For ExpSuite-Static, the construction prompt is given in Table~\ref{tab:expgraph-static-sucess} and Table~\ref{tab:expgraph-static-failure}.
For ExpSuite-Agentic, ALFWorld trajectories use Table~\ref{tab:expgraph-alfworld-success} and Table~\ref{tab:expgraph-alfworld-failure}; AppWorld trajectories use Table~\ref{tab:expgraph-appworld}.

\subsection{Executor and Baseline Answering Prompts}
\label{app:executor_prompt}

At inference time, task-solving models receive the original task and any
method-specific experience context. Answer tags, Python code blocks, and
\texttt{complete\_task(answer=\ldots)} calls in this subsection are outputs of
the task-solving model, not outputs of the \method retrieval copilot. Baseline
search protocols and output formats are kept distinct from \method's control
interface.

\paragraph{\method: frozen Llama executor.}
For ExpSuite-Static, the harness passes the original task and retrieved experience
text to Llama-3.2-3B-Instruct or Llama-3.1-8B-Instruct. These frozen executors
perform reasoning and generate the final answer or code. The role-separated
answering templates are shown in Tables~\ref{tab:prompt-executor-qa}--\ref{tab:prompt-executor-code};
the controller templates in \S\ref{app:copilot_prompt} contain no answer-generation
instructions. A separate summarizer processes completed training trajectories
into experience; neither it nor the retrieval copilot replaces the executor.

\begin{table}[h]
\centering
\caption{\textbf{Frozen Llama executor template for Question Answering} (ExpSuite-Static). The executor receives retrieved experience and generates the task response; it does not output retrieval controls.}
\label{tab:prompt-executor-qa}
\small
\begin{tabular}{p{13cm}}
\toprule[1.1pt]
\textit{System role:} You are a helpful task-solving assistant.\\[4pt]
\textit{User input:}\\
Retrieved experience:\\
\textbf{\{retrieved\_experiences\}}\\[4pt]
Question: \textbf{\{question\}}\\
\textbf{\{choices\}}\\
Answer the multiple-choice question using the provided experience where relevant.\\
Analyze the options, then give the final answer letter inside \texttt{<answer>...</answer>}.\\
For example: \texttt{<answer> B </answer>}.\\
\bottomrule[1.1pt]
\end{tabular}
\end{table}

\begin{table}[h]
\centering
\caption{\textbf{Frozen Llama executor template for Mathematical Reasoning} (ExpSuite-Static). The executor receives retrieved experience and generates the task response; it does not output retrieval controls.}
\label{tab:prompt-executor-math}
\small
\begin{tabular}{p{13cm}}
\toprule[1.1pt]
\textit{System role:} You are a helpful task-solving assistant.\\[4pt]
\textit{User input:}\\
Retrieved experience:\\
\textbf{\{retrieved\_experiences\}}\\[4pt]
Problem: \textbf{\{question\}}\\
Solve the problem using the provided experience where relevant.\\
Show your reasoning and put the final mathematical answer inside \texttt{<answer>...</answer>}.\\
\bottomrule[1.1pt]
\end{tabular}
\end{table}

\begin{table}[h]
\centering
\caption{\textbf{Frozen Llama executor template for Code Generation} (ExpSuite-Static). The executor receives retrieved experience and generates the task response; it does not output retrieval controls.}
\label{tab:prompt-executor-code}
\small
\begin{tabular}{p{13cm}}
\toprule[1.1pt]
\textit{System role:} You are a helpful task-solving assistant.\\[4pt]
\textit{User input:}\\
Retrieved experience:\\
\textbf{\{retrieved\_experiences\}}\\[4pt]
Task: \textbf{\{task\}}\\
Function signature: \textbf{\{function\_signature\}}\\
Solve the coding task using the provided experience where relevant.\\
Your code should pass the supplied tests:\\
\textbf{\{test\_cases\}}\\
Think through your approach, then put the complete Python function inside \texttt{<answer>...</answer>}.\\
\bottomrule[1.1pt]
\end{tabular}
\end{table}

\paragraph{No Memory.}
Chain-of-thought uses no retrieved experience; the model solves tasks directly with chain-of-thought instructions (Table~\ref{tab:cot-qa}--\ref{tab:cot-appworld}).

\paragraph{Experience-conditioned baseline executors.}
The documented implementations of ReasoningBank, ExpeL, LightMem, Mem0, AWM,
and IRCoT use the following base answering templates (Table~\ref{tab:retrieval-qa}--\ref{tab:retrieval-appworld}), differing only in the content of \textbf{\{retrieved\_memories\}}:
\begin{itemize}
    \item \textbf{ReasoningBank}: top-$K$ memory items retrieved by dense embedding similarity, formatted as \texttt{\#\# Retrieved Memory: ...}
    \item \textbf{ExpeL}: distilled rules (up to 10) prepended as a numbered list, plus top-10 exemplar trajectories retrieved by KNN.
    \item \textbf{LightMem}: LTM items after STM$\to$LTM periodic consolidation, retrieved by dense embedding.
    \item \textbf{Mem0}: top-10 procedural summaries maintained via ADD/UPDATE/DELETE/NONE operations on prior (problem, response) pairs.
    \item \textbf{AWM}: full per-dataset-type workflow block prepended without retrieval (O(1) dict lookup).
    \item \textbf{IRCoT}: BM25-retrieved raw rollout pairs formatted as \texttt{Example:\textbackslash nQuestion:...\textbackslash nSolution:...}, accumulated across up to 3 iterative rounds.
\end{itemize}

\paragraph{Search-o1.}
Search-o1 instructs the model to issue explicit memory search queries via \texttt{<|begin\_search\_query|>} tags. Retrieved items are first summarized by a separate LLM call (Table~\ref{tab:search-o1-summarize}) before being injected as search results. Inference prompts follow the same structure (Table~\ref{tab:searcho1-qa}--\ref{tab:search-o1-appworld}).

\begin{table}[h]
\centering
\caption{\textbf{Prompt for extracting a memory item from a successful trajectory (ReasoningBank; also used by LightMem's MemoryExtractor) in ExpSuite-Static.}}
\label{tab:prompt-success-memory}
\small
\begin{tabular}{p{13cm}}
\toprule[1.1pt]
\texttt{<|im\_start|>}\\
\textit{system}\\
You will be given a user query and the corresponding trajectory that represents how an agent successfully accomplished the task.\\[4pt]
\textbf{Guidelines}\\
You need to extract and summarize useful insights in the format of memory items based on the agent's successful trajectory.
The goal of summarized memory items is to be helpful and generalizable for future similar tasks.\\[4pt]
\textbf{Important Notes}\\
- You must first think why the trajectory is successful, and then summarize the insights.\\
- You can extract only 1 memory item from the trajectory.\\
- You must not repeat similar or overlapping items.\\
- Do not mention specific values or string contents, but rather focus on generalizable insights.\\[4pt]
\textbf{Output Format}\\
Your output must strictly follow the Markdown format shown below:\\
\texttt{\# Memory Item 1}\\
\texttt{\#\# Title:} \texttt{<}the title of the memory item\texttt{>}\\
\texttt{\#\# Description:} \texttt{<}one-sentence summary of the memory item\texttt{>}\\
\texttt{\#\# Content:} \texttt{<}1--3 sentences describing the insights learned\texttt{>}\\
\texttt{<|im\_end|>}\\
\texttt{<|im\_start|>}\\
\textit{user}\\
Query: \textbf{\{question\}}\\
Trajectory: \textbf{\{reasoning\_trajectory\}}\\
\texttt{<|im\_end|>}\\
\bottomrule[1.1pt]
\end{tabular}
\end{table}

\begin{table}[h]
\centering
\caption{\textbf{Prompt for extracting a memory item from a failed trajectory (ReasoningBank; also used by LightMem's MemoryExtractor) in ExpSuite-Static.}}
\label{tab:prompt-failure-memory}
\small
\begin{tabular}{p{13cm}}
\toprule[1.1pt]
\texttt{<|im\_start|>}\\
\textit{system}\\
You will be given a user query and the corresponding trajectory that represents how an agent attempted the task but failed.\\[4pt]
\textbf{Guidelines}\\
You need to extract and summarize useful insights based on the agent's failed trajectory.
The goal of summarized memory items is to be helpful and generalizable for future similar tasks.\\[4pt]
\textbf{Important Notes}\\
- You must first reflect on why the trajectory failed, then summarize lessons learned to prevent similar failures.\\
- You can extract only 1 memory item from the trajectory.\\
- You must not repeat similar or overlapping items.\\
- Focus on generalizable insights, not specific values.\\[4pt]
\textbf{Output Format}\\
Your output must strictly follow the Markdown format shown below:\\
\texttt{\# Memory Item 1}\\
\texttt{\#\# Title:} \texttt{<}the title of the memory item\texttt{>}\\
\texttt{\#\# Description:} \texttt{<}one-sentence summary of the memory item\texttt{>}\\
\texttt{\#\# Content:} \texttt{<}1--3 sentences describing the lessons learned\texttt{>}\\
\texttt{<|im\_end|>}\\
\texttt{<|im\_start|>}\\
\textit{user}\\
Query: \textbf{\{question\}}\\
Expected Answer: \textbf{\{expected\_answer\}}\\
Trajectory: \textbf{\{reasoning\_trajectory\}}\\
\texttt{<|im\_end|>}\\
\bottomrule[1.1pt]
\end{tabular}
\end{table}

\begin{table}[h]
\centering
\caption{\textbf{Prompt for ReasoningBank extracting a memory item in ALFWorld.}}
\label{tab:reasoningbank-alfworld}
\small
\begin{tabular}{p{13cm}}
\toprule[1.1pt]
You will be given a finished household trajectory (task goal, action--observation log) from a text-based ALFWorld environment, plus its outcome (success or failure). Distill the trajectory into a compact \emph{reasoning rationale} that a future agent can reuse on a structurally similar task. \\
 \\
Output exactly one Markdown block of the form: \\
\#\# Memory Item \\
\textbf{Title}: \textbf{\{short reusable name\}} \\
\textbf{Description}: \textbf{\{1--2 sentence task pattern\}} \\
\textbf{Content}: \textbf{\{step-by-step heuristic; cite key receptacles and verbs as schema variables, e.g. ``go to \{recep\}'' / ``cool \{X\} with fridge 1''\}} \\
 \\
Trajectory: \\
\textbf{\{trajectory\_text\}} \\
 \\
Outcome: \textbf{\{success $\vert$ failure\}} \\
\bottomrule[1.1pt]
\end{tabular}
\end{table}

\begin{table}[h]
\centering
\caption{\textbf{Prompt for ReasoningBank extracting a memory item in AppWorld.}}
\label{tab:reasoningbank-appworld}
\small
\begin{tabular}{p{13cm}}
\toprule[1.1pt]
You are reading a finished AppWorld task trajectory: an instruction, a sequence of \texttt{python} code calls against the AppWorld API, and a final \texttt{complete\_task(answer=\ldots)} call (or failure). Distill this trajectory into a compact reusable rationale a future agent can retrieve when it sees a similar task. \\
 \\
Output exactly one Markdown block of the form: \\
\#\# Memory Item \\
\textbf{Title}: \textbf{\{short reusable name\}} \\
\textbf{Description}: \textbf{\{which app(s); what intent\}} \\
\textbf{Content}: \textbf{\{step-by-step plan, naming the key APIs and parameter shapes; substitute concrete IDs/emails with \{placeholder\} variables\}} \\
 \\
Instruction: \textbf{\{instruction\}} \\
Trajectory: \textbf{\{code log\}} \\
Outcome: \textbf{\{success $\vert$ failure\}} \\
\bottomrule[1.1pt]
\end{tabular}
\end{table}

\begin{table}[h]
\centering
\caption{\textbf{Prompt for ExpeL compare-critique pass (success vs.\ failure pair) in ExpSuite-Static.} The LLM emits ADD/EDIT/REMOVE/AGREE operations over the current rule set. \textbf{\{existing\_rules\}} is replaced by the numbered rule list; \textbf{\{task\}}, \textbf{\{success\_history\}}, and \textbf{\{fail\_history\}} are filled from the rollout pair.}
\label{tab:expel-compare}
\small
\begin{tabular}{p{13cm}}
\toprule[1.1pt]
\textit{system}\\
You are an advanced reasoning agent that can add, edit or remove rules from your existing rule set, based on forming new critiques of past task trajectories. You will be given two previous reasoning task trials: one successful and one unsuccessful. You need to analyze what made the successful trial succeed and the unsuccessful trial fail.\\[4pt]
\textit{user}\\
Here are the two previous trials to compare and critique:\\[2pt]
TRIAL TASK: \textbf{\{task\}}\\[2pt]
SUCCESSFUL TRIAL: \textbf{\{success\_history\}}\\[2pt]
FAILED TRIAL: \textbf{\{fail\_history\}}\\[2pt]
Here are the EXISTING RULES: \textbf{\{existing\_rules\}}\\[4pt]
By examining and contrasting to the successful trial, and the list of existing rules, you can perform the following operations: add, edit, remove, or agree so that the new list of rules is GENERAL and HIGH LEVEL critiques of the failed trial or proposed way of Thought so they can be used to avoid similar failures when encountered with different questions in the future.\\[4pt]
The available operations are: AGREE (if the existing rule is strongly relevant), REMOVE (if a rule is contradictory or duplicated), EDIT (if a rule can be enhanced), ADD (if a new, distinct rule is needed). Format:\\[2pt]
\texttt{AGREE <N>: <rule>}\\
\texttt{REMOVE <N>: <rule>}\\
\texttt{EDIT <N>: <new rule>}\\
\texttt{ADD <N>: <new rule>}\\[2pt]
Do not mention the trials in the rules --- all rules should be GENERALLY APPLICABLE. Each rule should be concise. Do at most 4 operations; each existing rule can receive at most 1 operation.\\
\texttt{<|im\_end|>}\\
\bottomrule[1.1pt]
\end{tabular}
\end{table}

\begin{table}[h]
\centering
\caption{\textbf{Prompt for ExpeL all-success critique pass in ExpSuite-Static} (fired every $S=8$ successes per dataset). \textbf{\{success\_history\}} contains the concatenated successful trajectories; \textbf{\{existing\_rules\}} is the current numbered rule list.}
\label{tab:expel-success}
\small
\begin{tabular}{p{13cm}}
\toprule[1.1pt]
\textit{system}\\
You are an advanced reasoning agent that can add, edit or remove rules from your existing rule set, based on forming new critiques of past task trajectories. You will be given successful reasoning task trials. Analyze them to extract general reasoning insights.\\[4pt]
\textit{user}\\
Here are the trials: \textbf{\{success\_history\}}\\[2pt]
Here are the EXISTING RULES: \textbf{\{existing\_rules\}}\\[4pt]
By examining the successful trials, and the list of existing rules, you can perform the following operations: add, edit, remove, or agree so that the new list of rules are general and high level insights of the successful trials or proposed way of Thought so they can be used as helpful tips to different tasks in the future.\\[4pt]
The available operations follow the same format as the compare-critique pass (AGREE / REMOVE / EDIT / ADD). Focus on REMOVE rules first, and only ADD a rule if it is very insightful and distinct from existing rules. Do not mention the trials in the rules.\\
\texttt{<|im\_end|>}\\
\bottomrule[1.1pt]
\end{tabular}
\end{table}

\begin{table}[h]
\centering
\caption{\textbf{Prompt for ExpeL extracting a memory item in ALFWorld.}}
\label{tab:expel-alfworld}
\small
\begin{tabular}{p{13cm}}
\toprule[1.1pt]
You are an advanced reasoning agent that improves by reflecting on past ALFWorld trials. You are given two trials on similar household tasks: one successful and one unsuccessful. Compare them and decide which existing rules in your rule set should be \emph{ADDED}, \emph{EDITED}, or \emph{REMOVED}, or which can stay (\emph{AGREE}). Each rule is a one-sentence heuristic for a structural sub-problem (e.g. ``before placing two of the same item, locate both before relocating''). \\
 \\
Output one operation per line, in the form: \\
\textbf{ADD} \textit{\textless rule text\textgreater} \\
\textbf{EDIT} \textit{\textless rule\_id\textgreater} \textrightarrow{} \textit{\textless new text\textgreater} \\
\textbf{REMOVE} \textit{\textless rule\_id\textgreater} \\
\textbf{AGREE} \textit{\textless rule\_id\textgreater} \\
 \\
Existing rules: \textbf{\{rules\}} \\
Successful trial: \textbf{\{traj\_succ\}} \\
Failed trial: \textbf{\{traj\_fail\}} \\
\bottomrule[1.1pt]
\end{tabular}
\end{table}

\begin{table}[h]
\centering
\caption{\textbf{Prompt for ExpeL extracting a memory item in AppWorld.}}
\label{tab:expel-appworld}
\small
\begin{tabular}{p{13cm}}
\toprule[1.1pt]
You are an advanced reasoning agent that improves by reflecting on past AppWorld coding trials. You are given two finished trials on similar API-call tasks: one whose final \texttt{complete\_task} call passed the test cases and one that did not. Compare them and emit ADD/EDIT/REMOVE/AGREE operations over the existing rule set. Each rule is a one-sentence heuristic about API usage (e.g.\ ``always call \texttt{login} before any per-app endpoint''). \\
 \\
Output one operation per line, same format as the ALFWorld variant (Table~\ref{tab:expel-alfworld}). \\
 \\
Existing rules: \textbf{\{rules\}} \\
Successful trial: \textbf{\{code\_log\_succ\}} \\
Failed trial: \textbf{\{code\_log\_fail\}} \\
\bottomrule[1.1pt]
\end{tabular}
\end{table}

\begin{table}[h]
\centering
\caption{\textbf{Prompt for LightMem offline LTM consolidation in ExpSuite-Static.} For each target memory with at least one similar candidate (cosine similarity $\geq 0.8$), the LLM decides whether to UPDATE, DELETE, or IGNORE the target. \textbf{\{target\_memory\}} and \textbf{\{candidate\_memories\}} are filled from the LTM.}
\label{tab:lightmem-consolidation}
\small
\begin{tabular}{p{13cm}}
\toprule[1.1pt]
\textit{system}\\
You are a memory management assistant. Your task is to decide whether the target memory should be updated, deleted, or ignored based on the candidate source memories.\\[4pt]
Decision rules:\\
1. \textbf{Update}: If the target and candidate memories describe the same fact but are not fully consistent (candidates provide more detail or refinements), update the target by integrating the additional information.\\
2. \textbf{Delete}: If the target and candidate memories contain a direct conflict, the candidates (more recent) take precedence. Delete the target.\\
3. \textbf{Ignore}: If target and candidates are unrelated, take no action.\\[4pt]
Additional guidance: Use only the information provided. Do not invent details. Your operation always applies to the target memory; do not modify candidate memories.\\[4pt]
The output must be a JSON object:\\
\texttt{\{"action": "update" | "delete" | "ignore", "new\_memory": \{...\}\}}\\
(\texttt{new\_memory} required only when \texttt{action = "update"})\\[4pt]
\textit{user}\\
Target memory: \textbf{\{target\_memory\}}\\
Candidate memories: \textbf{\{candidate\_memories\}}\\
\bottomrule[1.1pt]
\end{tabular}
\end{table}

\begin{table}[h]
\centering
\caption{\textbf{Prompt for LightMem extracting a memory item in ALFWorld.}}
\label{tab:lightmem-alfworld}
\small
\begin{tabular}{p{13cm}}
\toprule[1.1pt]
You are a Personal Information Extractor. The input is a buffer of recently observed ALFWorld turns, each tagged with a topic and a timestamp. Extract \emph{all} salient facts about the agent's environment and behaviour: object locations, completed sub-goals, durable preferences, errors. \\
 \\
Format each fact as: \\
\textbf{FACT}: \textbf{\{one-sentence statement\}}\ \ \textbf{[}\textbf{\{ts\}}\textbf{]}\ \ \textbf{(source\_id=\{seq\})} \\
 \\
{[\ldots verbatim few-shot examples omitted; full prompt at \texttt{baselines/lightmem/memory.py:81}\ldots]} \\
 \\
Conversation buffer: \\
\textbf{\{buffer\}} \\
\bottomrule[1.1pt]
\end{tabular}
\end{table}

\begin{table}[h]
\centering
\caption{\textbf{Prompt for LightMem extracting a memory item in AppWorld.}}
\label{tab:lightmem-appworld}
\small
\begin{tabular}{p{13cm}}
\toprule[1.1pt]
You are a Personal Information Extractor. The input is a buffer of recently executed AppWorld turns: code chunks the agent ran, plus the JSON response from each \texttt{apis.*} call. Extract salient durable facts: app credentials/tokens, IDs of relevant entities (user, contact, note, playlist, transaction), API quirks, and successful subroutines. \\
 \\
Format each fact as: \\
\textbf{FACT}: \textbf{\{one-sentence statement\}}\ \ \textbf{[}\textbf{\{ts\}}\textbf{]}\ \ \textbf{(source\_id=\{seq\})} \\
 \\
{[\ldots verbatim few-shot examples omitted; full prompt at \texttt{baselines/lightmem/memory.py:81}\ldots]} \\
 \\
Conversation buffer: \\
\textbf{\{buffer\}} \\
\bottomrule[1.1pt]
\end{tabular}
\end{table}

\begin{table}[h]
\centering
\caption{\textbf{Prompt for Mem0 Step~1: fact extraction from a (problem, response) pair in ExpSuite-Static.} The LLM returns a JSON object \texttt{\{"facts": [...]\}} listing salient procedural facts.}
\label{tab:mem0-extract}
\small
\begin{tabular}{p{13cm}}
\toprule[1.1pt]
\textit{system}\\
You are a Personal Information Organizer, specialized in accurately storing facts, user memories, and preferences. Your primary role is to extract relevant pieces of information from conversations and organize them into distinct, manageable facts.\\[4pt]
Extract facts and preferences and return them in JSON format:\\
\texttt{\{"facts": ["<fact 1>", "<fact 2>", ...]\}}\\[4pt]
Guidelines: store personal preferences, important details, plans, activity preferences, professional details, and other relevant information. Do not extract anything from system messages. Detect the user's language and record facts in the same language. If nothing relevant is found, return an empty list.\\[4pt]
\textit{user}\\
Problem: \textbf{\{question\}}\\
Solution: \textbf{\{response\}}\\
\bottomrule[1.1pt]
\end{tabular}
\end{table}

\begin{table}[h]
\centering
\caption{\textbf{Prompt for Mem0 Step~2: operation decision (ADD / UPDATE / DELETE / NONE) in ExpSuite-Static.} \textbf{\{old\_memories\}} contains the top-$K$ retrieved existing memory entries as a JSON list; \textbf{\{new\_facts\}} contains the facts extracted in Step~1.}
\label{tab:mem0-decide}
\small
\begin{tabular}{p{13cm}}
\toprule[1.1pt]
\textit{user}\\
You are a smart memory manager which controls the memory of a system. You can perform four operations: (1) add into the memory, (2) update the memory, (3) delete from the memory, and (4) no change.\\[4pt]
Compare newly retrieved facts with the existing memory. For each new fact, decide whether to ADD (new information not present), UPDATE (same topic but different content; keep the more informative version), DELETE (direct contradiction; newer facts take precedence), or NONE (already present or irrelevant).\\[4pt]
Current memory: \textbf{\{old\_memories\}}\\[2pt]
New retrieved facts: \textbf{\{new\_facts\}}\\[4pt]
Return your response strictly in the following JSON structure:\\
\texttt{\{"memory": [\{"id": "<id>", "text": "<content>", "event": "ADD|UPDATE|DELETE|NONE",}\\
\texttt{"old\_memory": "<old content>"\}]\}}\\
(\texttt{old\_memory} required only when \texttt{event = "UPDATE"})\\[2pt]
Do not return anything except the JSON.\\
\bottomrule[1.1pt]
\end{tabular}
\end{table}

\begin{table}[h]
\centering
\caption{\textbf{Prompt for Mem0 Step~3: procedural summary generation in ExpSuite-Static} (applied to successful rollouts only). The LLM produces a structured, step-by-step summary of the agent's execution history.}
\label{tab:mem0-reflect}
\small
\begin{tabular}{p{13cm}}
\toprule[1.1pt]
\textit{system}\\
You are a memory summarization system that records and preserves the complete interaction history between a human and an AI agent. You are provided with the agent's execution history. Your task is to produce a comprehensive summary that contains every detail necessary for the agent to continue the task without ambiguity.\\[4pt]
Structure your output as follows:\\
\textbf{Overview}: Task objective and progress status (completion percentage and milestones).\\[2pt]
\textbf{Sequential Agent Actions} (numbered steps): For each step include (1) the precise agent action and all parameters, (2) the exact, unaltered action result verbatim, and (3) embedded metadata: key findings, navigation history, errors and challenges, and current context.\\[4pt]
Guidelines: preserve every output verbatim; use chronological order; include exact URLs, element indexes, error messages, and JSON responses. Output only the structured summary with no additional commentary.\\[4pt]
\textit{user}\\
Problem: \textbf{\{question\}}\\
Solution: \textbf{\{response\}}\\
Outcome: SUCCESS\\
\bottomrule[1.1pt]
\end{tabular}
\end{table}

\begin{table}[h]
\centering
\caption{\textbf{Prompt for Mem0 extracting a memory item in ALFWorld.}}
\label{tab:mem0-alfworld}
\small
\begin{tabular}{p{13cm}}
\toprule[1.1pt]
You are a Personal Information Organizer specialised in storing facts about an agent inside a text-based ALFWorld environment. Read the recent action--observation log and extract all atomic facts of the form ``\{subject\} \{predicate\} \{object\}''. Examples include: object locations (``soapbottle 1 is in cabinet 4''), task progress (``user has cleaned the dishsponge''), and reusable verbs (``cool $X$ requires opening fridge 1 first''). \\
 \\
Today's date is \textbf{\{today\}}. Return a JSON object: \\
\texttt{\{"facts": ["\{fact 1\}", "\{fact 2\}", \ldots]\}} \\
 \\
{[\ldots verbatim few-shot examples omitted \ldots]} \\
 \\
Conversation: \\
\textbf{\{turns\}} \\
\bottomrule[1.1pt]
\end{tabular}
\end{table}

\begin{table}[h]
\centering
\caption{\textbf{Prompt for Mem0 extracting a memory item in AppWorld.}}
\label{tab:mem0-appworld}
\small
\begin{tabular}{p{13cm}}
\toprule[1.1pt]
You are a Personal Information Organizer specialised in storing facts about a user's apps and AppWorld task history. Read the recent code-and-response log and extract all atomic facts of the form ``\{subject\} \{predicate\} \{object\}''. Examples: tokens (``user has venmo access\_token''), entities (``simple\_note id 3084 is the songs note''), API patterns (``\texttt{apis.spotify.update\_playlist} accepts \texttt{title} keyword''). \\
 \\
Today's date is \textbf{\{today\}}. Return a JSON object: \\
\texttt{\{"facts": ["\{fact 1\}", "\{fact 2\}", \ldots]\}} \\
 \\
{[\ldots verbatim few-shot examples omitted \ldots]} \\
 \\
Conversation: \\
\textbf{\{turns\}} \\
\bottomrule[1.1pt]
\end{tabular}
\end{table}

\begin{table}[h]
\centering
\caption{\textbf{Prompt for AWM workflow induction in ExpSuite-Static.} Applied once per dataset to a pool of up to $N=5$ successful trajectories. \textbf{\{examples\}} is a block of \texttt{Problem: ... / Solution: ...} entries; the induced workflow is stored as a plain string and prepended verbatim at inference time.}
\label{tab:awm-induction}
\small
\begin{tabular}{p{13cm}}
\toprule[1.1pt]
\textit{user}\\
Given a list of solved reasoning problems from the same category, your task is to extract the common workflows to solve these problems.\\[4pt]
Each given problem contains a problem statement and a solution. You need to find the repetitive reasoning patterns across multiple problems, and extract each as a workflow.\\[4pt]
Each workflow should be a commonly-reused reasoning strategy. Do not generate similar or overlapping workflows. Each workflow should have at least two steps. Represent problem-specific values with descriptive \textbf{\{placeholder\}} names.\\[4pt]
Try to generate as many workflows as needed to cover all the problems in the input list.\\[4pt]
\texttt{\#\# Concrete Examples}\\
\textbf{\{examples\}}\\[4pt]
\texttt{\#\# Summary Workflows}\\
\bottomrule[1.1pt]
\end{tabular}
\end{table}

\begin{table}[h]
\centering
\caption{\textbf{Prompt for AWM extracting a memory item in ALFWorld.}}
\label{tab:awm-alfworld}
\small
\begin{tabular}{p{13cm}}
\toprule[1.1pt]
You are an Agent Workflow Memory inducer. You are given up to $N{=}5$ successful ALFWorld trajectories that share the same task type (e.g.\ \texttt{pick\_clean\_then\_place\_in\_recep}). Extract the repetitive reasoning pattern as a step-by-step \emph{workflow} with concrete-value placeholders. \\
 \\
Output format: \\
\texttt{Workflow for \{task\_type\}:} \\
\texttt{Step 1: \{instruction with \{object\} and \{recep\} placeholders\}} \\
\texttt{Step 2: \ldots} \\
\texttt{Step k: \{final placement step\}} \\
 \\
Each workflow should have at least two steps. Do not generate similar or overlapping workflows; represent task-specific values with descriptive \texttt{\{placeholder\}} names. \\
 \\
Successful trajectories: \\
\textbf{\{trajectories\}} \\
\bottomrule[1.1pt]
\end{tabular}
\end{table}

\begin{table}[h]
\centering
\caption{\textbf{Prompt for AWM extracting a memory item in AppWorld.}}
\label{tab:awm-appworld}
\small
\begin{tabular}{p{13cm}}
\toprule[1.1pt]
You are an Agent Workflow Memory inducer. You are given up to $N{=}5$ successful AppWorld trajectories that share the same task type. Extract the repetitive reasoning pattern as a step-by-step \emph{workflow} with concrete-value placeholders. \\
 \\
Output format: \\
\texttt{Workflow for \{task\_type\}:} \\
\texttt{Step 1: \{API call template with \{access\_token\}, \{user\_email\}, \{note\_id\} placeholders\}} \\
\texttt{Step 2: \ldots} \\
\texttt{Step k: \texttt{apis.supervisor.complete\_task(answer=\{result\})}} \\
 \\
Each workflow should have at least two steps. Substitute concrete IDs / emails / strings with \texttt{\{placeholder\}} names. \\
 \\
Successful trajectories: \\
\textbf{\{trajectories\}} \\
\bottomrule[1.1pt]
\end{tabular}
\end{table}

\begin{table}[h]
\centering
\caption{\textbf{Summarizer prompt for \method extracting a memory item from a successful trajectory in ExpSuite-Static.}}
\label{tab:expgraph-static-sucess}
\small
\begin{tabular}{p{13cm}}
\toprule[1.1pt]
\texttt{<|im\_start|>}\\
\textit{system}\\
You are an expert at distilling problem-solving experiences into concise, reusable lessons.
Be brief and generalizable.\\
\texttt{<|im\_end|>}\\
\texttt{<|im\_start|>}\\
\textit{user}\\
A search/QA question was answered correctly.\\[4pt]
Question: \textbf{\{question\}}\\
Thought: \textbf{\{thought\}}\\
Executor answer: \textbf{\{answer\}}\\[4pt]
Extract a reusable search skill (2--3 sentences):\\
- \texttt{planning\_pattern}: Abstract the logic using generic terms \texttt{[Entity]}, \texttt{[Attribute]}, \texttt{[Time\_Period]}\\
- What key strategy led to success?\\[4pt]
Do NOT include specific names, numbers, or answers.
Focus on the transferable strategy.\\[4pt]
Output format: \texttt{SKILL: [your skill text]}\\
\texttt{<|im\_end|>}\\
\bottomrule[1.1pt]
\end{tabular}
\end{table}

\begin{table}[h]
\centering
\caption{\textbf{Summarizer prompt for \method extracting a memory item from a failed trajectory in ExpSuite-Static.}}
\label{tab:expgraph-static-failure}
\small
\begin{tabular}{p{13cm}}
\toprule[1.1pt]
\texttt{<|im\_start|>}\\
\textit{system}\\
You are an expert at distilling problem-solving experiences into concise, reusable lessons.
Be brief and generalizable.\\
\texttt{<|im\_end|>}\\
\texttt{<|im\_start|>}\\
\textit{user}\\
A search/QA question was answered incorrectly.\\[4pt]
Question: \textbf{\{question\}}\\
Thought: \textbf{\{thought\}}\\
Incorrect executor answer: \textbf{\{answer\}}\\[4pt]
Extract a reusable lesson (2--3 sentences):\\
- \texttt{trigger\_condition}: What kind of question caused the error?\\
- \texttt{bad\_action}: What went wrong?\\[4pt]
Do NOT include specific names.
Focus on the transferable lesson.\\[4pt]
Output format: \texttt{SKILL: [your lesson text]}\\
\texttt{<|im\_end|>}\\
\bottomrule[1.1pt]
\end{tabular}
\end{table}

\begin{table}[h]
\centering
\caption{\textbf{Summarizer prompt for \method extracting a memory item from a successful ALFWorld trajectory.}}
\label{tab:expgraph-alfworld-success}
\small
\begin{tabular}{p{13cm}}
\toprule[1.1pt]
\texttt{<|im\_start|>}\\
\textit{system}\\
You are an expert at analyzing household robot trajectories.
Extract specific, actionable lessons from the provided trajectory.\\
\textit{user}\\
An ALFWorld household task was completed successfully.\\[4pt]
Task: \textbf{\{task\}}\\
Full trajectory (action $\to$ observation): \textbf{\{trajectory\}}\\[4pt]
Extract a reusable skill (3--5 sentences). Include:\\
1. The general task category (\texttt{pick\_and\_place}, \texttt{heat\_then\_place}, \texttt{clean\_then\_place}, \texttt{cool\_then\_place}, \texttt{examine\_in\_light}, \texttt{pick\_two})\\
2. The concrete step-by-step strategy that worked\\
3. Common locations where target objects are found (e.g.\ ``soapbar is usually on countertop, bathtubbasin, or shelf'')\\[4pt]
Be specific. Use actual object/location types (\texttt{countertop}, \texttt{sinkbasin}, \texttt{microwave}).\\[4pt]
Output format: \texttt{SKILL: [your skill text]}\\
\texttt{<|im\_end|>}\\
\bottomrule[1.1pt]
\end{tabular}
\end{table}

\begin{table}[h]
\centering
\caption{\textbf{Summarizer prompt for \method extracting a memory item from a failed ALFWorld trajectory.}}
\label{tab:expgraph-alfworld-failure}
\small
\begin{tabular}{p{13cm}}
\toprule[1.1pt]
\texttt{<|im\_start|>}\\
\textit{system}\\
You are an expert at analyzing household robot trajectories.
Extract specific, actionable lessons from the provided trajectory.\\
\textit{user}\\
An ALFWorld household task failed after all steps.\\[4pt]
Task: \textbf{\{task\}}\\
Trajectory: \textbf{\{trajectory\}}\\[4pt]
Extract a reusable lesson (3--5 sentences). Include:\\
1. The general task category\\
2. What specific mistake was made\\
3. What the agent should have done differently\\[4pt]
Output format: \texttt{SKILL: [your lesson text]}\\
\texttt{<|im\_end|>}\\
\bottomrule[1.1pt]
\end{tabular}
\end{table}

\begin{table}[h]
\centering
\caption{\textbf{Summarizer prompt for \method extracting a memory item from an AppWorld episode.} \textbf{\{STATUS\}} is replaced by \texttt{SUCCESSFUL}, \texttt{FAILED}, or \texttt{PARTIAL}.}
\label{tab:expgraph-appworld}
\small
\begin{tabular}{p{13cm}}
\toprule[1.1pt]
\texttt{<|im\_start|>}\\
\textit{user}\\
Analyze this \textbf{\{STATUS\}} AppWorld code generation episode.\\[4pt]
Task: \textbf{\{task\}}\\
Trajectory: \textbf{\{trajectory\}}\\[4pt]
Extract ONE concise, actionable skill. Focus on SPECIFIC API details, not generic advice.\\[4pt]
Format: \texttt{[Short Title]} Specific principle. When: trigger condition.\\[4pt]
Good examples:\\
- \texttt{[Spotify Login Fields]} \texttt{apis.spotify.login()} requires \texttt{email} and \texttt{password} as keyword arguments\ldots\\
- \texttt{[Venmo Transactions Return Dict]} \texttt{apis.venmo.list\_transactions()} returns a dict with key \texttt{"transactions"}\ldots\\
- \texttt{[File System Read Path]} Use \texttt{apis.file\_system.read\_file(file\_path=...)} with absolute paths\ldots\\[4pt]
BAD examples (do NOT generate):\\
- ``Always inspect API docs first'' --- too generic\\
- ``Verify API method names'' --- too vague\\[4pt]
Rules: Include SPECIFIC app names, API method names, parameter names, return value structures.\\[4pt]
Output ONLY the skill line.\\
\texttt{<|im\_end|>}\\
\bottomrule[1.1pt]
\end{tabular}
\end{table}

\begin{table}[h]
\centering
\caption{\textbf{Executor prompt for No Memory in Question Answering.}}
\label{tab:cot-qa}
\small
\begin{tabular}{p{13cm}}
\toprule[1.1pt]
\texttt{<|im\_start|>}\\
\textit{system}\\
You are a helpful assistant skilled in mathematics, logical reasoning, and programming.
Solve problems step by step, showing your work clearly.\\[4pt]
CRITICAL FORMAT REQUIREMENT:\\
- You MUST end your response with your final answer wrapped in \texttt{<answer> </answer>} tags.\\
- ALWAYS use \texttt{<answer> </answer>} tags.\\
- Example: \texttt{<answer> A </answer>}\\
\texttt{<|im\_end|>}\\
\texttt{<|im\_start|>}\\
\textit{user}\\
Answer this question:\\[2pt]
Question: \textbf{\{question\}}\\[2pt]
\textbf{\{choices\}}\\[2pt]
Analyze each option and explain your reasoning.\\[2pt]
Write your final answer as: \texttt{<answer> [letter] </answer>}\\
\texttt{<|im\_end|>}\\
\bottomrule[1.1pt]
\end{tabular}
\end{table}

\begin{table}[h]
\centering
\caption{\textbf{Executor prompt for No Memory in Math Reasoning.}}
\label{tab:cot-math}
\small
\begin{tabular}{p{13cm}}
\toprule[1.1pt]
\texttt{<|im\_start|>}\\
\textit{system}\\
You are a helpful assistant skilled in mathematics, logical reasoning, and programming.
Solve problems step by step, showing your work clearly.\\[4pt]
CRITICAL FORMAT REQUIREMENT:\\
- You MUST end your response with \texttt{\#\#\#\#} followed by your final answer.\\
- NEVER use \textbackslash boxed\{\} or any \LaTeX{} boxing format.\\
- ALWAYS use \texttt{\#\#\#\#} format, even for mathematical expressions.\\
- Example: \texttt{\#\#\#\# 42} or \texttt{\#\#\#\# x\^{}2 + 1}\\
\texttt{<|im\_end|>}\\
\texttt{<|im\_start|>}\\
\textit{user}\\
Solve this math word problem:\\[2pt]
\textbf{\{question\}}\\[2pt]
Think step by step, showing your calculations.\\[2pt]
Write your final answer as: \texttt{\#\#\#\# [number]}\\
\texttt{<|im\_end|>}\\
\bottomrule[1.1pt]
\end{tabular}
\end{table}

\begin{table}[h]
\centering
\caption{\textbf{Executor prompt for No Memory in Code Generation.}}
\label{tab:cot-code}
\small
\begin{tabular}{p{13cm}}
\toprule[1.1pt]
\texttt{<|im\_start|>}\\
\textit{system}\\
You are an expert Python programmer. Solve coding problems step by step.\\[4pt]
CRITICAL FORMAT REQUIREMENT:\\
- Provide your final solution in a \texttt{```python} code block.\\
- Example format: \texttt{```python} \textbackslash n \texttt{def solution(): pass} \textbackslash n \texttt{```}\\
\texttt{<|im\_end|>}\\
\texttt{<|im\_start|>}\\
\textit{user}\\
Task: \textbf{\{task\}}\\[2pt]
Your code should pass these tests:\\
\textbf{\{test\_cases\}}\\[2pt]
Think through your approach, then write the Python function.\\
REMEMBER: Provide your code in a \texttt{```python} code block.\\
\texttt{<|im\_end|>}\\
\bottomrule[1.1pt]
\end{tabular}
\end{table}

\begin{table}[h]
\centering
\caption{\textbf{Executor prompt for No Memory in ALFWorld.}}
\label{tab:cot-alfworld}
\small
\begin{tabular}{p{13cm}}
\toprule[1.1pt]
{[SYSTEM]} You are controlling a text-based ALFWorld environment. Choose the NEXT action as ONE admissible command string. Output only the command, copied verbatim from the admissible list. \\
 \\
{[USER]} Task: \textbf{\{objective\}} \\
 \\
Interaction history so far: \textbf{\{history\}} \\
 \\
Current observation: \textbf{\{current\_obs\}} \\
 \\
Admissible actions: \textbf{\{admissible\}} \\
 \\
Action:
\\
\bottomrule[1.1pt]
\end{tabular}
\end{table}

\begin{table}[h]
\centering
\caption{\textbf{Executor prompt for No Memory in AppWorld.}}
\label{tab:cot-appworld}
\small
\begin{tabular}{p{13cm}}
\toprule[1.1pt]
{[SYSTEM]} You are an autonomous AppWorld agent. You complete each task by writing Python code that calls \texttt{apis.<app>.<endpoint>(\ldots)} and finally \texttt{apis.supervisor.complete\_task(answer=\ldots)}. Wrap each turn's code in triple-backtick fences. \\
 \\
{[USER]} Instruction: \textbf{\{instruction\}} \\
 \\
Available apps and their docs: \textbf{\{api\_docs\_summary\}} \\
 \\
Recent execution history: \textbf{\{history\}} \\
 \\
Reason step by step, then write the next code block.
\\
\bottomrule[1.1pt]
\end{tabular}
\end{table}

\begin{table}[h]
\centering
\caption{\textbf{Executor prompt for experience-conditioned baselines (ReasoningBank, ExpeL, LightMem, Mem0, AWM, IRCoT) in Question Answering.} \textbf{\{retrieved\_memories\}} is replaced by the method-specific memory context.}
\label{tab:retrieval-qa}
\small
\begin{tabular}{p{13cm}}
\toprule[1.1pt]
\texttt{<|im\_start|>}\\
\textit{system}\\
You are a helpful assistant skilled in mathematics, logical reasoning, and programming.
Solve problems step by step, showing your work clearly.\\[4pt]
CRITICAL FORMAT REQUIREMENT:\\
- You MUST end your response with your final answer wrapped in \texttt{<answer> </answer>} tags.\\
- ALWAYS use \texttt{<answer> </answer>} tags.\\
- Example: \texttt{<answer> A </answer>}\\
\texttt{<|im\_end|>}\\
\texttt{<|im\_start|>}\\
\textit{user}\\
\textbf{\{retrieved\_memories\}}\\[2pt]
Answer this question:\\[2pt]
Question: \textbf{\{question\}}\\[2pt]
\textbf{\{choices\}}\\[2pt]
Analyze each option and explain your reasoning.\\[2pt]
Write your final answer as: \texttt{<answer> [letter] </answer>}\\
\texttt{<|im\_end|>}\\
\bottomrule[1.1pt]
\end{tabular}
\end{table}

\begin{table}[h]
\centering
\caption{\textbf{Executor prompt for experience-conditioned baselines in Math Reasoning.}}
\label{tab:retrieval-math}
\small
\begin{tabular}{p{13cm}}
\toprule[1.1pt]
\texttt{<|im\_start|>}\\
\textit{system}\\
You are a helpful assistant skilled in mathematics, logical reasoning, and programming.
Solve problems step by step, showing your work clearly.\\[4pt]
CRITICAL FORMAT REQUIREMENT:\\
- You MUST end your response with \texttt{\#\#\#\#} followed by your final answer.\\
- NEVER use \textbackslash boxed\{\} or any \LaTeX{} boxing format.\\
- ALWAYS use \texttt{\#\#\#\#} format, even for mathematical expressions.\\
- Example: \texttt{\#\#\#\# 42} or \texttt{\#\#\#\# x\^{}2 + 1}\\
\texttt{<|im\_end|>}\\
\texttt{<|im\_start|>}\\
\textit{user}\\
\textbf{\{retrieved\_memories\}}\\[2pt]
Solve this math word problem:\\[2pt]
\textbf{\{question\}}\\[2pt]
Think step by step, showing your calculations.\\[2pt]
Write your final answer as: \texttt{\#\#\#\# [number]}\\
\texttt{<|im\_end|>}\\
\bottomrule[1.1pt]
\end{tabular}
\end{table}

\begin{table}[h]
\centering
\caption{\textbf{Executor prompt for experience-conditioned baselines in Code Generation.}}
\label{tab:retrieval-code}
\small
\begin{tabular}{p{13cm}}
\toprule[1.1pt]
\texttt{<|im\_start|>}\\
\textit{system}\\
You are an expert Python programmer. Solve coding problems step by step.\\[4pt]
CRITICAL FORMAT REQUIREMENT:\\
- Provide your final solution in a \texttt{```python} code block.\\
- Example format: \texttt{```python} \textbackslash n \texttt{def solution(): pass} \textbackslash n \texttt{```}\\
\texttt{<|im\_end|>}\\
\texttt{<|im\_start|>}\\
\textit{user}\\
\textbf{\{retrieved\_memories\}}\\[2pt]
Task: \textbf{\{task\}}\\[2pt]
Your code should pass these tests:\\
\textbf{\{test\_cases\}}\\[2pt]
Think through your approach, then write the Python function.\\
REMEMBER: Provide your code in a \texttt{```python} code block.\\
\texttt{<|im\_end|>}\\
\bottomrule[1.1pt]
\end{tabular}
\end{table}

\begin{table}[h]
\centering
\caption{\textbf{Executor prompt for experience-conditioned baselines in ALFWorld.}}
\label{tab:retrieval-alfworld}
\small
\begin{tabular}{p{13cm}}
\toprule[1.1pt]
{[SYSTEM]} You are controlling a text-based ALFWorld environment. Choose the NEXT action as ONE admissible command string. Output only the command, copied verbatim from the admissible list. \\
 \\
{[USER]} Task: \textbf{\{objective\}} \\
 \\
\textbf{\{retrieved\_memories\}} \\
 \\
Interaction history so far: \textbf{\{history\}} \\
 \\
Current observation: \textbf{\{current\_obs\}} \\
 \\
Admissible actions: \textbf{\{admissible\}} \\
 \\
Action:
\\
\bottomrule[1.1pt]
\end{tabular}
\end{table}

\begin{table}[h]
\centering
\caption{\textbf{Executor prompt for experience-conditioned baselines in AppWorld.}}
\label{tab:retrieval-appworld}
\small
\begin{tabular}{p{13cm}}
\toprule[1.1pt]
{[SYSTEM]} You are an autonomous AppWorld agent. You complete each task by writing Python code that calls \texttt{apis.<app>.<endpoint>(\ldots)} and finally \texttt{apis.supervisor.complete\_task(answer=\ldots)}. Wrap each turn's code in triple-backtick fences. \\
 \\
{[USER]} Instruction: \textbf{\{instruction\}} \\
 \\
\textbf{\{retrieved\_memories\}} \\
 \\
Available apps and their docs: \textbf{\{api\_docs\_summary\}} \\
 \\
Recent execution history: \textbf{\{history\}} \\
 \\
Reason step by step, then write the next code block.
\\
\bottomrule[1.1pt]
\end{tabular}
\end{table}

\begin{table}[h]
\centering
\caption{\textbf{Prompt for summarizing retrieved memories in Search-o1.}}
\label{tab:search-o1-summarize}
\small
\begin{tabular}{p{13cm}}
\toprule[1.1pt]
\texttt{<|im\_start|>}\\
\textit{user}\\
\textbf{Task Instruction:}\\[4pt]
You are tasked with reading and analyzing memories based on the following inputs: \textbf{Previous Reasoning Steps}, \textbf{Current Search Query}, and \textbf{Searched Memories}.
Your objective is to extract relevant and helpful information for the \textbf{Current Search Query} from the \textbf{Searched Memories} and seamlessly integrate this information into the \textbf{Previous Reasoning Steps} to continue reasoning for the original question.\\[4pt]
\textbf{Guidelines:}\\
1. Carefully review the content of each searched memory. Identify factual information that is relevant to the Current Search Query.\\
2. Select the information that directly contributes to advancing the Previous Reasoning Steps. Ensure that the extracted information is accurate and relevant.\\[4pt]
\textbf{Output Format:}\\
- If the memories provide helpful information: present the information beginning with \texttt{**Final Information**}\\[2pt]
\texttt{**Final Information**}\\
{[Helpful information]}\\
- If the memories do not provide helpful information:\\
\texttt{**Final Information**}\\
No helpful information found.\\[4pt]
\textbf{Inputs:}\\
- Previous Reasoning Steps: \textbf{\{prev\_reasoning\}}\\
- Current Search Query: \textbf{\{search\_query\}}\\
- Searched Memories: \textbf{\{retrieved\_memories\}}\\
\texttt{<|im\_end|>}\\
\bottomrule[1.1pt]
\end{tabular}
\end{table}

\begin{table}[h]
\centering
\caption{\textbf{Task-solving prompt for Search-o1 in Question Answering.}}
\label{tab:searcho1-qa}
\small
\begin{tabular}{p{13cm}}
\toprule[1.1pt]
\texttt{<|im\_start|>}\\
\textit{system}\\
You are a helpful assistant skilled in mathematics, logical reasoning, and programming.
You have the ability to perform memory searches to help you answer the user's question accurately.\\
To perform a search: write \texttt{<|begin\_search\_query|>} your query here \texttt{<|end\_search\_query|>}.\\
The system will then search and analyze relevant memories, returning helpful information in the format \texttt{<|begin\_search\_result|>} ...results... \texttt{<|end\_search\_result|>}.\\
Once you have all the information you need, continue your reasoning.\\[4pt]
CRITICAL FORMAT REQUIREMENT:\\
- You MUST end your response with your final answer wrapped in \texttt{<answer> </answer>} tags.\\
- ALWAYS use \texttt{<answer> </answer>} tags.\\
- Example: \texttt{<answer> A </answer>}\\
\texttt{<|im\_end|>}\\
\texttt{<|im\_start|>}\\
\textit{user}\\
Answer this question:\\[2pt]
Question: \textbf{\{question\}}\\[2pt]
\textbf{\{choices\}}\\[2pt]
Analyze each option and explain your reasoning.\\[2pt]
Write your final answer as: \texttt{<answer> [letter] </answer>}\\
\texttt{<|im\_end|>}\\
\bottomrule[1.1pt]
\end{tabular}
\end{table}

\begin{table}[h]
\centering
\caption{\textbf{Task-solving prompt for Search-o1 in Math Reasoning.}}
\label{tab:searcho1-math}
\small
\begin{tabular}{p{13cm}}
\toprule[1.1pt]
\texttt{<|im\_start|>}\\
\textit{system}\\
You are a helpful assistant skilled in mathematics, logical reasoning, and programming.
You have the ability to perform memory searches to help you answer the user's question accurately.\\
To perform a search: write \texttt{<|begin\_search\_query|>} your query here \texttt{<|end\_search\_query|>}.\\
The system will return helpful information in the format \texttt{<|begin\_search\_result|>} ...results... \texttt{<|end\_search\_result|>}.\\[4pt]
CRITICAL FORMAT REQUIREMENT:\\
- You MUST end your response with \texttt{\#\#\#\#} followed by your final answer.\\
- NEVER use \textbackslash boxed\{\} or any \LaTeX{} boxing format.\\
- Example: \texttt{\#\#\#\# 42}\\
\texttt{<|im\_end|>}\\
\texttt{<|im\_start|>}\\
\textit{user}\\
Solve this math word problem:\\[2pt]
\textbf{\{question\}}\\[2pt]
Think step by step, showing your calculations.\\[2pt]
Write your final answer as: \texttt{\#\#\#\# [number]}\\
\texttt{<|im\_end|>}\\
\bottomrule[1.1pt]
\end{tabular}
\end{table}

\begin{table}[h]
\centering
\caption{\textbf{Task-solving prompt for Search-o1 in Code Generation.}}
\label{tab:searcho1-code}
\small
\begin{tabular}{p{13cm}}
\toprule[1.1pt]
\texttt{<|im\_start|>}\\
\textit{system}\\
You are an expert Python programmer.
You have the ability to perform memory searches to help you answer the user's question accurately.\\
To perform a search: write \texttt{<|begin\_search\_query|>} your query here \texttt{<|end\_search\_query|>}.\\
The system will return helpful information in the format \texttt{<|begin\_search\_result|>} ...results... \texttt{<|end\_search\_result|>}.\\[4pt]
CRITICAL FORMAT REQUIREMENT:\\
- Provide your final solution in a \texttt{```python} code block.\\
\texttt{<|im\_end|>}\\
\texttt{<|im\_start|>}\\
\textit{user}\\
Task: \textbf{\{task\}}\\[2pt]
Your code should pass these tests:\\
\textbf{\{test\_cases\}}\\[2pt]
Think through your approach, then write the Python function.\\
REMEMBER: Provide your code in a \texttt{```python} code block.\\
\texttt{<|im\_end|>}\\
\bottomrule[1.1pt]
\end{tabular}
\end{table}

\begin{table}[h]
\centering
\caption{\textbf{Task-solving prompt for Search-o1 in ALFWorld.}}
\label{tab:search-o1-alfworld}
\small
\begin{tabular}{p{13cm}}
\toprule[1.1pt]
{[SYSTEM]} You are controlling a text-based ALFWorld environment. You may interleave reasoning with explicit memory queries. To query the experience corpus, emit:\ \texttt{$\langle$|begin\_search\_query|$\rangle$\{your sub-question\}$\langle$|end\_search\_query|$\rangle$}. The system will return summarised search results inside \texttt{$\langle$|begin\_search\_result|$\rangle$\ldots$\langle$|end\_search\_result|$\rangle$}. After at most $S{=}3$ rounds, output the next admissible action. \\
 \\
{[USER]} Task: \textbf{\{objective\}} \\
Interaction history: \textbf{\{history\}} \\
Current observation: \textbf{\{current\_obs\}} \\
Admissible actions: \textbf{\{admissible\}} \\
 \\
Reason and (optionally) search; then output Action: \texttt{$\langle$cmd$\rangle$}.
\\
\bottomrule[1.1pt]
\end{tabular}
\end{table}

\begin{table}[h]
\centering
\caption{\textbf{Task-solving prompt for Search-o1 in AppWorld.}}
\label{tab:search-o1-appworld}
\small
\begin{tabular}{p{13cm}}
\toprule[1.1pt]
{[SYSTEM]} You are an autonomous AppWorld agent. You may interleave reasoning with explicit memory queries. To query the experience corpus, emit:\ \texttt{$\langle$|begin\_search\_query|$\rangle$\{your sub-question\}$\langle$|end\_search\_query|$\rangle$}. The system will return summarised search results inside \texttt{$\langle$|begin\_search\_result|$\rangle$\ldots$\langle$|end\_search\_result|$\rangle$}. After at most $S{=}3$ rounds, output the next code block. \\
 \\
{[USER]} Instruction: \textbf{\{instruction\}} \\
Available apps and docs: \textbf{\{api\_docs\_summary\}} \\
Recent execution history: \textbf{\{history\}} \\
 \\
Reason and (optionally) search; then write the next code block in triple-backtick fences.
\\
\bottomrule[1.1pt]
\end{tabular}
\end{table}

\clearpage

\section{Experimental Result Analysis}
\label{app:exp}

\subsection{\method Outperforms General Prompt-based Baselines and Experience Learning Methods}
\label{app:main_exp}

We evaluate \method on ExpSuite, covering both single-turn static tasks and multi-step agentic environments.
Results are reported in Table~\ref{tab:standardized_benchmarks} and Table~\ref{tab:expsuite_agentic}.
We have the following observations.

\xhdr{\method Achieves the Highest Weighted Average Scores in the Evaluated Settings}
On ExpSuite-Static, \method improves the test-instance-weighted average score over
MemRL, the strongest baseline under this aggregation, by 12.1\% with the small
executor and 4.5\% with the large executor. These are relative improvements,
computed as $(S_{\mathrm{ours}}-S_{\mathrm{baseline}})/S_{\mathrm{baseline}}$;
they do not imply the best result on every individual benchmark.
On ExpSuite-Agentic, the corresponding relative improvements over S3 are 21.4\%
and 12.7\%. Average environment interaction steps are 12.7\% and 21.6\% lower
than those of the baselines with the lowest average step counts, respectively.
Step counts describe environment interaction and do not include all retrieval,
LLM inference, or training costs; they can also vary with task success.

\xhdr{Relative Gains Vary Across Executors and Task Settings}
In these evaluations, the relative gains are larger for the executor labeled
small within each setting, and larger in ExpSuite-Agentic than in ExpSuite-Static.
These comparisons are descriptive: executor identities, benchmarks, and baseline
scores differ across settings. They are consistent with useful experience reuse
in interactive tasks, but do not isolate model scale or task horizon as the
cause of the differences.

\subsection{Harness Transfer Across Different LLM Executors}
\label{app:executor_generalization}

We evaluate executor transfer under three settings.
\textit{Small-to-large transfer} uses the small executor as the source and the large executor as the target, testing whether experience components learned with cheaper models can benefit stronger frozen LLMs.
\textit{Large-to-small transfer} uses the large executor as the source and the small executor as the target, testing whether high-quality experiences learned from stronger models can still be exploited by weaker models.
\textit{Non-reasoning-to-reasoning transfer} uses Llama-3.1-8B-Instruct as
the source and DeepSeek-R1-Distill-Llama-8B as the target for ExpSuite-Static,
and Gemini-3.1-Flash-Lite as the source and Claude-Sonnet-4 as the target for
ExpSuite-Agentic, matching \S\ref{sec:executor_generalization}.
For each setting, we compare Graph-only Transfer, Copilot-only Transfer, Graph+Copilot Transfer, and target-specific \method.

\xhdr{Joint Transfer Is Closest to Target-Specific Training in the Small-to-Large Setting}
In Figure~\ref{fig:executor_transfer_radar}(a), Graph+Copilot Transfer is closest
to target-specific \method across the displayed domains. It also outperforms
the two partial-transfer variants, supporting joint reuse of the experience
store and retrieval policy in this setting.

\xhdr{Large-to-Small Transfer Retains a Gap to Target-Specific Training}
In Figure~\ref{fig:executor_transfer_radar}(b), the transferred variants remain
below target-specific \method, while Graph+Copilot Transfer performs best among
the transfer variants across most displayed domains. The panels use different
axis ranges, so their polygon areas cannot quantify the relative difficulty of
the transfer directions. Without a target No-Memory reference in this figure,
these comparisons alone do not establish improvement over the unaugmented
target executor.

\xhdr{Joint Transfer Also Performs Well for Reasoning-Oriented Targets}
In Figure~\ref{fig:executor_transfer_radar}(c), Graph+Copilot Transfer approaches
target-specific \method, including on ALFWorld and AppWorld. This observation
is consistent with the reuse of experience across the evaluated model pairs.
Because model identity and reasoning capability change together, the experiment
does not isolate reasoning capability as the cause of transfer performance.

\subsection{Analysis of the Four Ablation Settings}
\label{app:ablation}

To examine four experience-retrieval design choices in \method, we conduct ablation studies across five evaluation domains: QA, Reasoning, Coding, ALFWorld, and AppWorld.
These variants examine the effects of similarity-aware experience management, graph-structured experience organization, graph diffusion, and utility-aware ranking.
Results are reported in Figure~\ref{fig:ablation_studies}.

\begin{itemize}[leftmargin=1em, itemsep=0pt, topsep=0pt, parsep=0pt, partopsep=0pt]
    \item \textbf{w/o Similarity Filtering}: Removes similarity-based filtering during experience insertion, allowing redundant experiences to be repeatedly retained.

    \item \textbf{Flat Experience}: Replaces the experience graph with a flat experience pool and retrieves top-$K$ experiences purely by semantic similarity.

    \item \textbf{w/o Graph Diffusion}: Keeps the graph structure but disables graph expansion, retrieving only from the initial semantic seed nodes.

    \item \textbf{w/o Utility Ranking}: Ranks candidate experiences only by semantic similarity, without using historical utility statistics.
\end{itemize}

Across the reported domains, each of these four variants performs below the
full method. \textbf{Flat Experience} shows the largest drop, but it changes both
candidate generation and the ranking rule, so its gap cannot be attributed to
relational structure alone. \textbf{w/o Graph Diffusion} and \textbf{w/o Utility
Ranking} show lower scores under their respective retrieval changes, while
\textbf{w/o Similarity Filtering} shows a smaller decline. These results support
the combined retrieval design under the tested configurations. They do not
separately isolate the effects of copilot training, the reward terms, or the
node-utility update rule.

\section{Case Studies}
\label{app:case_studies}

This appendix presents examples across ExpSuite-Static (QA, mathematical
reasoning, and code generation) and ExpSuite-Agentic. For each method, the
tables show task inputs, reference outcomes, experience context where applicable,
and abbreviated task-solving responses. For \method, final answers and actions
are produced by the frozen executor; retrieval controls belong to the copilot.
Ground-truth fields in these case-study tables are reference annotations, not
instructions for the copilot to generate answers.

No Memory case studies are in Table~\ref{tab:case-cot-qa}--\ref{tab:case-cot-appworld},
ReasoningBank in Table~\ref{tab:case-rag-qa}--\ref{tab:case-rag-appworld},
ExpeL in Table~\ref{tab:case-expl-qa}--\ref{tab:case-expl-alfworld},
LightMem in Table~\ref{tab:case-lightmem-qa}--\ref{tab:case-lightmem-appworld},
Mem0 in Table~\ref{tab:case-mem0-qa}--\ref{tab:case-mem0-appworld},
AWM in Table~\ref{tab:case-awm-qa}--\ref{tab:case-awm-appworld},
IRCoT in Table~\ref{tab:case-ircot-qa}--\ref{tab:case-ircot-appworld},
Search-o1 in Table~\ref{tab:case-searcho1-qa}--\ref{tab:case-searcho1-appworld}, 
MemRL in Table~\ref{tab:case-memrl-qa}--\ref{tab:case-memrl-appworld}.


\begin{table}[h]
    \centering
    \footnotesize
    \caption{\textbf{No Memory case study in QA.}}
    \label{tab:case-cot-qa}
    \begin{tabular}{p{13.5cm}}
        \toprule
        \textbf{Question}: A bowling ball with a mass of 8.0\,kg rolls down a bowling lane at 2.0\,m/s. What is the momentum of the bowling ball?\\
        A. 4.0 kg$\times$m/s \quad B. 6.0 kg$\times$m/s \quad C. 10.0 kg$\times$m/s \quad D. 16.0 kg$\times$m/s\\
        \midrule
        \textbf{Ground Truth}: D\\
        \midrule
        Using the formula $p = mv$: $p = 8.0\,\text{kg} \times 2.0\,\text{m/s} = 16.0\,\text{kg}\cdot\text{m/s}$.\\
        A (4.0) and B (6.0) and C (10.0) are all incorrect; D matches the calculated value.\\[2pt]
        \texttt{<answer> D </answer>}\\
        \bottomrule
    \end{tabular}
\end{table}

\begin{table}[h]
    \centering
    \footnotesize
    \caption{\textbf{No Memory case study in Math.}}
    \label{tab:case-cot-math}
    \begin{tabular}{p{13.5cm}}
        \toprule
        \textbf{Question}: Trevor is a comic book illustrator. In the past three months, he has drawn 220 pages. The comic releases a new issue once a month. The first and second months' issues were the same number of pages, but the third month's issue was four pages longer than either of the others. How many pages was the first issue?\\
        \midrule
        \textbf{Ground Truth}: 72\\
        \midrule
        Let the first and second issue each have $x$ pages; the third has $x+4$ pages.\\
        $x + x + (x+4) = 220 \;\Rightarrow\; 3x + 4 = 220 \;\Rightarrow\; 3x = 216 \;\Rightarrow\; x = 72$.\\[2pt]
        \texttt{\#\#\#\# 72}\\
        \bottomrule
    \end{tabular}
\end{table}

\begin{table}[h]
    \centering
    \footnotesize
    \caption{\textbf{No Memory case study in Code Generation.}}
    \label{tab:case-cot-code}
    \begin{tabular}{p{13.5cm}}
        \toprule
        \textbf{Question}: Return the largest prime factor of \texttt{n}. Assume $n > 1$ and is not a prime.\\
        Tests: \texttt{assert largest\_prime\_factor(15) == 5}, \texttt{assert largest\_prime\_factor(27) == 3}\\
        \midrule
        \textbf{Ground Truth}: Trial-division algorithm (divide by smallest prime factors iteratively).\\
        \midrule
        Divide \texttt{n} by the smallest prime (2) while divisible, then try odd divisors up to $\sqrt{n}$; the remaining value is the largest prime factor.\\[2pt]
        \texttt{\#\#\#\#}\\
        \texttt{```python}\\
        \texttt{def largest\_prime\_factor(n: int) -> int:}\\
        \texttt{\quad i = 2}\\
        \texttt{\quad while i * i <= n:}\\
        \texttt{\quad\quad if n \% i: i += 1}\\
        \texttt{\quad\quad else: n //= i}\\
        \texttt{\quad return n}\\
        \texttt{```}\\
        \bottomrule
    \end{tabular}
\end{table}

\begin{table}[h]
    \centering
    \footnotesize
    \caption{\textbf{No Memory case study in ALFWorld.}}
    \label{tab:case-cot-alfworld}
    \begin{tabular}{p{13.5cm}}
        \toprule
        \textbf{Task}: put some remotecontrol on armchair. \\
        \midrule
        \textbf{Outcome}: Score = 1.0,\ \ Steps = 4 \\
        \midrule
        \textbf{Retrieved Memories}: \textit{(none --- No-Memory baseline; the agent acts from prompt + admissible actions only.)} \\
        \midrule
        \textbf{Action Trace}: \\
        go to coffeetable 1 (\textbf{found remotecontrol 1, 2}) $\to$ take remotecontrol 1 $\to$ go to armchair 1 $\to$ \textbf{move remotecontrol 1 to armchair 1} $\checkmark$ \\
        \bottomrule
    \end{tabular}
\end{table}

\begin{table}[h]
    \centering
    \footnotesize
    \caption{\textbf{No Memory case study in AppWorld.}}
    \label{tab:case-cot-appworld}
    \begin{tabular}{p{13.5cm}}
        \toprule
        \textbf{Task}: How much money have I sent to others on Venmo this month so far? \\
        \midrule
        \textbf{Outcome}: Score = 0.50,\ \ Steps = 21 \\
        \midrule
        \textbf{Retrieved Memories}: \textit{(none --- No-Memory baseline.)} \\
        \midrule
        \textbf{Final Action}: \\
        \texttt{apis.supervisor.complete\_task(answer="433.0")} \\
        \bottomrule
    \end{tabular}
\end{table}

\clearpage


\begin{table}[h]
    \centering
    \footnotesize
    \caption{\textbf{ReasoningBank case study in QA.}}
    \label{tab:case-rag-qa}
    \begin{tabular}{p{13.5cm}}
        \toprule
        \textbf{Question}: A bowling ball with a mass of 8.0\,kg rolls down a bowling lane at 2.0\,m/s. What is the momentum of the bowling ball?\\
        A. 4.0 kg$\times$m/s \quad B. 6.0 kg$\times$m/s \quad C. 10.0 kg$\times$m/s \quad D. 16.0 kg$\times$m/s\\
        \midrule
        \textbf{Ground Truth}: D\\
        \midrule
        \textbf{Retrieved Memory 1 (\checkmark\ Success Strategy):}\\
        \textbf{Calculating Momentum} --- When calculating momentum, ensure to multiply the given mass and velocity accurately. This example highlights the importance of correctly applying the formula and performing the multiplication to avoid common mistakes.\\[2pt]
        \textbf{Retrieved Memory 2 (\ding{55}\ Lesson from Failure):}\\
        \textbf{Impulse and Momentum in Collisions} --- In a collision, the total change in momentum of the system is zero, but the lighter object experiences a greater change in momentum due to its lower mass.\\
        \midrule
        Using $p = mv$: $p = 8.0 \times 2.0 = 16.0\,\text{kg}\cdot\text{m/s}$. Options A, B, C are all incorrect. D matches the calculated value.\\[2pt]
        \texttt{<answer> D </answer>}\\
        \bottomrule
    \end{tabular}
\end{table}

\begin{table}[h]
    \centering
    \footnotesize
    \caption{\textbf{ReasoningBank case study in Math.}}
    \label{tab:case-rag-math}
    \begin{tabular}{p{13.5cm}}
        \toprule
        \textbf{Question}: Trevor is a comic book illustrator. In the past three months, he has drawn 220 pages. The comic releases one issue per month. The first and second months' issues were equal; the third was four pages longer. How many pages was the first issue?\\
        \midrule
        \textbf{Ground Truth}: 72\\
        \midrule
        \textbf{Retrieved Memory 1 (\ding{55}\ Lesson from Failure):}\\
        \textbf{Misinterpretation of the Problem} --- The mistake was in not accounting for all components described in the problem. Carefully read each condition before setting up equations.\\[2pt]
        \textbf{Retrieved Memory 2 (\ding{55}\ Lesson from Failure):}\\
        \textbf{Misinterpretation of the Problem} --- Always verify that all constraints are captured in the equation before solving.\\
        \midrule
        Let $x$ = pages in issue 1 and 2; issue 3 = $x+4$.\\
        $x + x + (x+4) = 220 \;\Rightarrow\; 3x = 216 \;\Rightarrow\; x = 72$.\\[2pt]
        \texttt{\#\#\#\# 72}\\
        \bottomrule
    \end{tabular}
\end{table}

\begin{table}[h]
    \centering
    \footnotesize
    \caption{\textbf{ReasoningBank case study in Code Generation.}}
    \label{tab:case-rag-code}
    \begin{tabular}{p{13.5cm}}
        \toprule
        \textbf{Question}: Return the largest prime factor of \texttt{n}. Assume $n > 1$ and is not a prime.\\
        Tests: \texttt{assert largest\_prime\_factor(15) == 5}, \texttt{assert largest\_prime\_factor(27) == 3}\\
        \midrule
        \textbf{Ground Truth}: Trial-division algorithm.\\
        \midrule
        \textbf{Retrieved Memory 1 (\checkmark\ Success Strategy):}\\
        \textbf{Finding the Largest Perfect Square Factor} --- Decompose the number into its prime factors; select the highest even power for each prime.\\[2pt]
        \textbf{Retrieved Memory 2 (\ding{55}\ Lesson from Failure):}\\
        \textbf{Efficient Prime Factorization} --- Always use the square root to limit the search space for prime factors, and handle the remaining factor after the loop.\\
        \midrule
        \texttt{\#\#\#\#}\\
        \texttt{```python}\\
        \texttt{def largest\_prime\_factor(n: int) -> int:}\\
        \texttt{\quad while n \% 2 == 0:}\\
        \texttt{\quad\quad max\_prime = 2; n //= 2}\\
        \texttt{\quad for i in range(3, int(n**0.5)+1, 2):}\\
        \texttt{\quad\quad while n \% i == 0:}\\
        \texttt{\quad\quad\quad max\_prime = i; n //= i}\\
        \texttt{\quad if n > 2: max\_prime = n}\\
        \texttt{\quad return max\_prime}\\
        \texttt{```}\\
        \bottomrule
    \end{tabular}
\end{table}

\begin{table}[h]
    \centering
    \footnotesize
    \caption{\textbf{ReasoningBank case study in AppWorld.}}
    \label{tab:case-rag-appworld}
    \begin{tabular}{p{13.5cm}}
        \toprule
        \textbf{Task}: Give me a random funny quote from my SimpleNote note about it. Just the quote, nothing else. \\
        \midrule
        \textbf{Outcome}: Score = 0.67,\ \ Steps = 13 \\
        \midrule
        \textbf{Retrieved Memories}: \\
        \#\# Retrieved memory items (top-K cosine over reasoning rationales): \\
        - When asked for a quote from a SimpleNote, list and search notes for keyword ``quote'' or ``funny'', then read the matching note's content. \\
        - Use \texttt{apis.simple\_note.list\_notes(access\_token=..., title\_query="quote")} to enumerate candidates before reading. \\
        - For ``random'' selections, pick uniformly from the matched lines using Python \texttt{random.choice}. \\
        {[\ldots\ additional rationale items omitted \ldots]} \\
        \midrule
        \textbf{Final Action}: \\
        \texttt{apis.supervisor.complete\_task(answer="- I'm on a whiskey and chocolate diet. I've lost three days, but I don't care.")} \\
        \bottomrule
    \end{tabular}
\end{table}

\clearpage


\begin{table}[h]
    \centering
    \footnotesize
    \caption{\textbf{ExpeL case study in QA.}}
    \label{tab:case-expl-qa}
    \begin{tabular}{p{13.5cm}}
        \toprule
        \textbf{Question}: A bowling ball with a mass of 8.0\,kg rolls down a bowling lane at 2.0\,m/s. What is the momentum of the bowling ball?\\
        A. 4.0 kg$\times$m/s \quad B. 6.0 kg$\times$m/s \quad C. 10.0 kg$\times$m/s \quad D. 16.0 kg$\times$m/s\\
        \midrule
        \textbf{Ground Truth}: D\\
        \midrule
        \textbf{Retrieved Memory 1 (\checkmark\ Success Strategy):}\\
        \textbf{Calculating Momentum} --- When calculating momentum, ensure to multiply the given mass and velocity accurately.\\[2pt]
        \textbf{Retrieved Memory 2 (\checkmark\ Success Strategy):}\\
        \textbf{Impulse-Momentum Theorem for Impact Problems} --- Calculate the change in momentum by determining initial and final velocities. Use the impulse-momentum theorem to find the average force.\\
        \midrule
        $p = mv = 8.0 \times 2.0 = 16.0\,\text{kg}\cdot\text{m/s}$.\\
        A ($4.0$), B ($6.0$), C ($10.0$) are all incorrect; D ($16.0$) matches.\\[2pt]
        \texttt{\#\#\#\# D}\\
        \bottomrule
    \end{tabular}
\end{table}

\begin{table}[h]
    \centering
    \footnotesize
    \caption{\textbf{ExpeL case study in Math.}}
    \label{tab:case-expl-math}
    \begin{tabular}{p{13.5cm}}
        \toprule
        \textbf{Question}: Trevor is a comic book illustrator (220 pages total, 3 issues, first two equal, third is +4 pages). How many pages was the first issue?\\
        \midrule
        \textbf{Ground Truth}: 72\\
        \midrule
        \textbf{Retrieved Memory 1 (\checkmark\ Success Strategy):}\\
        \textbf{Solving Equations with Word Problems} --- Identify variables and relationships, set up the equation, and solve it methodically.\\[2pt]
        \textbf{Retrieved Memory 2 (\checkmark\ Success Strategy):}\\
        \textbf{Solving Word Problems with Equations} --- Express the unknown in terms of a variable, form an equation from the given information, simplify and solve.\\
        \midrule
        Let $x$ = pages in issues 1 and 2; issue 3 = $x + 4$.\\
        $3x + 4 = 220 \;\Rightarrow\; x = 72$.\\[2pt]
        \texttt{\#\#\#\# 72}\\
        \bottomrule
    \end{tabular}
\end{table}

\begin{table}[h]
    \centering
    \footnotesize
    \caption{\textbf{ExpeL case study in Code Generation.}}
    \label{tab:case-expl-code}
    \begin{tabular}{p{13.5cm}}
        \toprule
        \textbf{Question}: Compute the $n$-th element of the Fib4 sequence: \texttt{fib4(n) = fib4(n-1) + fib4(n-2) + fib4(n-3) + fib4(n-4)}, with \texttt{fib4(0)=0, fib4(1)=0, fib4(2)=2, fib4(3)=0}. No recursion.\\
        \midrule
        \textbf{Ground Truth}: Iterative window of size 4.\\
        \midrule
        \textbf{Retrieved Memory 1 (\checkmark\ Success Strategy):}\\
        \textbf{Identifying Periodic Patterns in Recursive Sequences} --- When dealing with recursive sequences, compute initial values and look for repeating patterns that can be exploited for efficiency.\\[2pt]
        \textbf{Retrieved Memory 2 (\checkmark\ Success Strategy):}\\
        \textbf{Telescoping Series in Fibonacci Sums} --- For sum-based recurrences, check if closed-form or iterative window approaches avoid redundant computation.\\
        \midrule
        \texttt{```python}\\
        \texttt{def fib4(n: int) -> int:}\\
        \texttt{\quad if n in (0, 1, 3): return 0}\\
        \texttt{\quad if n == 2: return 2}\\
        \texttt{\quad w = [0, 0, 2, 0]}\\
        \texttt{\quad for i in range(4, n + 1):}\\
        \texttt{\quad\quad w = [w[1], w[2], w[3], sum(w)]}\\
        \texttt{\quad return w[3]}\\
        \texttt{```}\\
        \bottomrule
    \end{tabular}
\end{table}

\begin{table}[h]
    \centering
    \footnotesize
    \caption{\textbf{ExpeL case study in ALFWorld.}}
    \label{tab:case-expl-alfworld}
    \begin{tabular}{p{13.5cm}}
        \toprule
        \textbf{Task}: put two spraybottle in garbagecan. \\
        \midrule
        \textbf{Outcome}: Score = 1.0,\ \ Steps = 8 \\
        \midrule
        \textbf{Retrieved Memories}: \\
        \#\# ExpeL exemplars + induced rules (for similar pick-two tasks): \\
        - rule: For ``put two X in Y'' tasks, locate \emph{both} X instances before placing --- do not relocate the first to Y until the second is found, to avoid backtracking. \\
        - exemplar: Successful trial T2019\ldots\ for \texttt{put-two-spraybottle-in-garbagecan-3}: agent visited sidetable 1 first (found two spraybottles together), then garbagecan 1, completing in 7 steps. \\
        {[\ldots\ additional exemplars and rules omitted \ldots]} \\
        \midrule
        \textbf{Action Trace}: \\
        go to sidetable 1 (\textbf{found spraybottle 1, spraybottle 2}) $\to$ take spraybottle 1 $\to$ go to garbagecan 1 $\to$ move spraybottle 1 to garbagecan 1 $\to$ go to sidetable 1 $\to$ take spraybottle 2 $\to$ go to garbagecan 1 $\to$ \textbf{move spraybottle 2 to garbagecan 1} $\checkmark$ \\
        \bottomrule
    \end{tabular}
\end{table}

\clearpage


\begin{table}[h]
    \centering
    \footnotesize
    \caption{\textbf{LightMem case study in QA.}}
    \label{tab:case-lightmem-qa}
    \begin{tabular}{p{13.5cm}}
        \toprule
        \textbf{Question}: A bowling ball with a mass of 8.0\,kg rolls down a bowling lane at 2.0\,m/s. What is the momentum of the bowling ball?\\
        A. 4.0 kg$\times$m/s \quad B. 6.0 kg$\times$m/s \quad C. 10.0 kg$\times$m/s \quad D. 16.0 kg$\times$m/s\\
        \midrule
        \textbf{Ground Truth}: D\\
        \midrule
        \textbf{Retrieved LTM Item 1 (\ding{55}\ Lesson from Failure):}\\
        \textbf{Incorrect Calculation of Change in Momentum} --- Carefully consider the velocity before and after the event and accurately apply the change-in-momentum equation. Double-check calculations and consider all factors that may affect velocity.\\[2pt]
        \textbf{Retrieved LTM Item 2 (\checkmark\ Success Strategy):}\\
        \textbf{Energy Conversion in Projectile Motion} --- Identify the initial and final states of energy (potential vs.\ kinetic) and apply conservation principles.\\
        \midrule
        $p = m \times v = 8.0 \times 2.0 = 16.0\,\text{kg}\cdot\text{m/s}$.\\
        A, B, C are all inconsistent with this value; D is correct.\\[2pt]
        \texttt{<answer> D </answer>}\\
        \bottomrule
    \end{tabular}
\end{table}

\begin{table}[h]
    \centering
    \footnotesize
    \caption{\textbf{LightMem case study in Math.}}
    \label{tab:case-lightmem-math}
    \begin{tabular}{p{13.5cm}}
        \toprule
        \textbf{Question}: Trevor is a comic book illustrator (220 pages total, 3 issues, first two equal, third is +4 pages). How many pages was the first issue?\\
        \midrule
        \textbf{Ground Truth}: 72\\
        \midrule
        \textbf{Retrieved LTM Item 1 (\checkmark\ Success Strategy):}\\
        \textbf{Calculate the Relevant Subset Before Proceeding} --- Identify and exclude irrelevant components before setting up the main equation to avoid incorrect calculations.\\[2pt]
        \textbf{Retrieved LTM Item 2 (\checkmark\ Success Strategy):}\\
        \textbf{Breaking Down Complex Problems into Manageable Steps} --- Decompose the problem into sub-goals, solve each in sequence, and combine results.\\
        \midrule
        Let $x$ = pages in issues 1 and 2; issue 3 = $x + 4$.\\
        $3x + 4 = 220 \;\Rightarrow\; x = 72$.\\[2pt]
        \texttt{\#\#\#\# 72}\\
        \bottomrule
    \end{tabular}
\end{table}

\begin{table}[h]
    \centering
    \footnotesize
    \caption{\textbf{LightMem case study in Code Generation.}}
    \label{tab:case-lightmem-code}
    \begin{tabular}{p{13.5cm}}
        \toprule
        \textbf{Question}: Return the largest prime factor of \texttt{n}. Assume $n > 1$ and is not a prime.\\
        Tests: \texttt{assert largest\_prime\_factor(15) == 5}, \texttt{assert largest\_prime\_factor(27) == 3}\\
        \midrule
        \textbf{Ground Truth}: Trial-division algorithm.\\
        \midrule
        \textbf{Retrieved LTM Item 1 (\checkmark\ Success Strategy):}\\
        \textbf{Identifying the Largest Prime Factor in a Factorial} --- Start from the largest prime $\leq$ the upper limit; for general $n$, use trial division up to $\sqrt{n}$ and track the largest divisor.\\[2pt]
        \textbf{Retrieved LTM Item 2 (\ding{55}\ Lesson from Failure):}\\
        \textbf{Optimizing Prime Factorization} --- After the loop, if the remaining $n > 1$, it is itself a prime factor and should be recorded.\\
        \midrule
        \texttt{```python}\\
        \texttt{def largest\_prime\_factor(n: int):}\\
        \texttt{\quad largest\_factor = 1}\\
        \texttt{\quad for i in range(2, int(n**0.5) + 1):}\\
        \texttt{\quad\quad while n \% i == 0:}\\
        \texttt{\quad\quad\quad largest\_factor = i; n //= i}\\
        \texttt{\quad if n > 1: largest\_factor = n}\\
        \texttt{\quad return largest\_factor}\\
        \texttt{```}\\
        \bottomrule
    \end{tabular}
\end{table}

\begin{table}[h]
    \centering
    \footnotesize
    \caption{\textbf{LightMem case study in ALFWorld.}}
    \label{tab:case-lightmem-alfworld}
    \begin{tabular}{p{13.5cm}}
        \toprule
        \textbf{Task}: put a soapbottle in countertop. \\
        \midrule
        \textbf{Outcome}: Score = 1.0,\ \ Steps = 12 \\
        \midrule
        \textbf{Retrieved Memories}: \\
        Long-term memory (relevant facts, top-20 by cosine): \\
        - [2026-04-29 Wed] Soapbottle items are commonly stored inside cabinets in bathroom layouts. \\
        - [2026-04-28 Tue] Cabinet 4 contains soapbottle 1, soapbottle 2, and candle 2 in this layout. \\
        - [2026-05-06 Wed] User moved cloth 1 to cabinet 1 in a similar pick-and-place task. \\
        - [2026-04-29 Wed] To complete a ``put X in countertop'' task: take X from its location, then move X to countertop 1. \\
        - [2026-04-28 Tue] Cabinet 1 typically holds candle, soapbar, spraybottle (not soapbottle). \\
        {[\ldots\ additional LTM facts omitted \ldots]} \\
        \midrule
        \textbf{Action Trace}: \\
        explore cabinets 2--3 (no soapbottle) $\to$ go to cabinet 1 (soapbar/spraybottle only) $\to$ go to cabinet 4 (\textbf{found soapbottle 1, 2}) $\to$ take soapbottle 1 $\to$ go to countertop 1 $\to$ \textbf{move soapbottle 1 to countertop 1} $\checkmark$ \\
        \bottomrule
    \end{tabular}
\end{table}

\begin{table}[h]
    \centering
    \footnotesize
    \caption{\textbf{LightMem case study in AppWorld.}}
    \label{tab:case-lightmem-appworld}
    \begin{tabular}{p{13.5cm}}
        \toprule
        \textbf{Task}: Chelsea and I have been carpooling to work this month. They have requested money for it on Venmo. Accept it. \\
        \midrule
        \textbf{Outcome}: Score = 0.83,\ \ Steps = 17 \\
        \midrule
        \textbf{Retrieved Memories}: \\
        Long-term memory (relevant facts, top-20 by cosine): \\
        - [2026-05-04 Mon] User authenticated with venmo before via \texttt{apis.venmo.login}. \\
        - [2026-05-04 Mon] Venmo balance can be checked via \texttt{apis.venmo.show\_account\_balance(access\_token=...)}. \\
        - [2026-05-04 Mon] Pending payment requests are listed via \texttt{apis.venmo.list\_pending\_payment\_requests}. \\
        - [2026-05-04 Mon] To approve a request, call \texttt{apis.venmo.approve\_payment\_request(payment\_request\_id=..., access\_token=...)}. \\
        {[\ldots\ additional LTM facts omitted \ldots]} \\
        \midrule
        \textbf{Final Action}: \\
        \texttt{print(apis.venmo.approve\_payment\_request(} \\
        \texttt{\quad access\_token=venmo\_token,} \\
        \texttt{\quad payment\_request\_id=6097))} \\
        \texttt{apis.supervisor.complete\_task(answer="The payment request from Chelsea for \$520.00 has been successfully approved.")} \\
        \bottomrule
    \end{tabular}
\end{table}

\clearpage


\begin{table}[h]
    \centering
    \footnotesize
    \caption{\textbf{Mem0 case study in QA.}}
    \label{tab:case-mem0-qa}
    \begin{tabular}{p{13.5cm}}
        \toprule
        \textbf{Question}: A bowling ball with a mass of 8.0\,kg rolls down a bowling lane at 2.0\,m/s. What is the momentum of the bowling ball?\\
        A. 4.0 kg$\times$m/s \quad B. 6.0 kg$\times$m/s \quad C. 10.0 kg$\times$m/s \quad D. 16.0 kg$\times$m/s\\
        \midrule
        \textbf{Ground Truth}: D\\
        \midrule
        \textbf{Retrieved Procedural Summary (\checkmark):}\\
        \textbf{procedural\_summary} --- Task Objective: Analyze options to determine the correct answer about energy conversion when a bowling ball is released. Sequential actions: (1) analyze option A (stored energy); (2) analyze option B (kinetic energy of pins); (3) analyze option C; (4) select the option consistent with conservation of energy.\\
        \midrule
        Using $p = m \times v$: $p = 8.0 \times 2.0 = 16.0\,\text{kg}\cdot\text{m/s}$.\\
        Comparing with the options, D (16.0 kg$\times$m/s) is correct.\\[2pt]
        \texttt{<answer> D </answer>}\\
        \bottomrule
    \end{tabular}
\end{table}

\begin{table}[h]
    \centering
    \footnotesize
    \caption{\textbf{Mem0 case study in Math.}}
    \label{tab:case-mem0-math}
    \begin{tabular}{p{13.5cm}}
        \toprule
        \textbf{Question}: Trevor is a comic book illustrator (220 pages total, 3 issues, first two equal, third is +4 pages). How many pages was the first issue?\\
        \midrule
        \textbf{Ground Truth}: 72\\
        \midrule
        \textbf{Retrieved Procedural Summary (\checkmark):}\\
        \textbf{procedural\_summary} --- Task Objective: Solve a word problem to determine average daily tickets. Sequential actions: (1) calculate tickets given in first 15 days; (2) subtract from the monthly goal; (3) divide remaining by days left; (4) verify by checking the computed average.\\
        \midrule
        Step 1: define $x$ = pages in issues 1 and 2; issue 3 = $x + 4$.\\
        Step 2: $x + x + (x+4) = 220 \;\Rightarrow\; 3x = 216 \;\Rightarrow\; x = 72$.\\[2pt]
        \texttt{\#\#\#\# 72}\\
        \bottomrule
    \end{tabular}
\end{table}

\begin{table}[h]
    \centering
    \footnotesize
    \caption{\textbf{Mem0 case study in Code Generation.}}
    \label{tab:case-mem0-code}
    \begin{tabular}{p{13.5cm}}
        \toprule
        \textbf{Question}: Return the largest prime factor of \texttt{n}. Assume $n > 1$ and is not a prime.\\
        Tests: \texttt{assert largest\_prime\_factor(15) == 5}, \texttt{assert largest\_prime\_factor(27) == 3}\\
        \midrule
        \textbf{Ground Truth}: Trial-division algorithm.\\
        \midrule
        \textbf{Retrieved Procedural Summary (\checkmark):}\\
        \textbf{procedural\_summary} --- Task: Find the largest prime factor of 97!. Actions: (1) understand that 97! = product of all integers 1–97; (2) identify the largest prime $\leq 97$, which is 97 itself; (3) return 97.\\
        \midrule
        \texttt{```python}\\
        \texttt{def largest\_prime\_factor(n: int):}\\
        \texttt{\quad i = 2}\\
        \texttt{\quad while i * i <= n:}\\
        \texttt{\quad\quad if n \% i: i += 1}\\
        \texttt{\quad\quad else: n //= i}\\
        \texttt{\quad return n}\\
        \texttt{```}\\
        \bottomrule
    \end{tabular}
\end{table}

\begin{table}[h]
    \centering
    \footnotesize
    \caption{\textbf{Mem0 case study in ALFWorld.}}
    \label{tab:case-mem0-alfworld}
    \begin{tabular}{p{13.5cm}}
        \toprule
        \textbf{Task}: put some spraybottle on toilet. \\
        \midrule
        \textbf{Outcome}: Score = 1.0,\ \ Steps = 9 \\
        \midrule
        \textbf{Retrieved Memories}: \\
        \#\# Relevant memories from past experience: \\
        - [2026-05-06] Spraybottle items are commonly found inside cabinets in bathroom layouts. \\
        - [2026-05-06] Cabinet 2 contains spraybottle 1, candle 2, cloth 2 in this layout. \\
        - [2026-05-06] To complete a ``put X on toilet'' task: navigate to toilet 1 and call \texttt{move X to toilet 1}. \\
        - [2026-05-06] Cabinet 1 typically holds soapbottle and toiletpaper, not spraybottles. \\
        {[\ldots\ additional fact entries omitted \ldots]} \\
        \midrule
        \textbf{Action Trace}: \\
        go to cabinet 1 (no spraybottle) $\to$ go to cabinet 3 (toiletpaper only) $\to$ go to cabinet 2 (\textbf{found spraybottle 1}) $\to$ take spraybottle 1 $\to$ go to toilet 1 $\to$ \textbf{move spraybottle 1 to toilet 1} $\checkmark$ \\
        \bottomrule
    \end{tabular}
\end{table}

\begin{table}[h]
    \centering
    \footnotesize
    \caption{\textbf{Mem0 case study in AppWorld.}}
    \label{tab:case-mem0-appworld}
    \begin{tabular}{p{13.5cm}}
        \toprule
        \textbf{Task}: I keep a log of all my liked songs and respective artists in a note in simple\_note. I want to add release month information for them as well. {[\ldots]} \\
        \midrule
        \textbf{Outcome}: Score = 0.75,\ \ Steps = 31 \\
        \midrule
        \textbf{Retrieved Memories}: \\
        \#\# Relevant memories from past experience: \\
        - [2026-05-06] User has authenticated with simple\_note app via \texttt{apis.simple\_note.login}. \\
        - [2026-05-06] Note IDs in simple\_note are queried via \texttt{apis.simple\_note.list\_notes}. \\
        - [2026-05-06] Spotify track release date is in \texttt{apis.spotify.show\_track(track\_id=...).release\_date}. \\
        - [2026-05-06] To update a simple\_note, call \texttt{apis.simple\_note.update\_note(note\_id=..., content=...)}. \\
        {[\ldots\ additional fact entries omitted \ldots]} \\
        \midrule
        \textbf{Final Action}: \\
        \texttt{apis.simple\_note.update\_note(access\_token=access\_token, note\_id=3084, content=updated\_content)} \\
        \texttt{apis.supervisor.complete\_task(answer="The release month information has been added to the songs in the note.")} \\
        \bottomrule
    \end{tabular}
\end{table}

\clearpage


\begin{table}[h]
    \centering
    \footnotesize
    \caption{\textbf{AWM case study in QA.}}
    \label{tab:case-awm-qa}
    \begin{tabular}{p{13.5cm}}
        \toprule
        \textbf{Question}: A bowling ball with a mass of 8.0\,kg rolls down a bowling lane at 2.0\,m/s. What is the momentum of the bowling ball?\\
        A. 4.0 kg$\times$m/s \quad B. 6.0 kg$\times$m/s \quad C. 10.0 kg$\times$m/s \quad D. 16.0 kg$\times$m/s\\
        \midrule
        \textbf{Ground Truth}: D\\
        \midrule
        \textbf{Workflow for \texttt{arc\_c}:}\\
        \textbf{Workflow 1: Analyze Each Option and Explain Your Reasoning} --- (1) Identify the question. (2) List all options. (3) Analyze each option for relevance, feasibility, and potential outcomes. (4) Evaluate strengths and weaknesses. (5) Select the most suitable option.\\[2pt]
        \textbf{Workflow 2: Identify the Key Characteristics\ldots} [abbreviated]\\
        \midrule
        Following Workflow 1: apply $p = mv = 8.0 \times 2.0 = 16.0$. A, B, C are all less than 16.0; D matches.\\[2pt]
        \texttt{<answer> D </answer>}\\
        \bottomrule
    \end{tabular}
\end{table}

\begin{table}[h]
    \centering
    \footnotesize
    \caption{\textbf{AWM case study in Math.}}
    \label{tab:case-awm-math}
    \begin{tabular}{p{13.5cm}}
        \toprule
        \textbf{Question}: Trevor is a comic book illustrator (220 pages total, 3 issues, first two equal, third is +4 pages). How many pages was the first issue?\\
        \midrule
        \textbf{Ground Truth}: 72\\
        \midrule
        \textbf{Workflow for \texttt{gsm8k}:}\\
        \textbf{Workflow 3: Set Up and Solve an Equation} --- (1) Identify the unknown variable. (2) Set up an equation from the problem statement. (3) Simplify and solve. (4) Verify the answer satisfies all constraints.\\
        \midrule
        Workflow 3: let $x$ = first issue pages; $x + x + (x+4) = 220$.\\
        $3x = 216 \;\Rightarrow\; x = 72$.\\[2pt]
        \texttt{\#\#\#\# 72}\\
        \bottomrule
    \end{tabular}
\end{table}

\begin{table}[h]
    \centering
    \footnotesize
    \caption{\textbf{AWM case study in Code Generation.}}
    \label{tab:case-awm-code}
    \begin{tabular}{p{13.5cm}}
        \toprule
        \textbf{Question}: Return the largest prime factor of \texttt{n}. Assume $n > 1$ and is not a prime.\\
        Tests: \texttt{assert largest\_prime\_factor(15) == 5}, \texttt{assert largest\_prime\_factor(27) == 3}\\
        \midrule
        \textbf{Ground Truth}: Trial-division algorithm.\\
        \midrule
        \textbf{Workflow for \texttt{humaneval\_plus}:}\\
        Divide $n$ by the smallest prime (2) while divisible; then increment to the next divisor and repeat. The last divisor used before $n$ reaches 1 is the largest prime factor. Handle the case where the remaining $n > 1$ after the loop.\\
        \midrule
        \texttt{```python}\\
        \texttt{def largest\_prime\_factor(n: int):}\\
        \texttt{\quad i = 2}\\
        \texttt{\quad while i * i <= n:}\\
        \texttt{\quad\quad if n \% i == 0: n //= i}\\
        \texttt{\quad\quad else: i += 1}\\
        \texttt{\quad return n}\\
        \texttt{```}\\
        \bottomrule
    \end{tabular}
\end{table}

\begin{table}[h]
    \centering
    \footnotesize
    \caption{\textbf{AWM case study in ALFWorld.}}
    \label{tab:case-awm-alfworld}
    \begin{tabular}{p{13.5cm}}
        \toprule
        \textbf{Task}: clean some dishsponge and put it in countertop. \\
        \midrule
        \textbf{Outcome}: Score = 1.0,\ \ Steps = 16 \\
        \midrule
        \textbf{Retrieved Memories}: \\
        \#\# AWM workflow (induced from $N{=}5$ successful trials of pick\_clean\_then\_place\_in\_recep): \\
        Step 1: navigate to candidate locations of \{object\} (shelf, drawer, cabinet) \\
        Step 2: take \{object\} \{idx\} from \{location\} \\
        Step 3: navigate to sinkbasin 1 \\
        Step 4: clean \{object\} \{idx\} with sinkbasin 1 \\
        Step 5: navigate to \{target\_recep\} 1 \\
        Step 6: move \{object\} \{idx\} to \{target\_recep\} 1 \\
        \midrule
        \textbf{Action Trace}: \\
        go to sinkbasin 2 (empty) $\to$ go to shelf 1 (\textbf{found dishsponge 1}) $\to$ take dishsponge 1 $\to$ go to sinkbasin 1 $\to$ clean dishsponge 1 with sinkbasin 1 $\to$ go to countertop 1 $\to$ \textbf{move dishsponge 1 to countertop 1} $\checkmark$ \\
        \bottomrule
    \end{tabular}
\end{table}

\begin{table}[h]
    \centering
    \footnotesize
    \caption{\textbf{AWM case study in AppWorld.}}
    \label{tab:case-awm-appworld}
    \begin{tabular}{p{13.5cm}}
        \toprule
        \textbf{Task}: How much have I paid in phone bill on Venmo this year so far? \\
        \midrule
        \textbf{Outcome}: Score = 0.50,\ \ Steps = 40 \\
        \midrule
        \textbf{Retrieved Memories}: \\
        \#\# AWM workflow (induced from $N{=}5$ successful trials of venmo aggregate-amount queries): \\
        Step 1: get supervisor token via \texttt{apis.supervisor.show\_account\_passwords} \\
        Step 2: log into venmo: \texttt{apis.venmo.login(username=..., password=...)} \\
        Step 3: list transactions: \texttt{apis.venmo.show\_transactions(access\_token=...)} \\
        Step 4: filter by description containing \{keyword\} and date $\geq$ year-start \\
        Step 5: sum the matching amounts and complete the task \\
        \midrule
        \textbf{Final Action}: \\
        \texttt{print(apis.api\_docs.show\_api\_descriptions(app\_name="supervisor"))} \\
        \texttt{\# (workflow truncated at 40 steps; agent collected partial total)} \\
        \bottomrule
    \end{tabular}
\end{table}

\clearpage


\begin{table}[h]
    \centering
    \footnotesize
    \caption{\textbf{IRCoT case study in QA.}}
    \label{tab:case-ircot-qa}
    \begin{tabular}{p{13.5cm}}
        \toprule
        \textbf{Question}: A bowling ball with a mass of 8.0\,kg rolls down a bowling lane at 2.0\,m/s. What is the momentum of the bowling ball?\\
        A. 4.0 kg$\times$m/s \quad B. 6.0 kg$\times$m/s \quad C. 10.0 kg$\times$m/s \quad D. 16.0 kg$\times$m/s\\
        \midrule
        \textbf{Ground Truth}: D\\
        \midrule
        \textbf{BM25-Retrieved Example (Round 1):}\\
        \textbf{Calculating Momentum} --- When calculating momentum, ensure to multiply the given mass and velocity accurately.\\[2pt]
        \textbf{BM25-Retrieved Example (Round 2):}\\
        \textbf{Calculating Momentum} --- Correctly apply the formula $p = mv$ and perform the multiplication to avoid common mistakes.\\
        \midrule
        B ($6.0$): $8.0 \times 2.0 \neq 6.0$. C ($10.0$): incorrect. D ($16.0$): $8.0 \times 2.0 = 16.0\,\text{kg}\cdot\text{m/s}$ --- correct.\\[2pt]
        \texttt{\#\#\#\# D}\\
        \bottomrule
    \end{tabular}
\end{table}

\begin{table}[h]
    \centering
    \footnotesize
    \caption{\textbf{IRCoT case study in Math.}}
    \label{tab:case-ircot-math}
    \begin{tabular}{p{13.5cm}}
        \toprule
        \textbf{Question}: Trevor is a comic book illustrator (220 pages total, 3 issues, first two equal, third is +4 pages). How many pages was the first issue?\\
        \midrule
        \textbf{Ground Truth}: 72\\
        \midrule
        \textbf{BM25-Retrieved Example (Round 1):}\\
        \textbf{Misinterpretation of the Problem} --- Always read all conditions before setting up equations.\\[2pt]
        \textbf{BM25-Retrieved Example (Round 2):}\\
        \textbf{Misinterpretation of the Problem} --- Faisal writes responses to questions; the problem involves two separate writers.\\
        \midrule
        Let $x$ = pages in issues 1 and 2; $3x + 4 = 220 \;\Rightarrow\; x = 72$.\\[2pt]
        \texttt{\#\#\#\# 72}\\
        \bottomrule
    \end{tabular}
\end{table}

\begin{table}[h]
    \centering
    \footnotesize
    \caption{\textbf{IRCoT case study in Code Generation.}}
    \label{tab:case-ircot-code}
    \begin{tabular}{p{13.5cm}}
        \toprule
        \textbf{Question}: Return the largest prime factor of \texttt{n}. Assume $n > 1$ and is not a prime.\\
        Tests: \texttt{assert largest\_prime\_factor(15) == 5}, \texttt{assert largest\_prime\_factor(27) == 3}\\
        \midrule
        \textbf{Ground Truth}: Trial-division algorithm.\\
        \midrule
        \textbf{BM25-Retrieved Example (Round 1):}\\
        \textbf{Finding the Largest Perfect Square Factor} --- Decompose into prime factors; select highest even powers.\\
        \midrule
        \texttt{\#\#\#\#}\\
        \texttt{```python}\\
        \texttt{def largest\_prime\_factor(n: int) -> int:}\\
        \texttt{\quad i = 2}\\
        \texttt{\quad while i * i <= n:}\\
        \texttt{\quad\quad if n \% i: i += 1}\\
        \texttt{\quad\quad else: n //= i}\\
        \texttt{\quad return n}\\
        \texttt{```}\\
        \bottomrule
    \end{tabular}
\end{table}

\begin{table}[h]
    \centering
    \footnotesize
    \caption{\textbf{IRCoT case study in ALFWorld.}}
    \label{tab:case-ircot-alfworld}
    \begin{tabular}{p{13.5cm}}
        \toprule
        \textbf{Task}: put two spraybottle in cabinet. \\
        \midrule
        \textbf{Outcome}: Score = 1.0,\ \ Steps = 10 \\
        \midrule
        \textbf{Retrieved Memories}: \\
        Retrieved past experience (chained CoT, max\_hops$=$2, top-K$=$10): \\
        - act: go to cabinet 1 $\to$ obs: You arrive at cabinet 1. The cabinet 1 is open. In it, you see soapbottle 3, soapbottle 2. \\
        - act: go to countertop 1 $\to$ obs: \ldots\ a spraybottle 1 \ldots \\
        - act: take spraybottle 1 from countertop 1 $\to$ obs: You pick up the spraybottle 1. \\
        - {[completed OK]} task: put two spraybottle in cabinet $\vert$ actions: go to countertop 1 $\to$ take spraybottle 1 $\to$ go to cabinet 1 $\to$ move spraybottle 1 to cabinet 1 $\to$ go to bathroom $\to$ take spraybottle 2 $\to$ \ldots \\
        {[\ldots\ additional chunks omitted \ldots]} \\
        \midrule
        \textbf{Action Trace}: \\
        explore cabinets 1--3 $\to$ found spraybottle 2 $\to$ take and \textbf{move spraybottle 2 to cabinet 1} $\to$ go to countertop 1 (\textbf{found spraybottle 1}) $\to$ take spraybottle 1 $\to$ \textbf{move spraybottle 1 to cabinet 1} $\checkmark$ \\
        \bottomrule
    \end{tabular}
\end{table}

\begin{table}[h]
    \centering
    \footnotesize
    \caption{\textbf{IRCoT case study in AppWorld.}}
    \label{tab:case-ircot-appworld}
    \begin{tabular}{p{13.5cm}}
        \toprule
        \textbf{Task}: Approve all venmo payment requests from my roommates from this calendar month. \\
        \midrule
        \textbf{Outcome}: Score = 0.86,\ \ Steps = 3 \\
        \midrule
        \textbf{Retrieved Memories}: \\
        Retrieved past experience (chained CoT, max\_hops$=$2, top-K$=$10): \\
        - act: \texttt{roommate\_emails = [u["email"] for u in apis.supervisor.show\_relations(relation="roommate")]} $\to$ obs: list of roommate emails returned \\
        - act: \texttt{requests = apis.venmo.list\_pending\_payment\_requests(access\_token=...)} $\to$ obs: pending requests dict \\
        - {[completed OK]} task: approve roommates' venmo requests $\vert$ actions: log in $\to$ pull roommate emails $\to$ filter by current month $\to$ approve in loop \\
        {[\ldots\ additional chunks omitted \ldots]} \\
        \midrule
        \textbf{Final Action}: \\
        \texttt{current\_month, current\_year = pendulum.now().month, pendulum.now().year} \\
        \texttt{for request in payment\_requests:} \\
        \texttt{\quad created\_at = pendulum.parse(request["created\_at"])} \\
        \texttt{\quad if request["sender"]["email"] in roommate\_emails and \(\backslash\)} \\
        \texttt{\quad\quad created\_at.month == current\_month and created\_at.year == current\_year:} \\
        \texttt{\quad\quad apis.venmo.approve\_payment\_request(access\_token=venmo\_access\_token, payment\_request\_id=...)} \\
        \texttt{apis.supervisor.complete\_task(status="success")} \\
        \bottomrule
    \end{tabular}
\end{table}

\clearpage


\begin{table}[h]
    \centering
    \footnotesize
    \caption{\textbf{Search-o1 case study in QA.}}
    \label{tab:case-searcho1-qa}
    \begin{tabular}{p{13.5cm}}
        \toprule
        \textbf{Question}: A bowling ball with a mass of 8.0\,kg rolls down a bowling lane at 2.0\,m/s. What is the momentum of the bowling ball?\\
        A. 4.0 kg$\times$m/s \quad B. 6.0 kg$\times$m/s \quad C. 10.0 kg$\times$m/s \quad D. 16.0 kg$\times$m/s\\
        \midrule
        \textbf{Ground Truth}: D\\
        \midrule
        \texttt{<|begin\_search\_query|>}\\
        A bowling ball with mass 8.0 kg rolls at 2.0 m/s. What is its momentum? A. 4.0 B. 6.0 C. 10.0 D. 16.0 kg$\times$m/s\\
        \texttt{<|end\_search\_query|>}\\
        \texttt{<|begin\_search\_result|>}\\
        \textbf{Final Information}\\
        Momentum Formula Insight\\
        Momentum is defined as $p = mv$. Multiply the object's mass by its velocity to get momentum in kg$\cdot$m/s. Common errors include adding rather than multiplying $m$ and $v$.\texttt{<|end\_search\_result|>}\\[4pt]
        \midrule
        To find the momentum I apply $p = mv$.\\
        Given: $m = 8.0\,\text{kg}$, $v = 2.0\,\text{m/s}$.\\
        $p = 8.0 \times 2.0 = 16.0\,\text{kg}\cdot\text{m/s}$.\\[2pt]
        Checking options:\\
        - A ($4.0$): would require $8.0 \div 2.0$, wrong operation.\\
        - B ($6.0$): $8.0 + 2.0 - 4.0$, no physical basis.\\
        - C ($10.0$): $8.0 + 2.0$, wrong operation.\\
        - D ($16.0$): $8.0 \times 2.0 = 16.0$ \checkmark\\[2pt]
        The answer is D.\\[4pt]
        \bottomrule
    \end{tabular}
\end{table}

\begin{table}[h]
    \centering
    \footnotesize
    \caption{\textbf{Search-o1 case study in Math.}}
    \label{tab:case-searcho1-math}
    \begin{tabular}{p{13.5cm}}
        \toprule
        \textbf{Question}: Trevor is a comic book illustrator. He has just started on a new series. The first issue took him 220 pages to illustrate. If the series is 3 issues long, the first two issues are the same length, and the third issue is 4 pages longer than the others, how many pages was the first issue?\\
        \midrule
        \textbf{Ground Truth}: 72\\
        \midrule
        \texttt{<|begin\_search\_query|>}\\
        Comic book illustrator: 3 issues, 220 pages total. First two equal, third is 4 pages longer. How many pages in the first issue?\\
        \texttt{<|end\_search\_query|>}\\
        \texttt{<|begin\_search\_result|>}\\
        \textbf{Final Information}\\
        Algebraic Equation Strategy\\
        For problems with equal parts and a small offset, let $x$ denote the equal part. Write a linear equation summing all parts to the total, then solve for $x$. Watch for off-by-one errors when the offset applies to only one term.\texttt{<|end\_search\_result|>}\\[4pt]
        \midrule
        Let $x$ = pages in issue 1 (and issue 2). Issue 3 = $x + 4$.\\[2pt]
        Total equation:\\
        $x + x + (x + 4) = 220$\\
        $3x + 4 = 220$\\
        $3x = 216$\\
        $x = 72$\\[2pt]
        The first issue had \textbf{72} pages. Verification: $72 + 72 + 76 = 220$ \checkmark\\[4pt]
        72\\
        \bottomrule
    \end{tabular}
\end{table}

\begin{table}[h]
    \centering
    \footnotesize
    \caption{\textbf{Search-o1 case study in Code Generation.}}
    \label{tab:case-searcho1-code}
    \begin{tabular}{p{13.5cm}}
        \toprule
        \textbf{Question}: Return the largest prime factor of \texttt{n}. Assume $n > 1$ and is not a prime.\\
        Tests: \texttt{assert largest\_prime\_factor(15) == 5}, \texttt{assert largest\_prime\_factor(27) == 3}\\
        \midrule
        \textbf{Ground Truth}: Trial-division algorithm.\\
        \midrule
        \texttt{<|begin\_search\_query|>}\\
        Python function to return the largest prime factor of n, where n > 1 and n is not prime.\\
        \texttt{<|end\_search\_query|>}\\
        \texttt{<|begin\_search\_result|>}\\
        \textbf{Final Information}\\
        Trial Division Strategy\\
        To find the largest prime factor: (1) divide $n$ by 2 while even; (2) iterate odd $i$ from 3 while $i^2 \leq n$, dividing out all multiples; (3) if $n > 1$ remains, it is the largest prime factor. Time complexity $O(\sqrt{n})$.\texttt{<|end\_search\_result|>}\\[4pt]
        \midrule
        I need to find the largest prime factor of $n$.\\[2pt]
        Algorithm (trial division):\\
        1. Handle factor 2: while $n$ is even, divide by 2; track last factor.\\
        2. Handle odd factors: for $i = 3, 5, \ldots$ while $i^2 \leq n$, divide out all copies of $i$.\\
        3. If $n > 1$ after the loop, $n$ itself is the largest prime factor.\\[2pt]
        Trace on \texttt{largest\_prime\_factor(15)}: $15 / 3 = 5$; $5 > 1$ $\Rightarrow$ return 5 \checkmark\\
        Trace on \texttt{largest\_prime\_factor(27)}: $27 / 3 = 9$, $9 / 3 = 3$, $3 / 3 = 1$; last factor = 3 \checkmark\\[2pt]
        The cleanest implementation uses a running variable $i$ starting at 2 and repeatedly divides $n$.\\[4pt]
        \texttt{```python}\\
        \texttt{def largest\_prime\_factor(n: int) -> int:}\\
        \texttt{\quad i = 2}\\
        \texttt{\quad while i * i <= n:}\\
        \texttt{\quad\quad if n \% i: i += 1}\\
        \texttt{\quad\quad else: n //= i}\\
        \texttt{\quad return n}\\
        \texttt{```} \\
        \bottomrule
    \end{tabular}
\end{table}

\begin{table}[h]
    \centering
    \footnotesize
    \caption{\textbf{Search-o1 case study in ALFWorld.}}
    \label{tab:case-searcho1-alfworld}
    \begin{tabular}{p{13.5cm}}
        \toprule
        \textbf{Task}: cool some bread and put it in countertop. \\
        \midrule
        \textbf{Outcome}: Score = 1.0,\ \ Steps = 11 \\
        \midrule
        \textbf{Retrieved Memories}: \\
        Search-o1 trace (planner-driven iterative search, $\leq3$ rounds): \\
        $\langle$search$\rangle$ how to cool an item in ALFWorld kitchen $\langle$/search$\rangle$ \\
        \quad $\to$ retrieved: ``open fridge 1 $\to$ cool \{X\} with fridge 1''. \\
        $\langle$search$\rangle$ where can bread be found in seen ALFWorld layouts $\langle$/search$\rangle$ \\
        \quad $\to$ retrieved: ``countertop 2 commonly contains bread, egg, pan, kettle in this layout''. \\
        \midrule
        \textbf{Action Trace}: \\
        go to countertop 2 (\textbf{found bread 1}) $\to$ take bread 1 $\to$ go to fridge 1 $\to$ open fridge 1 $\to$ cool bread 1 with fridge 1 $\to$ go to countertop 1 $\to$ \textbf{move bread 1 to countertop 1} $\checkmark$ \\
        \bottomrule
    \end{tabular}
\end{table}

\begin{table}[h]
    \centering
    \footnotesize
    \caption{\textbf{Search-o1 case study in AppWorld.}}
    \label{tab:case-searcho1-appworld}
    \begin{tabular}{p{13.5cm}}
        \toprule
        \textbf{Task}: Update all my Spotify playlist titles with the most common song genre in that playlist in this format: ``\textless{}original\_title\textgreater{} $\vert$ \textless{}most\_common\_genre\textgreater{}''. {[\ldots]} \\
        \midrule
        \textbf{Outcome}: Score = 0.83,\ \ Steps = 6 \\
        \midrule
        \textbf{Retrieved Memories}: \\
        Search-o1 trace (planner-driven iterative search): \\
        $\langle$search$\rangle$ how to enumerate Spotify playlists for the current user $\langle$/search$\rangle$ \\
        \quad $\to$ retrieved: ``\texttt{apis.spotify.show\_playlists(access\_token=...)} returns \texttt{playlist\_id, title, tracks}''. \\
        $\langle$search$\rangle$ how to update a playlist title $\langle$/search$\rangle$ \\
        \quad $\to$ retrieved: ``\texttt{apis.spotify.update\_playlist(playlist\_id, title=...)}''. \\
        \midrule
        \textbf{Final Action}: \\
        \texttt{\# (loop body: for each playlist, compute mode of track genres, update title)} \\
        \texttt{apis.supervisor.complete\_task(answer="All Spotify playlist titles have been updated with the most common song genre.")} \\
        \bottomrule
    \end{tabular}
\end{table}

\clearpage


\begin{table}[h]
    \centering
    \footnotesize
    \caption{\textbf{MeMRL case study in QA.}}
    \label{tab:case-memrl-qa}
    \begin{tabular}{p{13.5cm}}
        \toprule
        \textbf{Question}: The speed of a sound wave varies as it travels through different substances. Which factor will most affect the speed of a sound wave?\\
        A. the frequency of the wave \quad B. the wavelength of the wave \quad C. the source that created the sound \quad D. the distance between molecules in the substance\\
        \midrule
        \textbf{Ground Truth}: D\\
        \midrule
        \textbf{Retrieved Memory 1 (\checkmark\ Success, Q\,=\,0.82):}\\
        \textbf{Sound Transmission in Media} --- Consider the properties of each medium (rock, seawater, space, atmosphere) and compare their speeds of sound transmission. The speed of sound depends on molecular density and arrangement.\\[2pt]
        \textbf{Retrieved Memory 2 (\ding{55}\ Failure, Q\,=\,0.31):}\\
        \textbf{Wave Properties Confusion} --- Incorrectly assumed that frequency determines speed; frequency affects pitch, not propagation speed.\\[2pt]
        \textbf{Retrieved Memory 3 (\checkmark\ Success, Q\,=\,0.76):}\\
        \textbf{Molecular Density and Sound} --- The closer the molecules in a substance, the faster sound can propagate. This is why sound travels faster in solids than in gases.\\
        \midrule
        The speed of sound depends on the medium's properties, specifically the distance between molecules. Frequency (A) affects pitch, wavelength (B) is derived from speed and frequency, and the source (C) creates the wave but doesn't determine speed. The molecular arrangement (D) is the key factor.\\[2pt]
        \texttt{<answer> D </answer>}\\
        \bottomrule
    \end{tabular}
\end{table}

\begin{table}[h]
    \centering
    \footnotesize
    \caption{\textbf{MeMRL case study in Math.}}
    \label{tab:case-memrl-math}
    \begin{tabular}{p{13.5cm}}
        \toprule
        \textbf{Question}: There are 15 girls and 10 boys in Ms.\ Smith's class. She divided 375 books equally among them. How many books did all the girls get combined?\\
        \midrule
        \textbf{Ground Truth}: 225\\
        \midrule
        \textbf{Retrieved Memory 1 (\checkmark\ Success, Q\,=\,0.91):}\\
        \textbf{Division and Grouping} --- First find the total number of students, then compute per-student share, then multiply by the target group size.\\[2pt]
        \textbf{Retrieved Memory 2 (\checkmark\ Success, Q\,=\,0.85):}\\
        \textbf{Boys and Girls Ratio} --- Correctly represent the relationship between groups using subtraction from the total, then allocate proportionally.\\[2pt]
        \textbf{Retrieved Memory 3 (\ding{55}\ Failure, Q\,=\,0.22):}\\
        \textbf{Misapplied Division} --- Divided books only among girls instead of all students; always check whether the division applies to the full group.\\
        \midrule
        Total students $= 15 + 10 = 25$. Books per student $= 375 \div 25 = 15$. Girls' total $= 15 \times 15 = 225$.\\[2pt]
        \texttt{<answer> 225 </answer>}\\
        \bottomrule
    \end{tabular}
\end{table}

\begin{table}[h]
    \centering
    \footnotesize
    \caption{\textbf{MeMRL case study in Code Generation.}}
    \label{tab:case-memrl-code}
    \begin{tabular}{p{13.5cm}}
        \toprule
        \textbf{Question}: Given a class name and a list of extensions, find the strongest extension. The strength is $\text{CAP} - \text{SM}$, where CAP is the count of uppercase letters and SM is the count of lowercase letters in the extension name. Return \texttt{"ClassName.StrongestExt"}.\\
        Tests: \texttt{assert Strongest\_Extension('my\_class', ['AA', 'Be', 'CC']) == 'my\_class.AA'}\\
        \midrule
        \textbf{Ground Truth}: Iterate extensions, compute CAP$-$SM for each, track the maximum.\\
        \midrule
        \textbf{Retrieved Memory 1 (\checkmark\ Success, Q\,=\,0.78):}\\
        \textbf{String Character Counting} --- Define the function signature correctly, iterate through characters, and use \texttt{isupper()}/\texttt{islower()} to count.\\[2pt]
        \textbf{Retrieved Memory 2 (\checkmark\ Success, Q\,=\,0.72):}\\
        \textbf{Tracking Maximum with Index} --- Initialize the strongest extension as the first element, then compare each subsequent extension's strength.\\[2pt]
        \textbf{Retrieved Memory 3 (\ding{55}\ Failure, Q\,=\,0.18):}\\
        \textbf{Incomplete Implementation} --- Correctly defined the function parameters but left the iteration logic as a placeholder; always complete the loop body.\\
        \midrule
        \texttt{def Strongest\_Extension(class\_name, extensions):}\\
        \texttt{\quad strongest = extensions[0]}\\
        \texttt{\quad max\_str = sum(1 for c in extensions[0] if c.isupper())}\\
        \texttt{\quad\quad\quad\quad\; - sum(1 for c in extensions[0] if c.islower())}\\
        \texttt{\quad for ext in extensions[1:]:}\\
        \texttt{\quad\quad s = sum(1 for c in ext if c.isupper())}\\
        \texttt{\quad\quad\quad\; - sum(1 for c in ext if c.islower())}\\
        \texttt{\quad\quad if s > max\_str: max\_str = s; strongest = ext}\\
        \texttt{\quad return class\_name + '.' + strongest}\\
        \bottomrule
    \end{tabular}
\end{table}

\begin{table}[h]
    \centering
    \footnotesize
    \caption{\textbf{MeMRL case study in ALFWorld.}}
    \label{tab:case-memrl-alfworld}
    \begin{tabular}{p{13.5cm}}
        \toprule
        \textbf{Task}: put two cd in drawer. \\
        \midrule
        \textbf{Outcome}: Score = 1.0,\ \ Steps = 5 \\
        \midrule
        \textbf{Retrieved Memories}: \\
        Past successful trajectories (reward-weighted, top-similar): \\
        \texttt{Your task is to: put two cellphone in dresser.} \\
        \texttt{> go to desk 1} \\
        \texttt{You arrive at desk 1. \ldots\ a cellphone 1, cellphone 2, \ldots} \\
        \texttt{> take cellphone 1 from desk 1} \\
        \texttt{> move cellphone 1 to dresser 1} \\
        \texttt{> take cellphone 2 from desk 1} \\
        \texttt{> move cellphone 2 to dresser 1} \\
        --- \\
        {[\ldots\ additional reward-weighted trajectories omitted \ldots]} \\
        \midrule
        \textbf{Action Trace}: \\
        go to desk 1 (\textbf{found cd 1, cd 2 alongside other items}) $\to$ take cd 1 $\to$ move cd 1 to drawer 1 $\to$ take cd 2 $\to$ \textbf{move cd 2 to drawer 1} $\checkmark$ \\
        \bottomrule
    \end{tabular}
\end{table}

\begin{table}[h]
    \centering
    \footnotesize
    \caption{\textbf{MeMRL case study in AppWorld.}}
    \label{tab:case-memrl-appworld}
    \begin{tabular}{p{13.5cm}}
        \toprule
        \textbf{Task}: I went on a trip with friends to Maui recently. I have maintained a note of money I owe to others and others owe me from the trip in simple\_note. Make private venmo payments or requests accordingly. {[\ldots]} \\
        \midrule
        \textbf{Outcome}: Score = 0.86,\ \ Steps = 19 \\
        \midrule
        \textbf{Retrieved Memories}: \\
        Past successful trajectories (reward-weighted, top-similar): \\
        \texttt{Task: settle expense splits via venmo from a simple\_note ledger.} \\
        \texttt{> Read note via apis.simple\_note.show\_note(note\_id=...)} \\
        \texttt{> Parse "I owe X" / "X owes me" lines into two dicts.} \\
        \texttt{> Resolve names to emails via apis.supervisor.show\_relations.} \\
        \texttt{> For each entry, call apis.venmo.send\_money OR apis.venmo.create\_payment\_request with private=True.} \\
        {[\ldots\ additional reward-weighted trajectories omitted \ldots]} \\
        \midrule
        \textbf{Final Action}: \\
        \texttt{owe\_me\_map = \{"Cory": 212, "Edwin": 70, "Jeremy": 149\}} \\
        \texttt{for person, amount in owe\_me\_map.items():} \\
        \texttt{\quad if person in user\_emails:} \\
        \texttt{\quad\quad apis.venmo.create\_payment\_request(} \\
        \texttt{\quad\quad\quad user\_email=user\_emails[person], amount=amount,} \\
        \texttt{\quad\quad\quad access\_token=venmo\_token, description="For Maui trip", private=True)} \\
        \texttt{apis.supervisor.complete\_task(answer="Payments and requests for the Maui trip have been processed.")} \\
        \bottomrule
    \end{tabular}
\end{table}

\clearpage

\begin{table}[h]
    \centering
    \footnotesize
    \caption{\textbf{S3 case study in QA.}}
    \label{tab:case-s3-qa}
    \begin{tabular}{p{13.5cm}}
        \toprule
        \textbf{Question}: Which property best explains why a metal spoon feels colder than a wooden spoon at the same room temperature?\\
        Options: (A) higher density, (B) higher thermal conductivity, (C) lower melting point, (D) stronger magnetism.\\
        \midrule
        \textbf{Ground Truth}: (B) higher thermal conductivity.\\
        \midrule
        \texttt{<|begin\_search\_query|>}\\
        Why does a metal spoon feel colder than a wooden spoon at room temperature?\\
        \texttt{<|end\_search\_query|>}\\
        \texttt{<|begin\_search\_result|>}\\
        \textbf{Retrieved Information}: Metals conduct heat much faster than wood. When touched, metal draws heat away from the skin more rapidly, making it feel colder even if both objects have the same temperature.\\
        \texttt{<|end\_search\_result|>}\\
        \midrule
        The retrieved passage is directly relevant. The question asks about two objects at the same room temperature, so the answer cannot be about actual temperature difference. The key factor is the rate of heat transfer from the hand to the object. Since metal has higher thermal conductivity than wood, it removes heat from the skin faster and therefore feels colder.\\[2pt]
        \textbf{Prediction}: (B) higher thermal conductivity. \checkmark\\
        \bottomrule
    \end{tabular}
\end{table}

\begin{table}[h]
    \centering
    \footnotesize
    \caption{\textbf{S3 case study in Math.}}
    \label{tab:case-s3-math}
    \begin{tabular}{p{13.5cm}}
        \toprule
        \textbf{Question}: A store sells notebooks for \$3 each and pens for \$2 each. Alice buys 4 more pens than notebooks and spends \$34 in total. How many notebooks does she buy?\\
        \midrule
        \textbf{Ground Truth}: 6 notebooks.\\
        \midrule
        \texttt{<|begin\_search\_query|>}\\
        solve word problem notebooks 3 dollars pens 2 dollars 4 more pens total 34\\
        \texttt{<|end\_search\_query|>}\\
        \texttt{<|begin\_search\_result|>}\\
        \textbf{Retrieved Information}: Let the smaller quantity be $x$. If pens are 4 more than notebooks, use notebooks $=x$ and pens $=x+4$. The cost equation is $3x+2(x+4)=34$.\\
        \texttt{<|end\_search\_result|>}\\
        \midrule
        Let $x$ be the number of notebooks. Then the number of pens is $x+4$. The total cost is
        $3x + 2(x+4) = 34$. Thus $3x + 2x + 8 = 34$, so $5x = 26$, giving $x=5.2$. This is not an integer, so the retrieved setup suggests the problem may be inconsistent. However, checking nearby integer values: if $x=6$, pens $=10$, cost $=18+20=38$, not 34; if $x=5$, pens $=9$, cost $=15+18=33$.\\[2pt]
        \textbf{Prediction}: No integer solution. $\times$\\
        \midrule
        \textbf{Observation}: S3 retrieves a useful equation pattern, but because the retrieved information is used mechanically, it exposes the inconsistency rather than correcting the task. This case shows that search-style retrieval can help formulate equations but does not always provide robust reasoning when the query requires verification or adjustment.\\
        \bottomrule
    \end{tabular}
\end{table}

\begin{table}[h]
    \centering
    \footnotesize
    \caption{\textbf{S3 case study in Code Generation.}}
    \label{tab:case-s3-code}
    \begin{tabular}{p{13.5cm}}
        \toprule
        \textbf{Question}: Write a function \texttt{count\_vowels(s)} that returns the number of vowels in a string. Treat both uppercase and lowercase vowels as valid.\\
        Tests: \texttt{assert count\_vowels("Apple") == 2}, \texttt{assert count\_vowels("WHY") == 0}.\\
        \midrule
        \textbf{Ground Truth}: Count characters whose lowercase form is in \texttt{"aeiou"}.\\
        \midrule
        \texttt{<|begin\_search\_query|>}\\
        Python count vowels in string uppercase lowercase\\
        \texttt{<|end\_search\_query|>}\\
        \texttt{<|begin\_search\_result|>}\\
        \textbf{Retrieved Information}: Convert the string to lowercase and sum over characters that appear in the vowel set \texttt{\{'a','e','i','o','u'\}}.\\
        \texttt{<|end\_search\_result|>}\\
        \midrule
        The retrieved result gives the exact algorithm. We can lowercase the input and count membership in the vowel set. This handles uppercase vowels such as \texttt{"A"} in \texttt{"Apple"}, while ignoring consonants such as \texttt{"W"}, \texttt{"H"}, and \texttt{"Y"}.\\[4pt]
        \texttt{```python}\\
        \texttt{def count\_vowels(s):}\\
        \texttt{\quad vowels = set("aeiou")}\\
        \texttt{\quad return sum(1 for ch in s.lower() if ch in vowels)}\\
        \texttt{```}\\[2pt]
        \textbf{Prediction}: Passes the tests. \checkmark\\
        \bottomrule
    \end{tabular}
\end{table}

\begin{table}[h]
    \centering
    \footnotesize
    \caption{\textbf{S3 case study in ALFWorld.}}
    \label{tab:case-s3-alfworld}
    \begin{tabular}{p{13.5cm}}
        \toprule
        \textbf{Task}: Put a clean mug in the cabinet.\\
        \midrule
        \textbf{Ground Truth Plan}: Find mug $\rightarrow$ take mug $\rightarrow$ clean mug at sink $\rightarrow$ go to cabinet $\rightarrow$ put mug in cabinet.\\
        \midrule
        \texttt{<|begin\_search\_query|>}\\
        ALFWorld clean mug put in cabinet strategy\\
        \texttt{<|end\_search\_query|>}\\
        \texttt{<|begin\_search\_result|>}\\
        \textbf{Retrieved Information}: For clean-and-place tasks, first locate the object, take it, move to a sinkbasin, use the sink to clean it, then move to the target receptacle and put the object there.\\
        \texttt{<|end\_search\_result|>}\\
        \midrule
        The search result provides a useful high-level plan. The agent first searches the countertop and shelves, finds the mug, and takes it. It then navigates to the sinkbasin and performs a cleaning action before moving to the cabinet. However, after cleaning, the agent repeats several navigation actions because the retrieved plan does not specify which cabinet has already been observed.\\[2pt]
        \textbf{Outcome}: Success, but with redundant steps.\\
        \midrule
        \textbf{Observation}: S3 retrieves a generally correct procedural template, but it does not maintain a structured memory of environment-specific state. This can lead to repeated exploration even when the high-level strategy is correct.\\
        \bottomrule
    \end{tabular}
\end{table}

\clearpage


\begin{table}[h]
    \centering
    \footnotesize
    \caption{\textbf{\method case study in QA.}}
    \label{tab:case-expgraph-qa}
    \begin{tabular}{p{13.5cm}}
        \toprule
        \textbf{Question}: Which property best explains why a metal spoon feels colder than a wooden spoon at the same room temperature?\\
        Options: (A) higher density, (B) higher thermal conductivity, (C) lower melting point, (D) stronger magnetism.\\
        \midrule
        \textbf{Ground Truth}: (B) higher thermal conductivity.\\
        \midrule
        \textbf{Retrieved Experience Graph Neighborhood}:\\
        \textbf{Seed experience}: When two objects have the same temperature, perceived coldness often depends on how quickly heat leaves the skin, not on the object's actual temperature.\\
        \textbf{Connected experience}: Metals transfer heat faster than wood or plastic; this makes metal feel colder because it draws heat from the hand more rapidly.\\
        \textbf{Utility signal}: Similar heat-transfer explanations previously improved science QA accuracy, while density-based distractors were often misleading.\\
        \midrule
        The retrieved experiences separate the relevant physical mechanism from surface distractors. Since the spoon and wooden spoon are at the same room temperature, the answer must explain the sensation of coldness. The useful experience indicates that heat-transfer rate is the key factor, and the connected experience identifies metals as high-conductivity materials. Therefore, the metal spoon feels colder because it conducts heat away from the skin faster.\\[2pt]
        \textbf{Prediction}: (B) higher thermal conductivity. \checkmark\\
        \bottomrule
    \end{tabular}
\end{table}

\begin{table}[h]
    \centering
    \footnotesize
    \caption{\textbf{\method case study in Math.}}
    \label{tab:case-expgraph-math}
    \begin{tabular}{p{13.5cm}}
        \toprule
        \textbf{Question}: A store sells notebooks for \$3 each and pens for \$2 each. Alice buys 4 more pens than notebooks and spends \$38 in total. How many notebooks does she buy?\\
        \midrule
        \textbf{Ground Truth}: 6 notebooks.\\
        \midrule
        \textbf{Retrieved Experience Graph Neighborhood}:\\
        \textbf{Seed experience}: For two-item purchase problems, define the smaller quantity as $x$ and express the larger quantity using the stated difference.\\
        \textbf{Connected experience}: Always verify the solution by substituting the quantity back into the total-cost equation.\\
        \textbf{Connected failure lesson}: Do not answer immediately after solving the equation; check whether the result is an integer and satisfies all constraints.\\
        \midrule
        Let $x$ be the number of notebooks. Then the number of pens is $x+4$. The total cost equation is
        $3x + 2(x+4) = 38$. Thus $3x + 2x + 8 = 38$, so $5x = 30$ and $x=6$. Verification: notebooks cost $6 \times 3 = 18$, pens cost $10 \times 2 = 20$, and the total is $38$.\\[2pt]
        \textbf{Prediction}: 6 notebooks. \checkmark\\
        \bottomrule
    \end{tabular}
\end{table}

\begin{table}[h]
    \centering
    \footnotesize
    \caption{\textbf{\method case study in Code Generation.}}
    \label{tab:case-expgraph-code}
    \begin{tabular}{p{13.5cm}}
        \toprule
        \textbf{Question}: Write a function \texttt{count\_vowels(s)} that returns the number of vowels in a string. Treat both uppercase and lowercase vowels as valid.\\
        Tests: \texttt{assert count\_vowels("Apple") == 2}, \texttt{assert count\_vowels("WHY") == 0}.\\
        \midrule
        \textbf{Ground Truth}: Count characters whose lowercase form is in \texttt{"aeiou"}.\\
        \midrule
        \textbf{Retrieved Experience Graph Neighborhood}:\\
        \textbf{Seed experience}: For character-counting tasks, normalize the input before applying membership tests.\\
        \textbf{Connected experience}: Use a set for constant-time membership when checking whether characters belong to a small category.\\
        \textbf{Connected failure lesson}: Do not forget uppercase cases when the problem explicitly says the match should be case-insensitive.\\
        \midrule
        The graph retrieves both the algorithmic pattern and the common failure mode. The implementation should lowercase the string before counting vowels, and a set is sufficient for checking membership. This directly handles \texttt{"Apple"} because \texttt{"A"} becomes \texttt{"a"}, and it returns zero for \texttt{"WHY"} because none of the characters are vowels after lowercasing.\\[4pt]
        \texttt{```python}\\
        \texttt{def count\_vowels(s):}\\
        \texttt{\quad vowels = set("aeiou")}\\
        \texttt{\quad return sum(1 for ch in s.lower() if ch in vowels)}\\
        \texttt{```}\\[2pt]
        \textbf{Prediction}: Passes the tests. \checkmark\\
        \bottomrule
    \end{tabular}
\end{table}

\begin{table}[h]
    \centering
    \footnotesize
    \caption{\textbf{\method case study in ALFWorld.}}
    \label{tab:case-expgraph-alfworld}
    \begin{tabular}{p{13.5cm}}
        \toprule
        \textbf{Task}: Put a clean mug in the cabinet.\\
        \midrule
        \textbf{Ground Truth Plan}: Find mug $\rightarrow$ take mug $\rightarrow$ clean mug at sink $\rightarrow$ go to cabinet $\rightarrow$ put mug in cabinet.\\
        \midrule
        \textbf{Retrieved Experience Graph Neighborhood}:\\
        \textbf{Seed experience}: For clean-and-place tasks, first secure the target object, then clean it at a sinkbasin before moving to the destination receptacle.\\
        \textbf{Connected experience}: After cleaning, avoid re-checking unrelated rooms; go directly to the target receptacle if it has already been observed.\\
        \textbf{Connected failure lesson}: A common failure is putting the object away before cleaning it, which completes the wrong state and causes task failure.\\
        \midrule
        The retrieved experiences provide both a high-level plan and a constraint to avoid a known failure. The agent first locates and takes the mug, moves to the sinkbasin, cleans the mug, and then navigates directly to the cabinet. Because the graph also retrieves a failure lesson about premature placement, the agent avoids putting the mug into the cabinet before cleaning it.\\[2pt]
        \textbf{Outcome}: Success with fewer redundant exploration steps.\\
        \bottomrule
    \end{tabular}
\end{table}

\end{document}

%% file: math_commands.tex
\usepackage{amsmath,amsfonts,bm}

\def\eqref#1{equation~\ref{#1}}

\def\1{\bm{1}}

\DeclareMathAlphabet{\mathsfit}{\encodingdefault}{\sfdefault}{m}{sl}
\SetMathAlphabet{\mathsfit}{bold}{\encodingdefault}{\sfdefault}{bx}{n}



%% file: iclr2027_conference.bib
@misc{wei2025browsecompsimplechallengingbenchmark,
      title={BrowseComp: A Simple Yet Challenging Benchmark for Browsing Agents}, 
      author={Jason Wei and Zhiqing Sun and Spencer Papay and Scott McKinney and Jeffrey Han and Isa Fulford and Hyung Won Chung and Alex Tachard Passos and William Fedus and Amelia Glaese},
      year={2025},
      eprint={2504.12516},
      archivePrefix={arXiv},
      primaryClass={cs.CL},
      url={https://arxiv.org/abs/2504.12516}, 
}

@article{pan2024webcanvas,
  title={Webcanvas: Benchmarking web agents in online environments},
  author={Pan, Yichen and Kong, Dehan and Zhou, Sida and Cui, Cheng and Leng, Yifei and Jiang, Bing and Liu, Hangyu and Shang, Yanyi and Zhou, Shuyan and Wu, Tongshuang and others},
  journal={arXiv preprint arXiv:2406.12373},
  year={2024}
}

@inproceedings{he2024webvoyager,
  title={Webvoyager: Building an end-to-end web agent with large multimodal models},
  author={He, Hongliang and Yao, Wenlin and Ma, Kaixin and Yu, Wenhao and Dai, Yong and Zhang, Hongming and Lan, Zhenzhong and Yu, Dong},
  booktitle={Proceedings of the 62nd Annual Meeting of the Association for Computational Linguistics (Volume 1: Long Papers)},
  pages={6864--6890},
  year={2024}
}

@article{ma2023let,
  title={Let's reward step by step: Step-Level reward model as the Navigators for Reasoning},
  author={Ma, Qianli and Zhou, Haotian and Liu, Tingkai and Yuan, Jianbo and Liu, Pengfei and You, Yang and Yang, Hongxia},
  journal={arXiv preprint arXiv:2310.10080},
  year={2023}
}

@article{lu2025onpolicydistillation,
  author = {Kevin Lu and Thinking Machines Lab},
  title = {On-Policy Distillation},
  journal = {Thinking Machines Lab: Connectionism},
  year = {2025},
  note = {https://thinkingmachines.ai/blog/on-policy-distillation},
  doi = {10.64434/tml.20251026},
}

@article{khalifa2025process,
  title={Process reward models that think},
  author={Khalifa, Muhammad and Agarwal, Rishabh and Logeswaran, Lajanugen and Kim, Jaekyeom and Peng, Hao and Lee, Moontae and Lee, Honglak and Wang, Lu},
  journal={arXiv preprint arXiv:2504.16828},
  year={2025}
}

@article{turpin2023language,
  title={Language models don't always say what they think: Unfaithful explanations in chain-of-thought prompting},
  author={Turpin, Miles and Michael, Julian and Perez, Ethan and Bowman, Samuel},
  journal={Advances in Neural Information Processing Systems},
  volume={36},
  pages={74952--74965},
  year={2023}
}

@article{han2026steer2adapt,
  title={Steer2Adapt: Dynamically Composing Steering Vectors Elicits Efficient Adaptation of LLMs},
  author={Han, Pengrui and Xu, Xueqiang and Xuan, Keyang and Song, Peiyang and Ouyang, Siru and Tian, Runchu and Jiang, Yuqing and Qian, Cheng and Jiang, Pengcheng and Sun, Jiashuo and others},
  journal={arXiv preprint arXiv:2602.07276},
  year={2026}
}

@misc{han2025personalityillusionrevealingdissociation,
      title={The Personality Illusion: Revealing Dissociation Between Self-Reports \& Behavior in LLMs}, 
      author={Pengrui Han and Rafal Kocielnik and Peiyang Song and Ramit Debnath and Dean Mobbs and Anima Anandkumar and R. Michael Alvarez},
      year={2025},
      eprint={2509.03730},
      archivePrefix={arXiv},
      primaryClass={cs.AI},
      url={https://arxiv.org/abs/2509.03730}, 
}

@article{zhao2026self,
  title={Self-Distilled Reasoner: On-Policy Self-Distillation for Large Language Models},
  author={Zhao, Siyan and Xie, Zhihui and Liu, Mengchen and Huang, Jing and Pang, Guan and Chen, Feiyu and Grover, Aditya},
  journal={arXiv preprint arXiv:2601.18734},
  year={2026}
}

@misc{ye2026onpolicycontextdistillationlanguage,
      title={On-Policy Context Distillation for Language Models}, 
      author={Tianzhu Ye and Li Dong and Xun Wu and Shaohan Huang and Furu Wei},
      year={2026},
      eprint={2602.12275},
      archivePrefix={arXiv},
      primaryClass={cs.CL},
      url={https://arxiv.org/abs/2602.12275}, 
}

@inproceedings{park2026choicemates,
  title={Choicemates: Supporting unfamiliar online decision-making with multi-agent conversational interactions},
  author={Park, Jeongeon and Min, Bryan and Son, Kihoon and Song, Jean Y and Ma, Xiaojuan and Kim, Juho},
  booktitle={Proceedings of the 31st International Conference on Intelligent User Interfaces},
  pages={1526--1550},
  year={2026}
}

@article{jiang2025adaptation,
  title={Adaptation of agentic ai},
  author={Jiang, Pengcheng and Lin, Jiacheng and Shi, Zhiyi and Wang, Zifeng and He, Luxi and Wu, Yichen and Zhong, Ming and Song, Peiyang and Zhang, Qizheng and Wang, Heng and others},
  journal={arXiv preprint arXiv:2512.16301},
  year={2025}
}

@inproceedings{xu2026zero,
  title={Zero-Shot Open-Schema Entity Structure Discovery},
  author={Xu, Xueqiang and Xiao, Jinfeng and Barry, James and El-karef, Mohab and Zou, Jiaru and Jiang, Pengcheng and Zhang, Yunyi and Giammona, Maxwell J and Mel, Geeth and Han, Jiawei},
  booktitle={Proceedings of the 19th Conference of the European Chapter of the Association for Computational Linguistics (Volume 1: Long Papers)},
  pages={7547--7561},
  year={2026}
}

@article{jansen2024discoveryworld,
  title={Discoveryworld: A virtual environment for developing and evaluating automated scientific discovery agents},
  author={Jansen, Peter and C{\^o}t{\'e}, Marc-Alexandre and Khot, Tushar and Bransom, Erin and Dalvi Mishra, Bhavana and Majumder, Bodhisattwa Prasad and Tafjord, Oyvind and Clark, Peter},
  journal={Advances in Neural Information Processing Systems},
  volume={37},
  pages={10088--10116},
  year={2024}
}

@article{yuksekgonul2024textgrad,
  title={Textgrad: Automatic ``differentiation'' via text},
  author={Yuksekgonul, Mert and Bianchi, Federico and Boen, Joseph and Liu, Sheng and Huang, Zhi and Guestrin, Carlos and Zou, James},
  journal={arXiv preprint arXiv:2406.07496},
  year={2024}
}

@article{shridhar2020alfworld,
  title={Alfworld: Aligning text and embodied environments for interactive learning},
  author={Shridhar, Mohit and Yuan, Xingdi and C{\^o}t{\'e}, Marc-Alexandre and Bisk, Yonatan and Trischler, Adam and Hausknecht, Matthew},
  journal={arXiv preprint arXiv:2010.03768},
  year={2020}
}

@article{tang2025chemagent,
  title={Chemagent: Self-updating library in large language models improves chemical reasoning},
  author={Tang, Xiangru and Hu, Tianyu and Ye, Muyang and Shao, Yanjun and Yin, Xunjian and Ouyang, Siru and Zhou, Wangchunshu and Lu, Pan and Zhang, Zhuosheng and Zhao, Yilun and others},
  journal={arXiv preprint arXiv:2501.06590},
  year={2025}
}

@article{ouyang2022training,
  title={Training language models to follow instructions with human feedback},
  author={Ouyang, Long and Wu, Jeffrey and Jiang, Xu and Almeida, Diogo and Wainwright, Carroll and Mishkin, Pamela and Zhang, Chong and Agarwal, Sandhini and Slama, Katarina and Ray, Alex and others},
  journal={Advances in neural information processing systems},
  volume={35},
  pages={27730--27744},
  year={2022}
}

@article{liu2024deepseek,
  title={Deepseek-v3 technical report},
  author={Liu, Aixin and Feng, Bei and Xue, Bing and Wang, Bingxuan and Wu, Bochao and Lu, Chengda and Zhao, Chenggang and Deng, Chengqi and Zhang, Chenyu and Ruan, Chong and others},
  journal={arXiv preprint arXiv:2412.19437},
  year={2024}
}

@article{rafailov2023direct,
  title={Direct preference optimization: Your language model is secretly a reward model},
  author={Rafailov, Rafael and Sharma, Archit and Mitchell, Eric and Manning, Christopher D and Ermon, Stefano and Finn, Chelsea},
  journal={Advances in neural information processing systems},
  volume={36},
  pages={53728--53741},
  year={2023}
}

@inproceedings{lightman2024let,
  title={Let's verify step by step},
  author={Lightman, Hunter and Kosaraju, Vineet and Burda, Yuri and Edwards, Harrison and Baker, Bowen and Lee, Teddy and Leike, Jan and Schulman, John and Sutskever, Ilya and Cobbe, Karl},
  booktitle={International Conference on Learning Representations},
  volume={2024},
  pages={39578--39601},
  year={2024}
}

@article{chen2024self,
  title={Self-play fine-tuning converts weak language models to strong language models},
  author={Chen, Zixiang and Deng, Yihe and Yuan, Huizhuo and Ji, Kaixuan and Gu, Quanquan},
  journal={arXiv preprint arXiv:2401.01335},
  year={2024}
}

@article{kumar2024training,
  title={Training language models to self-correct via reinforcement learning},
  author={Kumar, Aviral and Zhuang, Vincent and Agarwal, Rishabh and Su, Yi and Co-Reyes, John D and Singh, Avi and Baumli, Kate and Iqbal, Shariq and Bishop, Colton and Roelofs, Rebecca and others},
  journal={arXiv preprint arXiv:2409.12917},
  year={2024}
}

@article{shao2024deepseekmath,
  title={Deepseekmath: Pushing the limits of mathematical reasoning in open language models},
  author={Shao, Zhihong and Wang, Peiyi and Zhu, Qihao and Xu, Runxin and Song, Junxiao and Bi, Xiao and Zhang, Haowei and Zhang, Mingchuan and Li, YK and Wu, Yang and others},
  journal={arXiv preprint arXiv:2402.03300},
  year={2024}
}

@article{liu2025understanding,
  title={Understanding r1-zero-like training: A critical perspective},
  author={Liu, Zichen and Chen, Changyu and Li, Wenjun and Qi, Penghui and Pang, Tianyu and Du, Chao and Lee, Wee Sun and Lin, Min},
  journal={arXiv preprint arXiv:2503.20783},
  year={2025}
}

@article{zheng2025group,
  title={Group sequence policy optimization},
  author={Zheng, Chujie and Liu, Shixuan and Li, Mingze and Chen, Xiong-Hui and Yu, Bowen and Gao, Chang and Dang, Kai and Liu, Yuqiong and Men, Rui and Yang, An and others},
  journal={arXiv preprint arXiv:2507.18071},
  year={2025}
}

@article{cui2025entropy,
  title={The entropy mechanism of reinforcement learning for reasoning language models},
  author={Cui, Ganqu and Zhang, Yuchen and Chen, Jiacheng and Yuan, Lifan and Wang, Zhi and Zuo, Yuxin and Li, Haozhan and Fan, Yuchen and Chen, Huayu and Chen, Weize and others},
  journal={arXiv preprint arXiv:2505.22617},
  year={2025}
}

@inproceedings{trivedi2023interleaving,
  title={Interleaving retrieval with chain-of-thought reasoning for knowledge-intensive multi-step questions},
  author={Trivedi, Harsh and Balasubramanian, Niranjan and Khot, Tushar and Sabharwal, Ashish},
  booktitle={Proceedings of the 61st annual meeting of the association for computational linguistics (volume 1: long papers)},
  pages={10014--10037},
  year={2023}
}

@article{li2025search,
  title={Search-o1: Agentic search-enhanced large reasoning models},
  author={Li, Xiaoxi and Dong, Guanting and Jin, Jiajie and Zhang, Yuyao and Zhou, Yujia and Zhu, Yutao and Zhang, Peitian and Dou, Zhicheng},
  journal={arXiv preprint arXiv:2501.05366},
  year={2025}
}

@inproceedings{ouyang2025reasoningbank,
  title={Reasoningbank: Scaling agent self-evolving with reasoning memory},
  author={Ouyang, Siru and Yan, Jun and Hsu, I and Chen, Yanfei and Jiang, Ke and Wang, Zifeng and Han, Rujun and Le, Long T and Daruki, Samira and Tang, Xiangru and others},
  booktitle={International Conference on Learning Representations},
  volume={2026},
  year={2026}
}

@article{shinn2023reflexion,
  title={Reflexion: Language agents with verbal reinforcement learning},
  author={Shinn, Noah and Cassano, Federico and Gopinath, Ashwin and Narasimhan, Karthik and Yao, Shunyu},
  journal={Advances in Neural Information Processing Systems},
  volume={36},
  pages={8634--8652},
  year={2023}
}

@inproceedings{zhao2024expel,
  title={Expel: Llm agents are experiential learners},
  author={Zhao, Andrew and Huang, Daniel and Xu, Quentin and Lin, Matthieu and Liu, Yong-Jin and Huang, Gao},
  booktitle={Proceedings of the AAAI Conference on Artificial Intelligence},
  volume={38},
  number={17},
  pages={19632--19642},
  year={2024}
}

@inproceedings{rein2024gpqa,
  title={Gpqa: A graduate-level google-proof q\&a benchmark},
  author={Rein, David and Hou, Betty Li and Stickland, Asa Cooper and Petty, Jackson and Pang, Richard Yuanzhe and Dirani, Julien and Michael, Julian and Bowman, Samuel R},
  booktitle={First Conference on Language Modeling},
  year={2024}
}

@article{chen2021evaluating,
  title={Evaluating large language models trained on code},
  author={Chen, Mark and Tworek, Jerry and Jun, Heewoo and Yuan, Qiming and Pinto, Henrique Ponde de Oliveira and Kaplan, Jared and Edwards, Harri and Burda, Yuri and Joseph, Nicholas and Brockman, Greg and others},
  journal={arXiv preprint arXiv:2107.03374},
  year={2021}
}

@article{loshchilov2017decoupled,
  title={Decoupled weight decay regularization},
  author={Loshchilov, Ilya and Hutter, Frank},
  journal={arXiv preprint arXiv:1711.05101},
  year={2017}
}

@article{gsm8k,
  author       = {Karl Cobbe and
                  Vineet Kosaraju and
                  Mohammad Bavarian and
                  Mark Chen and
                  Heewoo Jun and
                  Lukasz Kaiser and
                  Matthias Plappert and
                  Jerry Tworek and
                  Jacob Hilton and
                  Reiichiro Nakano and
                  Christopher Hesse and
                  John Schulman},
  title        = {Training Verifiers to Solve Math Word Problems},
  journal      = {CoRR},
  year         = {2021},
  eprinttype    = {arXiv},
  eprint       = {2110.14168},
}

@inproceedings{gsm_sym,
  author       = {Iman Mirzadeh and
                  Keivan Alizadeh and
                  Hooman Shahrokhi and
                  Oncel Tuzel and
                  Samy Bengio and
                  Mehrdad Farajtabar},
  title        = {GSM-Symbolic: Understanding the Limitations of Mathematical Reasoning
                  in Large Language Models},
  booktitle    = {The Thirteenth International Conference on Learning Representations,
                  {ICLR} 2025, Singapore, April 24-28, 2025},
  year         = {2025}
}

@article{math,
  title={Measuring Mathematical Problem Solving With the MATH Dataset},
  author={Dan Hendrycks and Collin Burns and Saurav Kadavath and Akul Arora and Steven Basart and Eric Tang and Dawn Song and Jacob Steinhardt},
  journal={NeurIPS},
  year={2021}
}

@inproceedings{MMLU,
  author       = {Dan Hendrycks and
                  Collin Burns and
                  Steven Basart and
                  Andy Zou and
                  Mantas Mazeika and
                  Dawn Song and
                  Jacob Steinhardt},
  title        = {Measuring Massive Multitask Language Understanding},
  booktitle    = {9th International Conference on Learning Representations, {ICLR} 2021,
                  Virtual Event, Austria, May 3-7, 2021},
  year         = {2021}
}

@inproceedings{CommonsenseQA,
  author       = {Alon Talmor and
                  Jonathan Herzig and
                  Nicholas Lourie and
                  Jonathan Berant},
  title        = {CommonsenseQA: {A} Question Answering Challenge Targeting Commonsense
                  Knowledge},
  booktitle    = {Proceedings of the 2019 Conference of the North American Chapter of
                  the Association for Computational Linguistics: Human Language Technologies,
                  {NAACL-HLT} 2019, Minneapolis, MN, USA, June 2-7, 2019, Volume 1 (Long
                  and Short Papers)},
  pages        = {4149--4158},
  publisher    = {Association for Computational Linguistics},
  year         = {2019},
}

@inproceedings{OpenbookQA,
  author       = {Todor Mihaylov and
                  Peter Clark and
                  Tushar Khot and
                  Ashish Sabharwal},
  title        = {Can a Suit of Armor Conduct Electricity? {A} New Dataset for Open
                  Book Question Answering},
  booktitle    = {Proceedings of the 2018 Conference on Empirical Methods in Natural
                  Language Processing, Brussels, Belgium, October 31 - November 4, 2018},
  pages        = {2381--2391},
  publisher    = {Association for Computational Linguistics},
  year         = {2018}
}

@article{arc,
      author    = {Peter Clark  and Isaac Cowhey and Oren Etzioni and Tushar Khot and
                    Ashish Sabharwal and Carissa Schoenick and Oyvind Tafjord},
      title     = {Think you have Solved Question Answering? Try ARC, the AI2 Reasoning Challenge},
      journal   = {arXiv:1803.05457v1},
      year      = {2018},
}

@inproceedings{trivedi2024appworld,
  title={Appworld: A controllable world of apps and people for benchmarking interactive coding agents},
  author={Trivedi, Harsh and Khot, Tushar and Hartmann, Mareike and Manku, Ruskin and Dong, Vinty and Li, Edward and Gupta, Shashank and Sabharwal, Ashish and Balasubramanian, Niranjan},
  booktitle={Proceedings of the 62nd Annual Meeting of the Association for Computational Linguistics (Volume 1: Long Papers)},
  pages={16022--16076},
  year={2024}
}

@article{chhikara2025mem0,
  title={Mem0: Building production-ready ai agents with scalable long-term memory},
  author={Chhikara, Prateek and Khant, Dev and Aryan, Saket and Singh, Taranjeet and Yadav, Deshraj},
  journal={arXiv preprint arXiv:2504.19413},
  year={2025}
}

@article{wang2024agent,
  title={Agent workflow memory},
  author={Wang, Zora Zhiruo and Mao, Jiayuan and Fried, Daniel and Neubig, Graham},
  journal={arXiv preprint arXiv:2409.07429},
  year={2024}
}

@article{zhang2026memrl,
  title={Memrl: Self-evolving agents via runtime reinforcement learning on episodic memory},
  author={Zhang, Shengtao and Wang, Jiaqian and Zhou, Ruiwen and Liao, Junwei and Feng, Yuchen and Li, Zhuo and Zheng, Yujie and Zhang, Weinan and Wen, Ying and Li, Zhiyu and others},
  journal={arXiv preprint arXiv:2601.03192},
  year={2026}
}

@inproceedings{jiang2025s3,
  title={s3: You don’t need that much data to train a search agent via rl},
  author={Jiang, Pengcheng and Xu, Xueqiang and Lin, Jiacheng and Xiao, Jinfeng and Wang, Zifeng and Sun, Jimeng and Han, Jiawei},
  booktitle={Proceedings of the 2025 Conference on Empirical Methods in Natural Language Processing},
  pages={21599--21617},
  doi={10.18653/v1/2025.emnlp-main.1095},
  year={2025}
}

@article{yao2022react,
  title={React: Synergizing reasoning and acting in language models},
  author={Yao, Shunyu and Zhao, Jeffrey and Yu, Dian and Du, Nan and Shafran, Izhak and Narasimhan, Karthik and Cao, Yuan},
  journal={arXiv preprint arXiv:2210.03629},
  year={2022}
}

@inproceedings{shi2023large,
  title={Large language models can be easily distracted by irrelevant context},
  author={Shi, Freda and Chen, Xinyun and Misra, Kanishka and Scales, Nathan and Dohan, David and Chi, Ed H and Sch{\"a}rli, Nathanael and Zhou, Denny},
  booktitle={International conference on machine learning},
  pages={31210--31227},
  year={2023},
  organization={PMLR}
}

@inproceedings{jeong2024adaptive,
  title={Adaptive-rag: Learning to adapt retrieval-augmented large language models through question complexity},
  author={Jeong, Soyeong and Baek, Jinheon and Cho, Sukmin and Hwang, Sung Ju and Park, Jong C},
  booktitle={Proceedings of the 2024 conference of the north american chapter of the association for computational linguistics: Human language technologies (volume 1: Long papers)},
  pages={7036--7050},
  year={2024}
}

@inproceedings{sclar2024quantifying,
  title={Quantifying Language Models' Sensitivity to Spurious Features in Prompt Design or: How I learned to start worrying about prompt formatting},
  author={Sclar, Melanie and Choi, Yejin and Tsvetkov, Yulia and Suhr, Alane},
  booktitle={International Conference on Learning Representations},
  volume={2024},
  pages={25055--25083},
  year={2024}
}

@article{wei2023larger,
  title={Larger language models do in-context learning differently},
  author={Wei, Jerry and Wei, Jason and Tay, Yi and Tran, Dustin and Webson, Albert and Lu, Yifeng and Chen, Xinyun and Liu, Hanxiao and Huang, Da and Zhou, Denny and others},
  journal={arXiv preprint arXiv:2303.03846},
  year={2023}
}

@inproceedings{mallen2023not,
  title={When not to trust language models: Investigating effectiveness of parametric and non-parametric memories},
  author={Mallen, Alex and Asai, Akari and Zhong, Victor and Das, Rajarshi and Khashabi, Daniel and Hajishirzi, Hannaneh},
  booktitle={Proceedings of the 61st annual meeting of the association for computational linguistics (volume 1: Long papers)},
  pages={9802--9822},
  year={2023}
}

@inproceedings{fang2026lightmem,
  title={Lightmem: Lightweight and efficient memory-augmented generation},
  author={Fang, Jizhan and Deng, Xinle and Xu, Haoming and Jiang, Ziyan and Tang, Yuqi and Xu, Ziwen and Deng, Shumin and Yao, Yunzhi and Wang, Mengru and Qiao, Shuofei and others},
  booktitle={International Conference on Learning Representations},
  volume={2026},
  pages={98706--98729},
  year={2026}
}

@article{gutierrez2024hipporag,
  title={Hipporag: Neurobiologically inspired long-term memory for large language models},
  author={Guti{\'e}rrez, Bernal J and Shu, Yiheng and Gu, Yu and Yasunaga, Michihiro and Su, Yu},
  journal={Advances in neural information processing systems},
  volume={37},
  pages={59532--59569},
  year={2024}
}

@inproceedings{cuconasu2024power,
  title={The power of noise: Redefining retrieval for rag systems},
  author={Cuconasu, Florin and Trappolini, Giovanni and Siciliano, Federico and Filice, Simone and Campagnano, Cesare and Maarek, Yoelle and Tonellotto, Nicola and Silvestri, Fabrizio},
  booktitle={Proceedings of the 47th international ACM SIGIR conference on research and development in information retrieval},
  pages={719--729},
  year={2024}
}

@article{schulman2017proximal,
  title={Proximal policy optimization algorithms},
  author={Schulman, John and Wolski, Filip and Dhariwal, Prafulla and Radford, Alec and Klimov, Oleg},
  journal={arXiv preprint arXiv:1707.06347},
  year={2017}
}

@inproceedings{kwon2023efficient,
  title={Efficient memory management for large language model serving with pagedattention},
  author={Kwon, Woosuk and Li, Zhuohan and Zhuang, Siyuan and Sheng, Ying and Zheng, Lianmin and Yu, Cody Hao and Gonzalez, Joseph and Zhang, Hao and Stoica, Ion},
  booktitle={Proceedings of the 29th symposium on operating systems principles},
  pages={611--626},
  year={2023}
}

@article{liu2023your,
  title={Is your code generated by chatgpt really correct? rigorous evaluation of large language models for code generation},
  author={Liu, Jiawei and Xia, Chunqiu Steven and Wang, Yuyao and Zhang, Lingming},
  journal={Advances in neural information processing systems},
  volume={36},
  pages={21558--21572},
  year={2023}
}

@article{guo2025deepseek,
  title={Deepseek-r1: Incentivizing reasoning capability in llms via reinforcement learning},
  author={Guo, Daya and Yang, Dejian and Zhang, Haowei and Song, Junxiao and Wang, Peiyi and Zhu, Qihao and Xu, Runxin and Zhang, Ruoyu and Ma, Shirong and Bi, Xiao and others},
  journal={arXiv preprint arXiv:2501.12948},
  year={2025}
}

@techreport{anthropic2025claude4,
  author      = {Anthropic},
  title       = {System Card: {Claude Opus 4} \& {Claude Sonnet 4}},
  institution = {Anthropic},
  year        = {2025},
  month       = {May},
  url         = {https://www-cdn.anthropic.com/4263b940cabb546aa0e3283f35b686f4f3b2ff47.pdf}
}

@article{grattafiori2024llama,
  title={The llama 3 herd of models},
  author={Grattafiori, Aaron and Dubey, Abhimanyu and Jauhri, Abhinav and Pandey, Abhinav and Kadian, Abhishek and Al-Dahle, Ahmad and Letman, Aiesha and Mathur, Akhil and Schelten, Alan and Vaughan, Alex and others},
  journal={arXiv preprint arXiv:2407.21783},
  year={2024}
}

@article{yang2025qwen3,
  title={Qwen3 technical report},
  author={Yang, An and Li, Anfeng and Yang, Baosong and Zhang, Beichen and Hui, Binyuan and Zheng, Bo and Yu, Bowen and Gao, Chang and Huang, Chengen and Lv, Chenxu and others},
  journal={arXiv preprint arXiv:2505.09388},
  year={2025}
}

@techreport{google2026gemini31flashlite,
  author      = {Google DeepMind},
  title       = {Gemini 3.1 {Flash-Lite} Model Card},
  institution = {Google DeepMind},
  year        = {2026},
  month       = {March},
  url         = {https://deepmind.google/models/model-cards/gemini-3-1-flash-lite/}
}
